\documentclass[10pt,twocolumn,letterpaper]{article}

\usepackage{iccv}
\usepackage{times}
\usepackage{epsfig}
\usepackage{graphicx}
\usepackage{amsmath}
\usepackage{amssymb}
\usepackage{mathcomp}
\usepackage[normalem]{ulem}
\usepackage{xcolor}
\usepackage{todonotes}
\usepackage[group-separator={,}]{siunitx}
\usepackage{bm}
\usepackage{booktabs}
\usepackage{capt-of}
\usepackage{enumitem}
\usepackage[accsupp]{axessibility}

\usepackage[pagebackref=true,breaklinks=true,letterpaper=true,colorlinks,bookmarks=false]{hyperref}

\usepackage[capitalize]{cleveref}
\crefname{section}{Sec.}{Secs.}
\Crefname{section}{Section}{Sections}
\Crefname{table}{Table}{Tables}
\crefname{table}{Tab.}{Tabs.}

\DeclareRobustCommand{\removed}{\bgroup\markoverwith{\color{red}\rule[.5ex]{2pt}{1pt}}\ULon} 

\iccvfinalcopy 

\def\iccvPaperID{4819} 

\ificcvfinal\fi

\begin{document}

\title{DeePoint: Visual Pointing Recognition and Direction Estimation}

\author{Shu Nakamura$^*$\quad Yasutomo Kawanishi$^\dagger$ \quad Shohei Nobuhara$^*$ \quad Ko Nishino$^{*,\dagger}$\\
\begin{tabular}{cc}
$^*$Graduate School of Informatics, Kyoto University     & $^\dagger$RIKEN \\
https://vision.ist.i.kyoto-u.ac.jp/     &
https://grp.riken.jp/
\end{tabular}
}


\makeatletter
\let\@oldmaketitle\@maketitle%
\renewcommand{\@maketitle}{
    \@oldmaketitle%
    \centering
    \includegraphics[width=\linewidth]{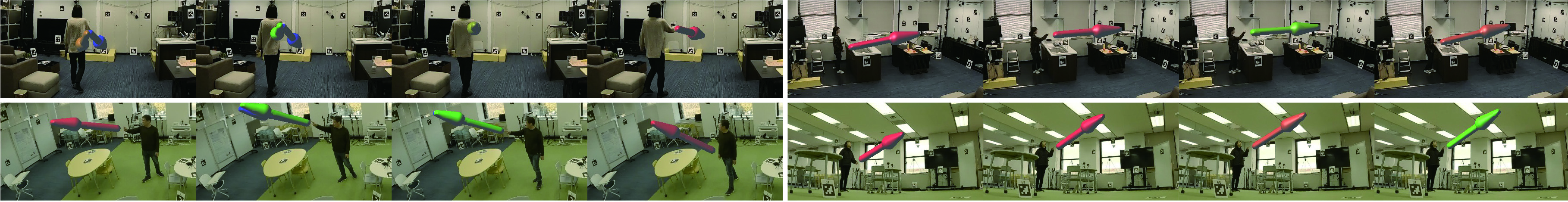}
\captionof{figure}{We introduce DeePoint, a neural pointing recognition and 3D direction estimator. DeePoint trained on our newly constructed DP Dataset recognizes when a person is pointing and estimates its 3D direction from video frames captured from a fixed-view camera. Each arrow depicts the pointing direction and its color is green when the person is pointing, red when not. DeePoint successfully recognizes when a pointing starts and ends and can estimate its 3D direction from the complex spatio-temporal coordination of the person's body.}
    \label{fig:opening}
    \vspace{4mm}
}
\makeatother

\maketitle
\ificcvfinal\thispagestyle{empty}\fi

\begin{abstract}

In this paper, we realize automatic visual recognition and direction estimation of pointing. We introduce the first neural pointing understanding method based on two key contributions. The first is the introduction of a first-of-its-kind large-scale dataset for pointing recognition and direction estimation, which we refer to as the DP Dataset. DP Dataset consists of more than 2 million frames of 33 people pointing in various styles annotated for each frame with pointing timings and 3D directions. The second is DeePoint, a novel deep network model for joint recognition and 3D direction estimation of pointing. DeePoint is a Transformer-based network which fully leverages the spatio-temporal coordination of the body parts, not just the hands. Through extensive experiments, we demonstrate the accuracy and efficiency of DeePoint. We believe DP Dataset and DeePoint will serve as a sound foundation for visual human intention understanding. 

\end{abstract}

\section{Introduction}

Gauging a person's intent from passive visual observations is one of the key goals of computer vision research. Successful visual intent understanding would be essential for a wide range of applications including personal assistance, elderly care, and surveillance. Visual recognition of the gesticulations of a person is essential for this as they directly express those intents. Pointing, the act of extending one's (usually index) finger towards something in the person's view to call attention to it, is particularly important as it conveys explicit information about the person's interactions with the environment including conversations with others.

Despite the broad interest in gesture recognition, research on visual understanding of pointing has been surprisingly limited. Visual pointing interpretation requires both recognition (is the person pointing) and direction estimation (which direction is the person pointing). Past works have relied on special cameras, such as RGB-D sensors, or required the person to point in a specific way. For in-the-wild natural pointing understanding, we must be able to recognize and estimate their directions in 3D from regular RGB cameras. A typical scenario we consider is a person in a room pointing at various things around her while freely moving around, which is observed by cameras fixed to room corners.

Pointing recognition and direction estimation from fixed-view cameras is particularly challenging. The person is usually small in the view and the fingers can hardly be discerned. The hand can even be completely occluded by the person's body. The pointing gesture would also typically span only about half a second, which makes its recognition in the video hard. Estimating the direction becomes even more challenging. In a full HD video frame captured with a fixed corner camera in a typical living room, the index finger would span only about 30 pixels. Analytical modeling such as line regression to such observations would be futile. Even if that were possible, due to intra- and inter-personal variations of pointing, such estimations would be prone to error. Accounting for those variations would naturally necessitate a learning-based approach that directly regresses to the intended directions. This is also, however, not straightforward, as the task is inherently spatio-temporal and, most important, large-scale data of pointing is difficult to collect and currently devoid in the community. 

In this paper, as illustrated in \cref{fig:opening}, we make two key contributions to realize automatic visual recognition and direction estimation of pointing. The first is the introduction of a comprehensive dataset for pointing recognition and direction estimation. We refer to this as the DP Dataset. It consists of \num{2800000} frames of \num{33} people pointing at various directions in different styles captured in \num{2} different rooms. Each of these frames is annotated with whether the person is pointing or not, and, when pointing, the 3D direction intended by the pointing person. This first of its kind large-scale collection and annotation of natural pointing gestures is achieved semi-automatically with a combination of multi-view geometry and audio processing.

The second key contribution is DeePoint, a novel deep network model for joint recognition and 3D direction estimation of pointing. To overcome the challenges stemming from the fixed-view observations from a distance, our key idea is to leverage the whole-body appearance and motion to detect and estimate 3D pointing. For this, we introduce a Transformer model, inspired by the STLT~\cite{Radevski2021}, that fully leverages the spatio-temporal coordination of the body including the head and joints in addition to the hand. By incorporating the appearance of these as tokens and through cascaded attention transforms in space and time, we show that pointing gestures can be detected in time and their 3D directions can be estimated accurately. 

We conduct extensive experiments to evaluate the effectiveness of DeePoint. We first evaluate the accuracy of recognition and direction estimation on the DP Dataset. We then evaluate the generalizability of DeePoint across different people and scenes. Through ablation studies, we also show that the spatio-temporal modeling of the body appearance and movements are essential for the task. We also conduct comparative studies with related works, including evaluation on the PKU-MMD dataset~\cite{Liu17_PKUMMD}. The experimental results collectively demonstrate the accuracy and efficiency of DeePoint. Our future work includes incorporating environmental cues and audio including spoken words to enhance the accuracy of pointing direction estimation, the challenge of which lies in realizing this without overfitting to the particular context. We believe DeePoint provides a sound foundation for these further studies.

\section{Related Works}

Gesture recognition has been a major topic of research in the computer vision community but research specific to pointing recognition and its direction estimation is fairly limited. We review works relevant to our approach of using the whole body for visual pointing understanding and also on construction of large-scale real-world datasets for human behavior understanding. 
\subsection{Pointing Recognition}

Early works of pointing recognition used wearable devices to measure pointing directions directly.
Various devices such as magnetometers~\cite{Bolt:1980} and IMUs~\cite{Denis2018} have been adopted.
Since the person to be measured must wear a dedicated device for pointing, however, the applications of these methods were limited.

Most past pointing recognition methods require special camera setups. These works include those that require multiple cameras~\cite{Fukumoto1992,Watanabe2000,Cipolla1994,Fujita2015,Roland2004,Nickel2003}, RGB-D cameras~\cite{Shukla:2015,Fernandez:2015,Azari2019,Dhingra2020}, or depth sensors~\cite{Das2018,Das2021}. From the visual observations captured with these specialized cameras or setups, these methods estimate pointing direction in mainly two ways: geometry-based or learning-based.

Geometry-based approaches first locate 3D coordinates of specific body parts, \eg, face, hand, or fingertip, and calculate the direction of pointing by extending the line connecting them. Results of such methods can be very noisy as the detection and triangulation of these body parts can be unreliable. Learning-based approaches estimate the 3D direction from the observed appearance of the body parts, \eg, hands or arms.
Both of these approaches can achieve accurate pointing direction recognition when certain imaging conditions are met, \eg, the person is in fairly near distance from the camera and showing a perfect side-view of the pointing. They, however, fundamentally rely on multi-view observations or direct depth perception, which preclude their use with regular RGB cameras.

A few recent works achieve pointing recognition from a single RGB image to alleviate the needs for special camera settings. Estimating pointing direction in 3D is inherently challenging, as 3D locations of body parts or detailed appearance of hands cannot be captured by a normal camera. Past works resolve this by limiting the allowed postures of the target person, for instance, by requiring the person to stand upright with her arm fully extended when pointing~\cite{Zuzana2007,Shiratori2021}. Jaiswal et al.~\cite{Shruti2018} introduced a ConvNet pointing direction estimation, but is limited to when the person is standing in front the camera with her body facing towards it. These methods, in essence, recognize a special pre-defined body posture as pointing, which does not generalize across people and scenes. Our DeePoint, in contrast, realizes automatic visual recognition and direction estimation of pointing by a person freely moving, regardless of walking or sitting, in a room-size area from a single view of a regular camera. To our knowledge, this is the first work to achieve 3D pointing understanding in the wild.

\subsection{Action Recognition}

Pointing can be viewed as a special gesture or action. General gesture and action recognition research has a long history in computer vision. Many benchmark datasets have been released, such as UCF101~\cite{Khurram_2012_arxiv} and ActivityNet~\cite{caba2015activitynet}. As spatio-temporal visual information becomes crucial for recognizing actions, a variety of approaches for capturing temporal relations of body and other contextual movements have been proposed such as CNN+LSTM~\cite{Donahue_2015_CVPR} and 3DResNet~\cite{Hara_2017_ICCV}.

More recently, Transformers~\cite{VaswaniSPUJGKP17} have been applied to learn such spatio-temporal coordination through their attention mechanism~\cite{Wang_2021_ICCV,Yang_2022_CVPR}. Among them, Radevski et al.~\cite{Radevski2021} proposed a two-stage transformer model which captures spatial relationships of object with the first transformer for each frame and temporal relationships of their movements with the second transformer. We build on this idea of decoupling spatial and temporal information aggregation with two cascaded Transformer encoders and extend it to encoding body postures and their temporal coordination to achieve accurate pointing recognition and direction estimation.

\subsection{3D Direction Annotation}

For learning-based 3D direction estimation, annotation of images with 3D vectors becomes essential. This task is, however, extremely challenging, if not impossible, to achieve manually as the annotator needs to somehow indicate the projected 2D direction from a 3D ground truth in mind on the 2D image plane. Past works have mitigated this difficulty by exploring automatic means to directly obtain the 3D ground truth. Das~\cite{Das2018, Das2021} attached a colored marker or an IMU to the index finger to obtain ground-truth pointing directions. This is possible for their method as they rely on direct depth perception for pointing recognition and artificial appearance of the person does not affect the input.

Other methods leverage multi-view geometry of cameras to compute 3D directions of 3D gaze and pointing.
Kellnhofer \etal~\cite{gaze360_2019} proposed Gaze360, a large-scale 3D gaze tracking dataset. The data was collected with an omnidirectional camera that simultaneously captures subjects and their gaze targets. By using an AR marker as the gaze target, the authors realize automatic annotation of the 3D location of the target. Nonaka \etal~\cite{Nonaka_2022_CVPR} introduced GAFA, a 3D gaze dataset with per-frame 3D gaze annotations. The gaze directions were captured with an eyeglass gaze tracker. For the ground truth head and body orientations, they used body- and head-mounted cameras and AR markers attached to compute the 3D orientations via SLAM. We automatically annotate our DP Dataset with accurate 3D pointing directions by identifying the pointed AR markers in the scene from audio and by computing the 3D directions to them with multi-view geometry. We also obtain the pointing timing and duration with synchronized audio. We believe this multi-modal automatic annotation would be useful in other dataset annotation tasks.

\section{DP Dataset}

Our first key contribution is the first-of-its-kind large-scale dataset for pointing recognition and 3D direction estimation. We make this dataset and code available to the public\footnote{\url{https://github.com/kyotovision-public/deepoint}}.

\subsection{Dataset Capture}
A large-scale dataset of videos capturing people pointing in various directions as they naturally roam around and sit and stand in an environment with accurate timing and 3D direction annotations is essential for exploring learning-based approaches to visual pointing understanding. The dataset desiderata include variations in the people spanning age and gender, the viewpoint and viewing directions, the pointing styles and timings including duration, the pointed directions, the behaviors of people such as standing, walking, and sitting, and the overall environments in which the people are immersed. Also, in order to use the natural appearance of people, they should not wear specific measurement devices that affect their appearance, such as motion capture devices, special markers, or gaze measurement devices, as a learning-based approach would overfit to them. To the best of our knowledge, there are no large-scale public datasets for pointing recognition and direction estimation that fulfill these.

We introduce \textit{DP Dataset}, a first-of-its-kind large-scale pointing dataset, which consists of \num{2800000} frames of \num{33} people of various ages and different genders pointing in a wide range of directions in different styles in two different rooms captured from a variety of viewing directions with multiple fixed-view cameras at room-scale distances. Most important, the dataset includes annotations of pointing timings and their 3D directions for each and every frame.

\begin{figure}[t]
    \centering
    \begin{tabular}{cc}
    \includegraphics[width=0.45\linewidth]{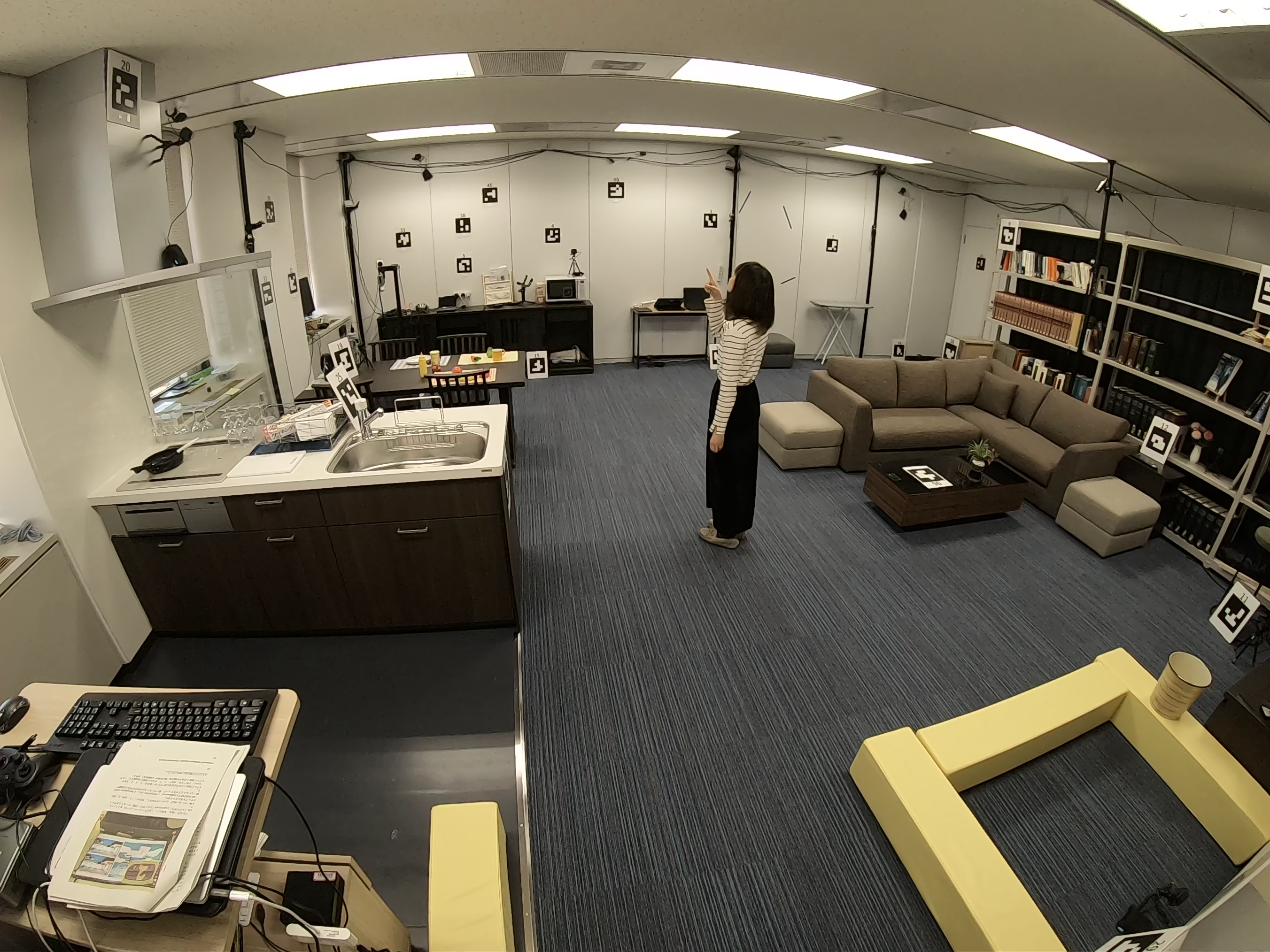}&
    \includegraphics[width=0.45\linewidth]{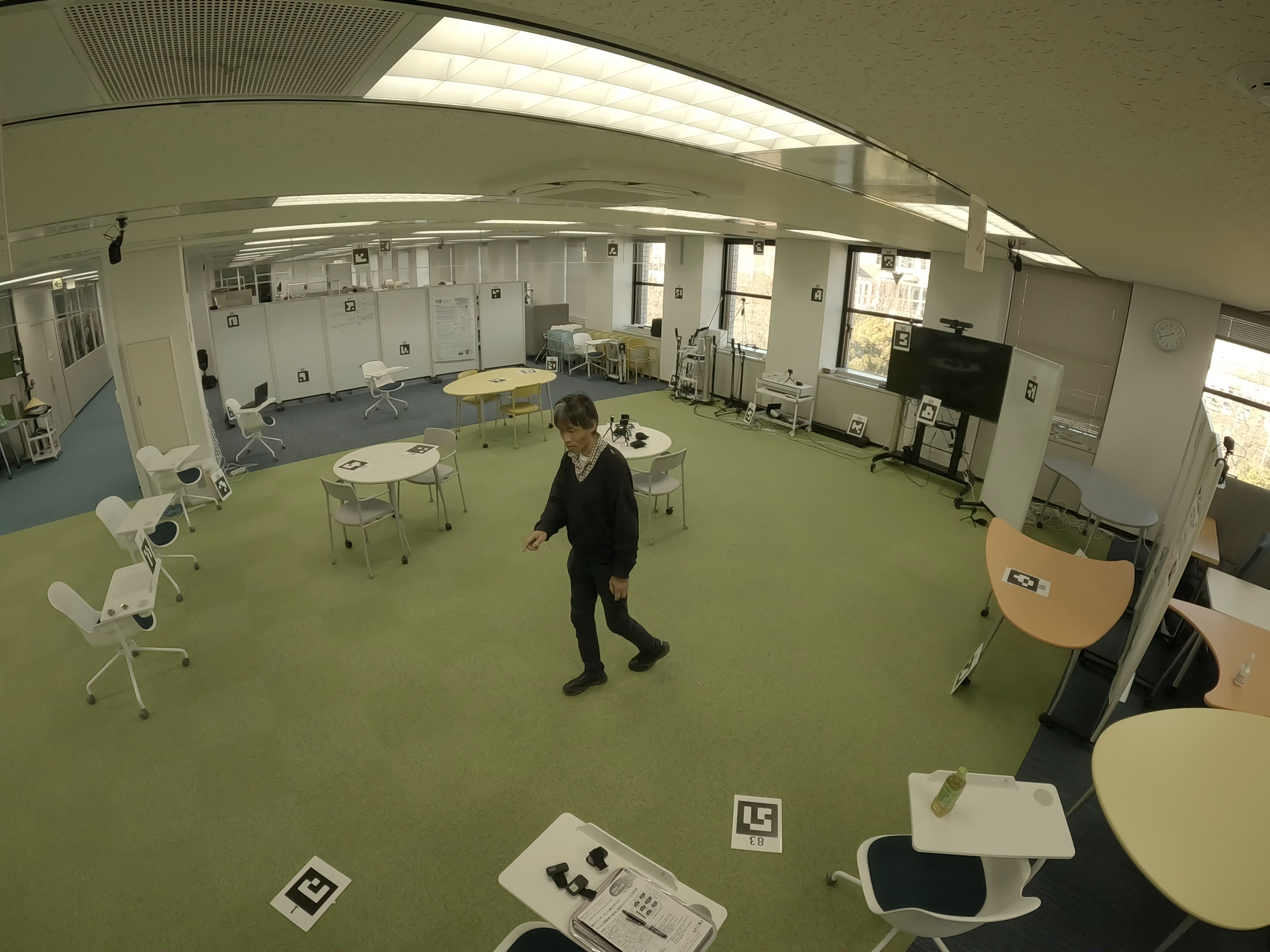}\\
    (i) Living Room & (ii) Office
    \end{tabular}
    \caption{The two environments of DP Dataset. The example frames are captured by the cameras outlined in red in Fig.~\ref{fig:camera_layouts}.}
    \label{fig:environments}
\end{figure}
\begin{figure}[t]
    \centering
    \begin{tabular}{cc}
    \includegraphics[width=0.4\linewidth]{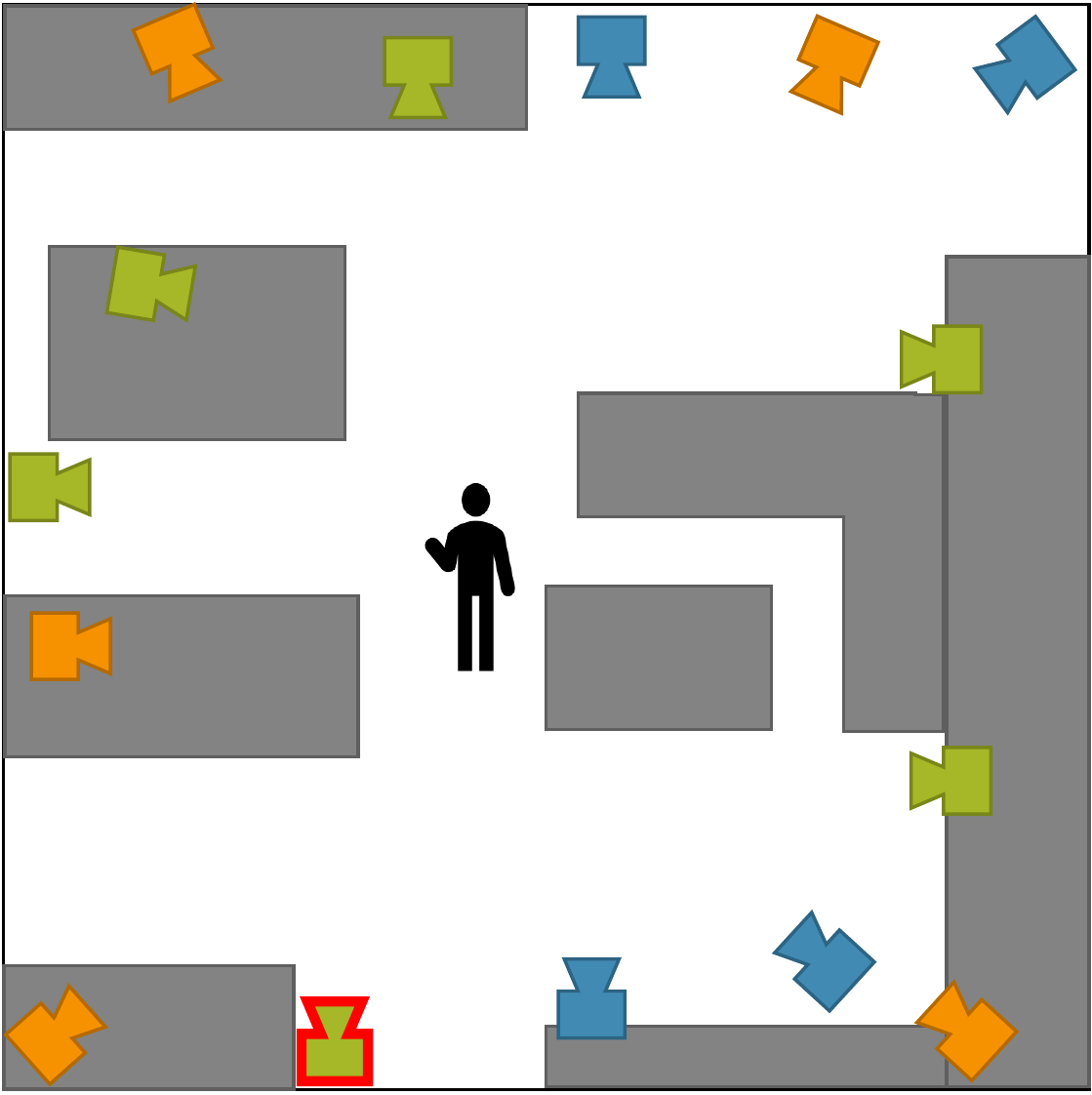}&
    \includegraphics[width=0.4\linewidth]{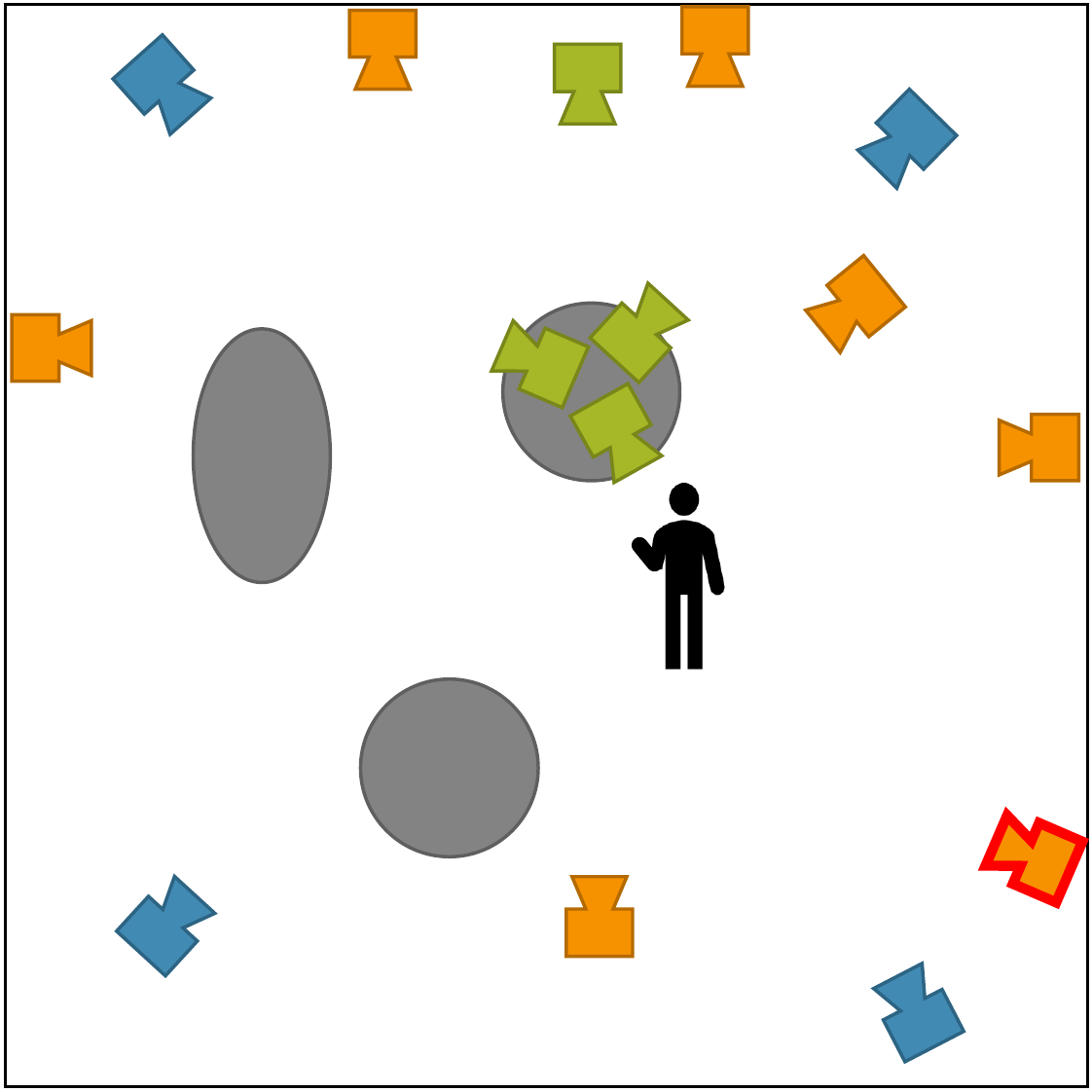}\\
    (i) Living Room & (ii) Office
    \end{tabular}
    \caption{Camera layout of DP Dataset. We mount cameras at fixed viewpoints in the room to capture the pointing gestures from a variety of directions at once. The orange cameras are installed on the ceiling, the blue ones are put on the floor, and the green ones are installed on the mid-level. The gray objects depict tables, sofas, and obstacles.}
    \label{fig:camera_layouts}
\end{figure}
\begin{figure}
    \centering
    \includegraphics[width=\linewidth]{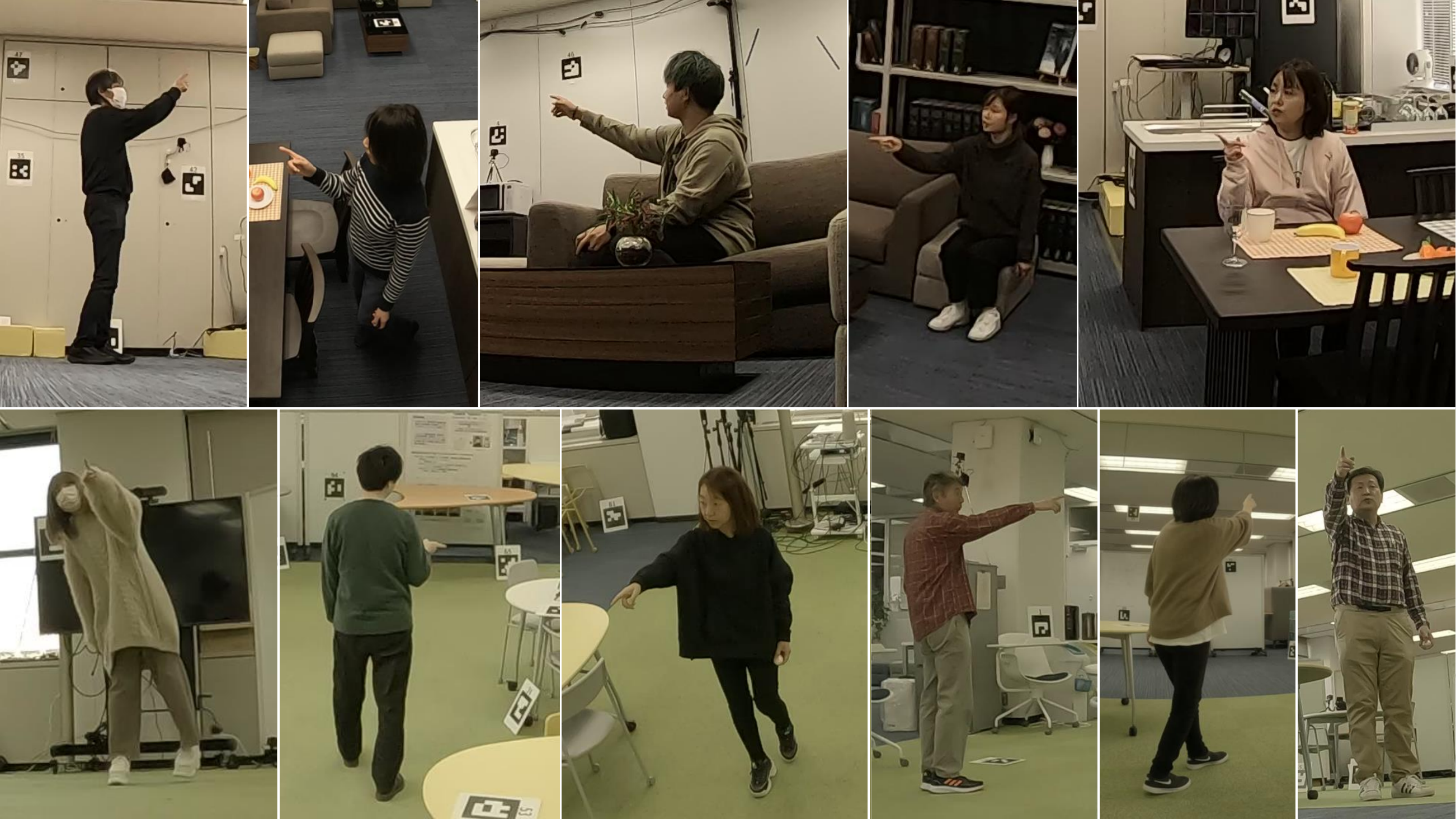}
    \caption{Example frames from the DP Dataset (people are cropped).}
    \label{fig:dpdataset}
\end{figure}

As shown in \cref{fig:environments}, we constructed a data capture imaging setup for two different rooms. One is a living room with a kitchen and sofa, and another is an open office with chairs, desks, and whiteboard, which we refer to as Living Room and Office, respectively. Both rooms are about \SI{64}{\meter\squared}. As depicted in \cref{fig:camera_layouts}, we installed 15 GoPro cameras in each room to capture people in them and calibrated all the cameras so that we could triangulate the 3D position of each joint and marker in the environment. They were installed in various locations in the room pointing towards the center so that a person in the room can be captured from all directions roughly uniformly. For this, we mounted the cameras on the tables, walls, the floor, and the ceiling. All cameras were synchronized at the beginning of the capture.

We captured videos in 2.7K resolution at 60fps and drop the frame rate to 15fps for the dataset we use for our experiments. The raw dataset can also be released upon request. To annotate the pointing direction, we installed roughly 40 ArUco~\cite{Romeroramirez_2018_IMAVIS} markers randomly on tables, walls, the floor, and the ceiling, in each room. Each marker is observed by multiple cameras and its 3D location is recovered with triangulation. 

\begin{figure}[t]
    \centering
    \includegraphics[width=\linewidth,bb={0 0 1554 879},clip]{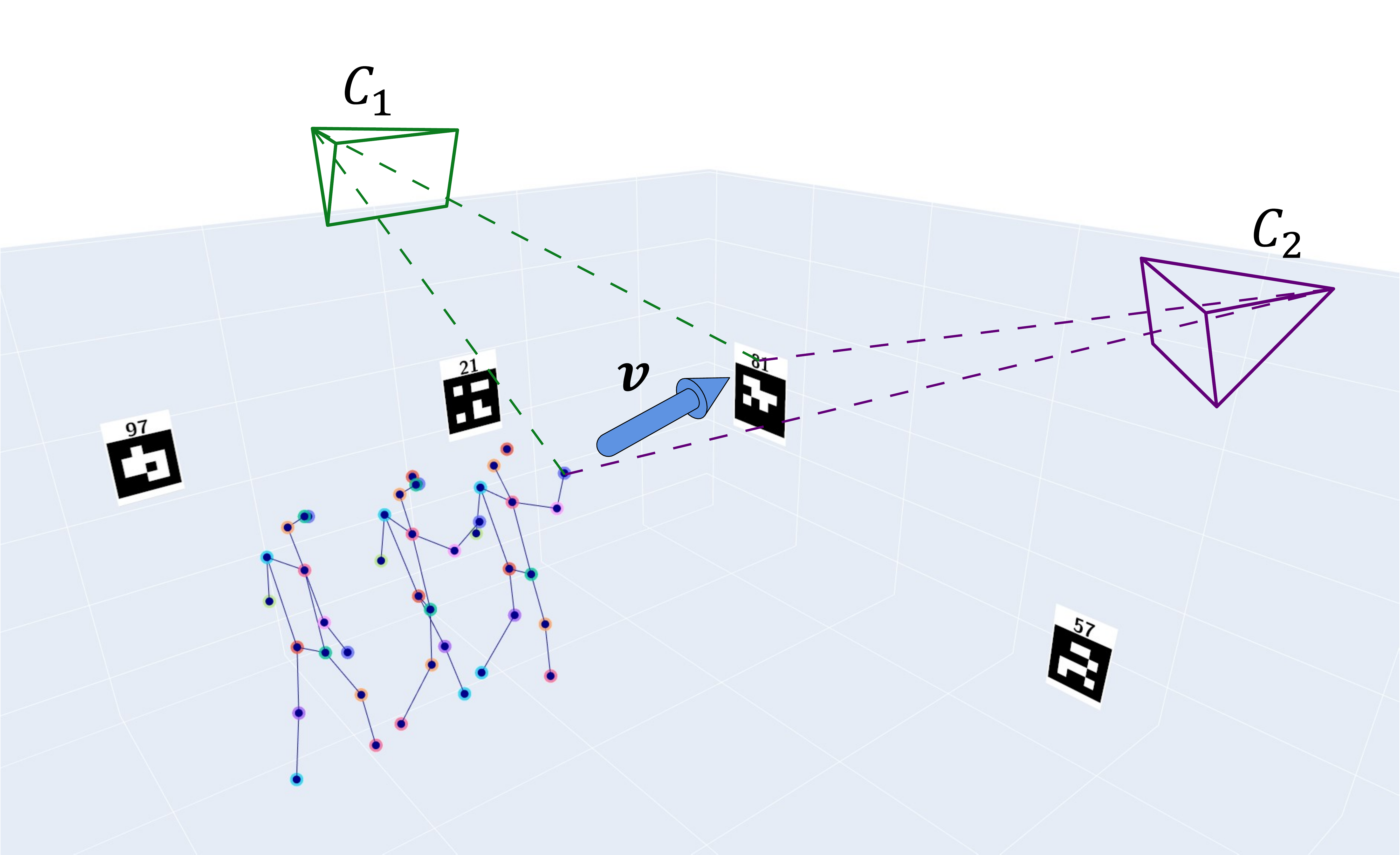}
    \caption{By identifying the marker to which the person is pointing from recorded audio and triangulating the hand location, we compute the unit 3D vector annotation for the 3D pointing direction.}
    \label{fig:dataset_triangulation}
\end{figure}

As shown in \cref{fig:dpdataset}, pointing style varies from person to person and the dataset should capture this variation as much as possible. For the dataset, we collected a total of 33 male and female participants, uniformly ranging in generations from their twenties to sixties. We captured each participant separately for about 5 minutes in each room. Each participant was free to walk around the room and point to markers freely selected by themselves with their dominant hand, but asked to verbalize the marker ID and click and hold down on a handheld wireless mouse when pointing. They were also allowed to point to the markers while sitting on a chair or sofa. They pointed to a marker once every 3 to 5 seconds while moving in the room. For each session, we collected 15 videos from the different fixed-view cameras.

\subsection{Pointing Timing and 3D Direction Annotation}

We fully annotate DP Dataset with pointing timings, \ie, the start and end of a pointing instance, and the 3D directions for all pointing instances. We annotate the pointing timings by asking the participants to indicate the start and duration of when he or she points to a marker. This is achieved by providing the participants with a small click button, for which we simply used a tiny wireless mouse, held in the non-dominant hand so that it is not visible from the camera. Participants pressed the button when they started pointing to a marker and held it down until their pointing gesture finished. The duration is typically less than a second for a natural pointing behavior. 

As depicted in \cref{fig:dataset_triangulation}, we automatically annotate the 3D directions of each pointing instance with multi-view geometry. Participants were asked to verbally express which marker they were pointing to, whose voice was recorded by the observing cameras. By manually identifying the marker ID from the recorded voice, we know the 3D coordinates of the target pointing direction. To recover the other end of the 3D vector, \ie, from where that marker is pointed, we first apply 2D pose estimation to the videos captured by the cameras and calculate the 3D hand locations based on triangulation using only high-confident 2D pose estimation results. We use OpenPifPaf~\cite{Kreiss2021} as the pose estimator, but any method that is sufficiently accurate can replace it. Accurate pointing directions were calculated from each pair of a 3D hand location and the pointed 3D marker location. We describe each pointing direction as a 3D unit vector.

\begin{figure}[t]
    \centering
    \includegraphics[width=\linewidth,bb=0 0 576 239,clip]{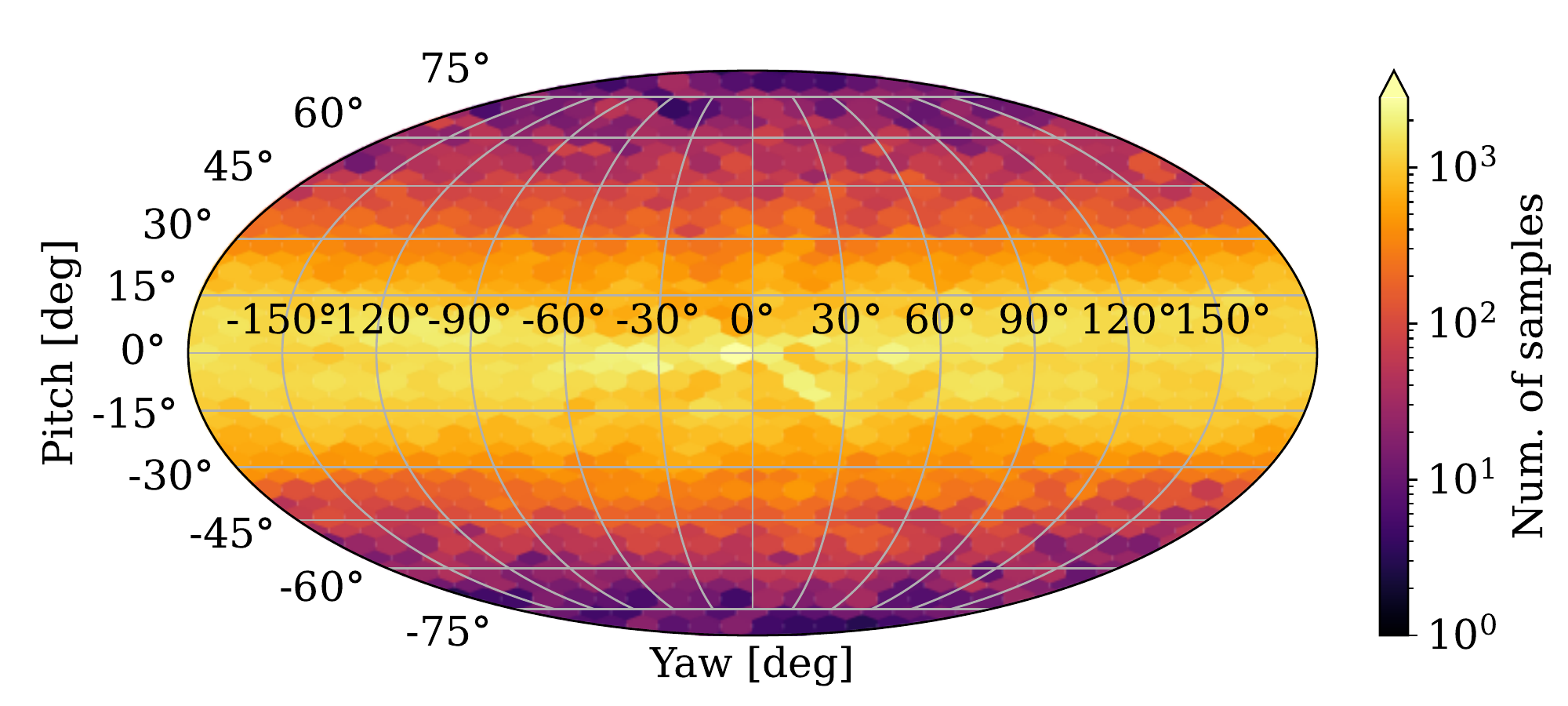}
    \caption{The 3D angular distribution of pointing directions in the DP Dataset shown with Mollweide projection. A variety of pointing behaviors with a wide range of pitch and yaw are captured in the dataset.}
    \label{fig:projection_world}
\end{figure}

In total, the dataset contains about \num{2800000} frames of 33 people. Frames with pointing span \num{770000} frames capturing \num{6355} unique pointing instances. Although the number of located ArUco markers in a room was limited to about 40, we were able to collect a large variety of pointing directions in the dataset as the participants were allowed to move around and change their postures freely in the rooms. \Cref{fig:projection_world} shows the 3D angular distribution of pointing directions. The histogram clearly shows that we were able to capture a wide variety of pointings in the dataset. Note that each of these instances are captured with a wide range of viewing directions using the 15 cameras. 

\section{DeePoint}
\def\JE{Joint Encoder\xspace}
\def\TE{Temporal Encoder\xspace}

\begin{figure}[t]
    \centering
    \includegraphics[width=\linewidth]{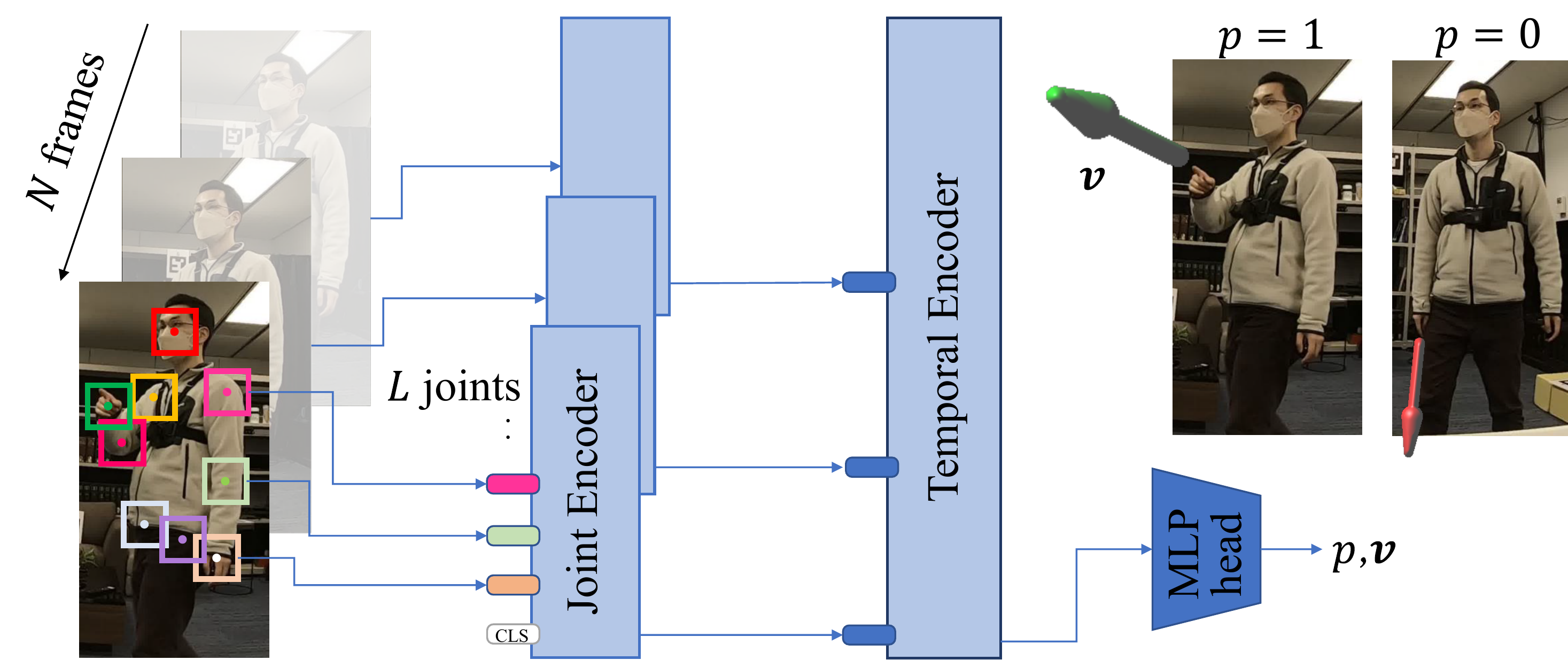}
    \caption{DeePoint consists of two Transformer encoders which we refer to as \JE and \TE. \JE learns to model the spatial coordination of body parts and \TE learns to extract their temporal coordination to jointly recognize and estimate the 3D direction of pointing from RGB video frames.}
    \label{fig:model architecture}
\end{figure}

We introduce DeePoint, a novel method for accurate pointing recognition and 3D direction estimation. Unlike past works, the method does not rely on specific poses taken by the target person and only requires regular RGB video frames as input. As depicted in \cref{fig:model architecture}, DeePoint is a Transformer-based model which leverages attention for spatio-temporal information aggregation as first introduced for video understanding by Radevski \etal~\cite{Radevski2021}. In contrast to learning the spatio-temporal coordination of objects in a scene for video understanding, we leverage the STLT architecture~\cite{Radevski2021} to learn the structured spatio-temporal coordination of body parts of a person when she is pointing and simultaneously detect and estimate its 3D direction. Given a sequence of input frames, DeePoint first detects the joints using an off-the-shelf 2D human pose estimator~\cite{Kreiss2021}, and extracts visual features around them. The visual features are first processed by \JE in a frame-wise manner, and then fed to \TE to integrate features from multiple frames. The output of \TE is transformed by an MLP head to the probability $p$ indicating whether the target is in a pointing action, and its 3D direction $\bm \nu$ in the camera coordinate system.

\paragraph{Pose Feature Extraction}
We use the output of the third block of ResNet-34~\cite{he2016deep} pre-trained by ImageNet-1K~\cite{deng2009imagenet} as the backbone for extracting \num{256}-channel visual features from each of the input video frames, and apply ROI align~\cite{He_2017_ICCV} around each joint to obtain \numproduct{3 x 3 x 256} feature vectors of a constant size regardless of the apparent joint size. These feature vectors are then projected by a learnable linear layer to \num{192} dimensions.

Though these visual features around the joints collectively cover a certain area over the target person, they do not explicitly describe the relative positions between the joints, \ie, the pose. We encode the 2D pose information by two additional features per joint: the joint indices and the 2D relative position \wrt the midpoint of the shoulders normalized by the bounding box size. Both the joint indices and the normalized relative positions are projected to the size of the visual features so that they are added as positional encodings~\cite{VaswaniSPUJGKP17}. These projections are randomly initialized and refined in the training.

For undetected joints or joints with confidence below a certain threshold due to, for example, occlusions, corresponding visual features are padded with a dummy tensor and ignored in the subsequent steps by masked attention~\cite{VaswaniSPUJGKP17}.

\paragraph{\JE}

The \JE takes the visual features of human joints and a class token, and computes multi-head attention~\cite{VaswaniSPUJGKP17} between them. We set \JE to accept $L=17$ tokens corresponding to the joints detected by pose estimation. \JE processes such tokens with 6 iterations of attention layers, and returns the output corresponding to the class token for the last attention layer as the final output.

\paragraph{\TE}

Our \TE takes the output of \JE of the current frame together with those of past $N$ frames as input tokens.  \TE has 6 layers of multi-head attention, and the output token corresponding to the current frame at the last layer is used as the output of \TE.

\paragraph{MLP head}

The output of \TE is transformed by an MLP into the pointing probability $p$ and the pointing direction $\bm \nu$.  The pointing probability $p$ is implemented as binary classification and the MLP outputs a 2-dimensional vector normalized by the sigmoid function. The pointing direction $\bm \nu$ is first regressed as a 3-dimensional vector of arbitrary norm, and then normalized to be a unit vector.

\paragraph{Training}

We train DeePoint using DP Dataset in a supervised manner, by measuring the cross entropy of $p$ and the angular error of $\bm \nu$ between their ground truths.  The weighting parameter to balance these two terms is determined empirically. During training, we randomly sampled frames so that pointing and non-pointing frames appear evenly. 

\section{Experimental Results}

\def\OursA{\textit{{DP}}\xspace}
\def\OursB{\textit{{DP-B}}\xspace}
\def\OursC{\textit{{DP-BI}}\xspace}
\def\SplitT{\textit{{Split-T}}\xspace}
\def\SplitS{\textit{{Split-S}}\xspace}
\def\SplitP{\textit{{Split-P}}\xspace}

\if11
\begin{figure*}[t]
    \centering
    \def\Pa#1{\includegraphics[width=0.22\linewidth,bb={115 130 729 393},clip]{#1}}
    \def\Pb#1{\includegraphics[width=0.22\linewidth,bb={0   248 640 522},clip]{#1}}
    \def\Pc#1{\includegraphics[width=0.22\linewidth,bb={320 166 960 440},clip]{#1}}
    \def\Pd#1{\includegraphics[width=0.22\linewidth,bb={146 178 786 452},clip]{#1}}
    \def\Pe#1{\includegraphics[width=0.22\linewidth,bb={178 172 739 412},clip]{#1}}
    \def\Pf#1{\includegraphics[width=0.22\linewidth,bb={8   267 808 610},clip]{#1}}
    \def\Pg#1{\includegraphics[width=0.22\linewidth,bb={421 213 960 444},clip]{#1}}
    \def\Ph#1{\includegraphics[width=0.22\linewidth,bb={82  171 722 445},clip]{#1}}
\begin{tikzpicture}[x=1.02mm,y=1mm]
    \newcount\vX
    \newcount\vY
    \newcount\vdX
    \newcount\vdY
    \newcount\vXi
    \newcount\vYi
    \vdX = 41
    \vdY = -16
    \vXi = -24
    \vYi = -100

    \vX = \vXi
    \node[inner sep=0pt, font=\scriptsize] (a) at (\vX,\vY) {$t=0$};
    \advance\vY by \vdY
    \node[inner sep=0pt, font=\scriptsize] (a) at (\vX,\vY) {$t=3$};
    \advance\vY by \vdY
    \node[inner sep=0pt, font=\scriptsize] (a) at (\vX,\vY) {$t=6$};
    \advance\vY by \vdY
    \node[inner sep=0pt, font=\scriptsize] (a) at (\vX,\vY) {$t=9$};
    \advance\vY by \vdY
    \node[inner sep=0pt, font=\scriptsize] (a) at (\vX,\vY) {$t=12$};
    \advance\vY by \vdY
    \node[inner sep=0pt, font=\scriptsize] (a) at (\vX,\vY) {$t=15$};
    \advance\vY by \vdY

    \vX = 0
    \vY = 0
    \node at (\vX,\vY) {\Pa{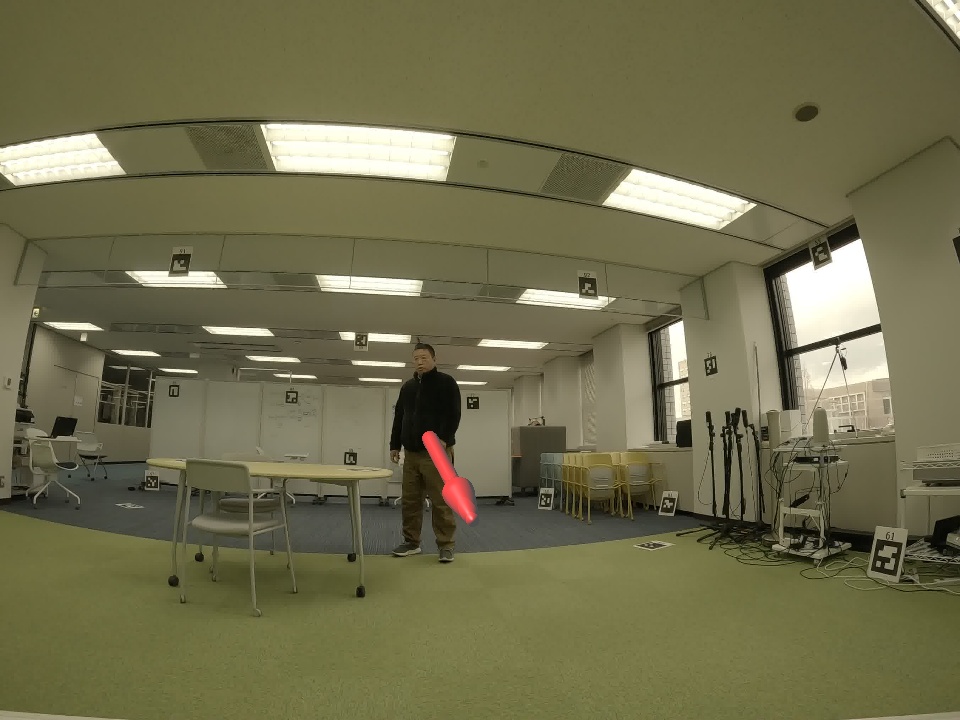}};
    \advance\vY by \vdY
    \node at (\vX,\vY) {\Pa{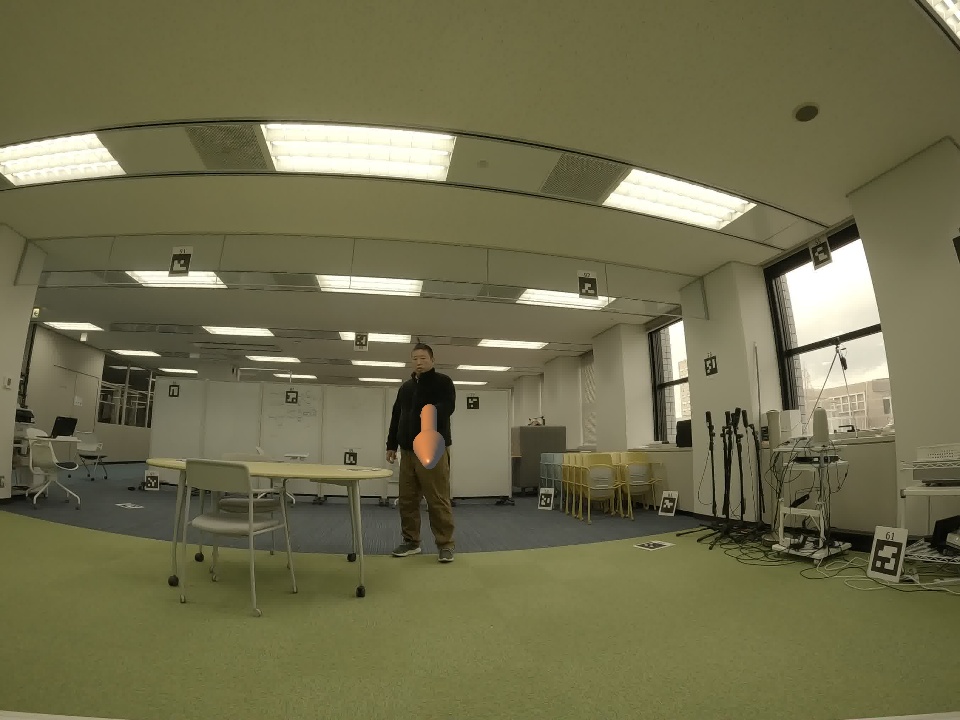}};
    \advance\vY by \vdY
    \node at (\vX,\vY) {\Pa{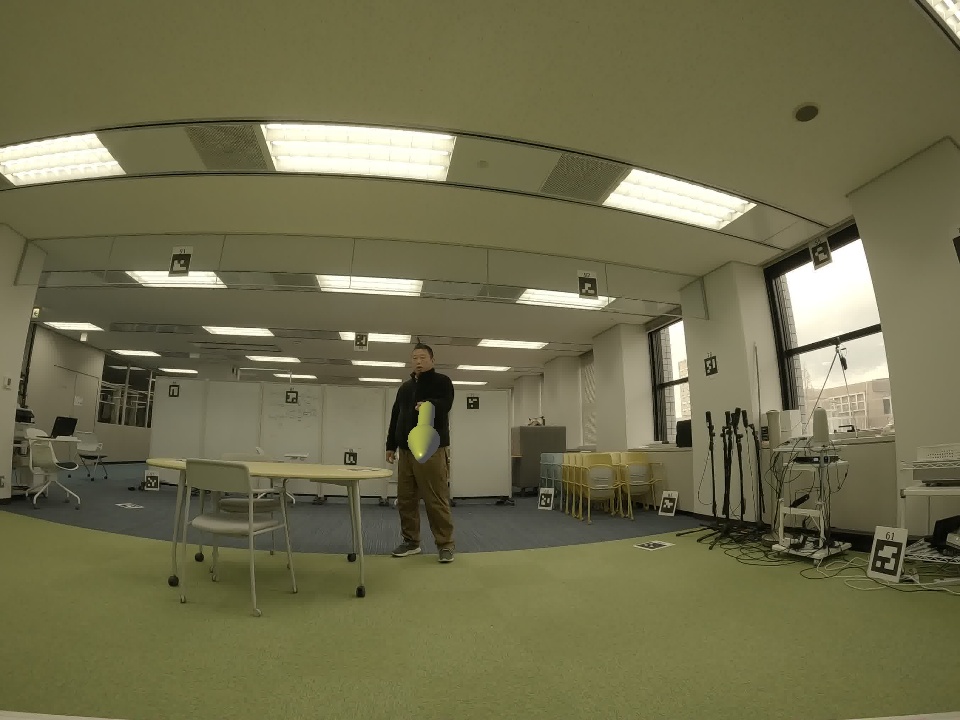}};
    \advance\vY by \vdY
    \node at (\vX,\vY) {\Pa{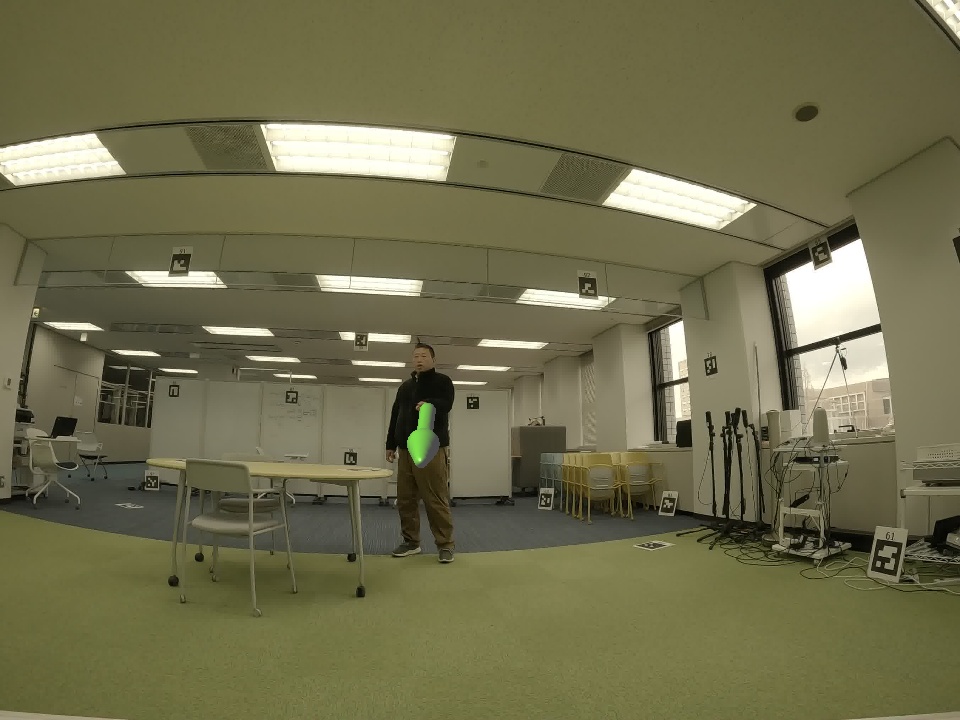}};
    \advance\vY by \vdY
    \node at (\vX,\vY) {\Pa{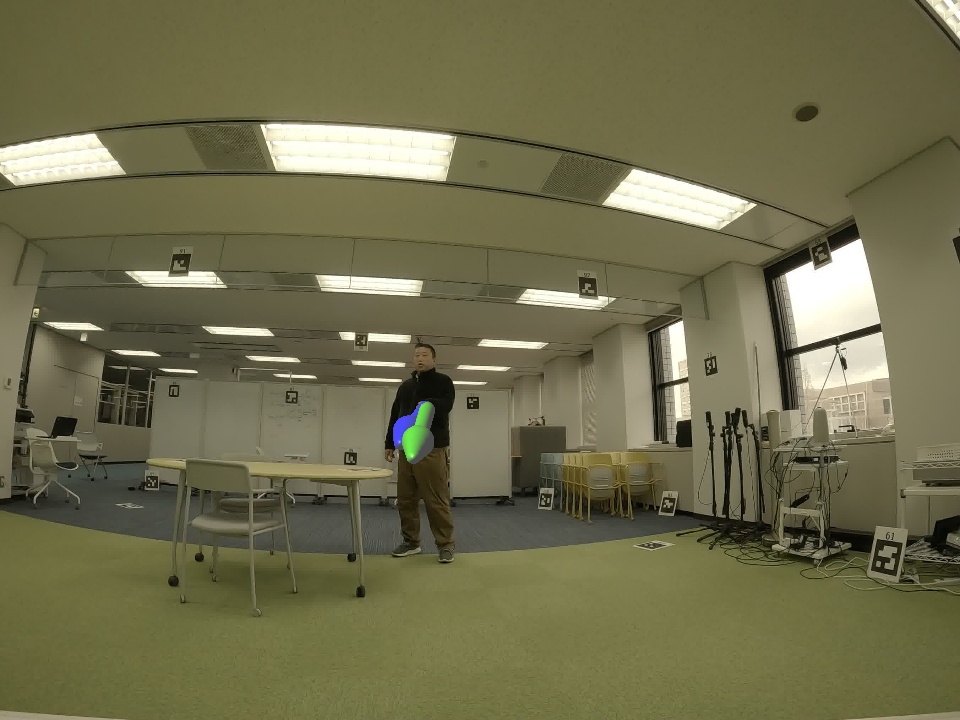}};
    \advance\vY by \vdY
    \node at (\vX,\vY) {\Pa{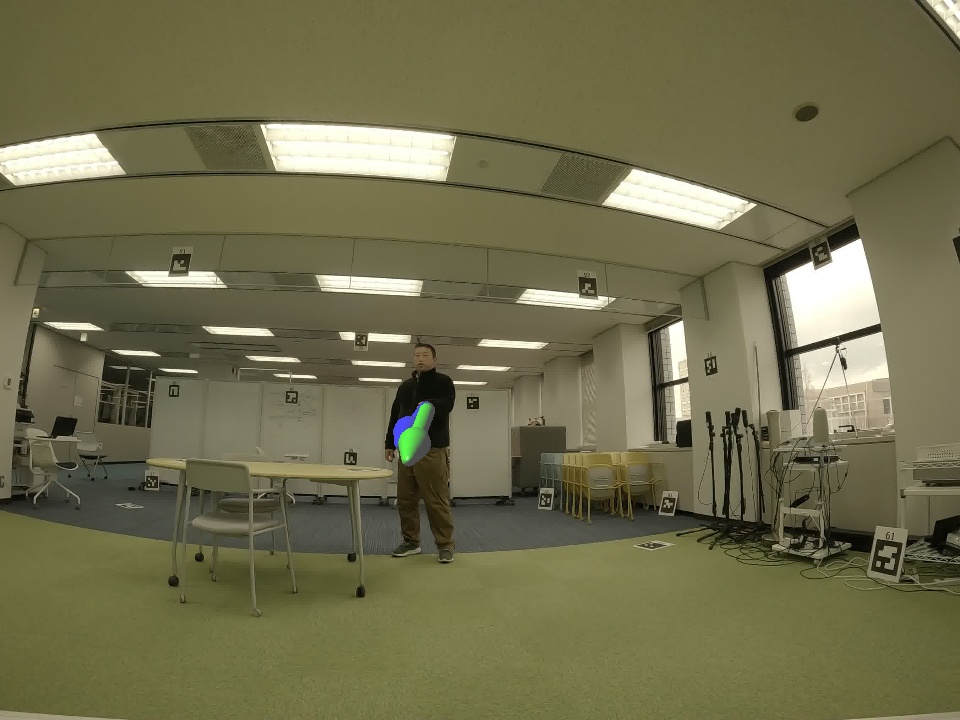}};

    \advance\vX by \vdX
    \vY = 0
    \node at (\vX,\vY) {\Pb{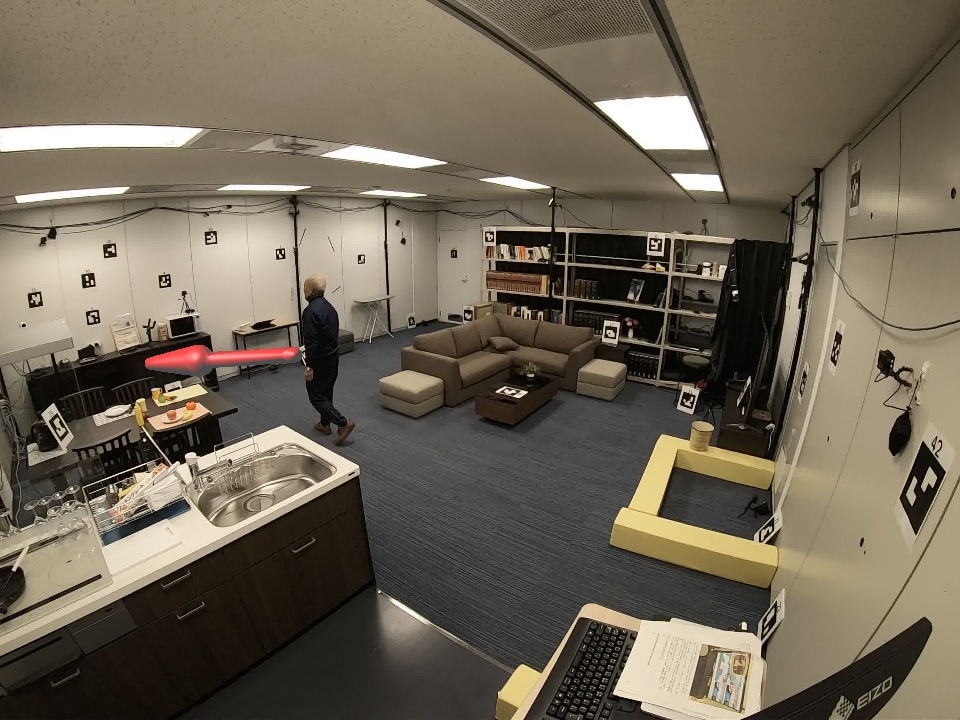}};
    \advance\vY by \vdY
    \node at (\vX,\vY) {\Pb{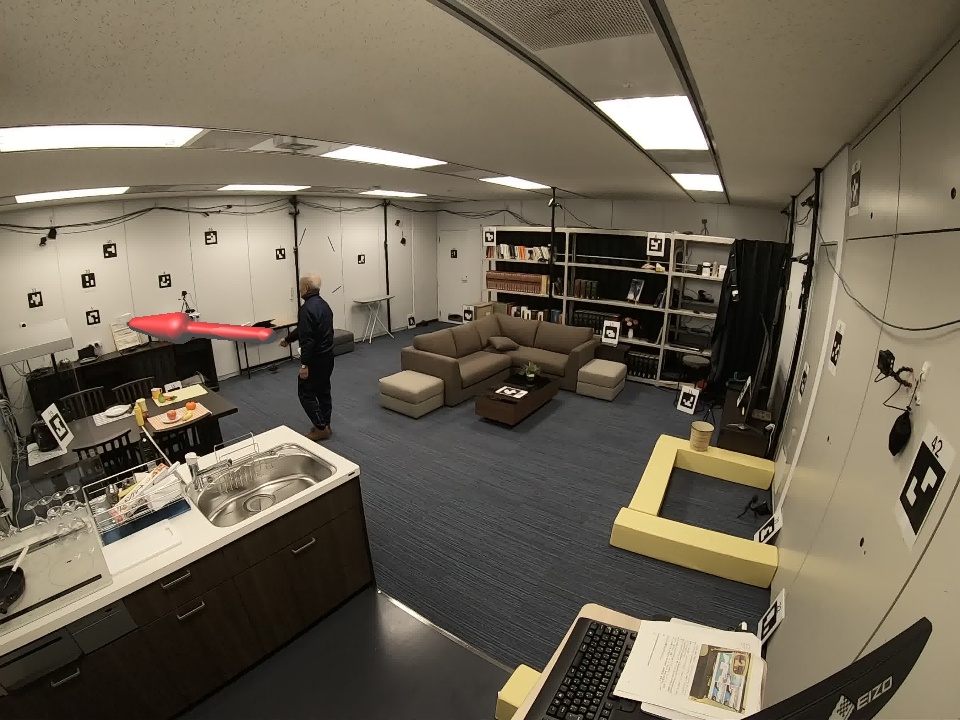}};
    \advance\vY by \vdY
    \node at (\vX,\vY) {\Pb{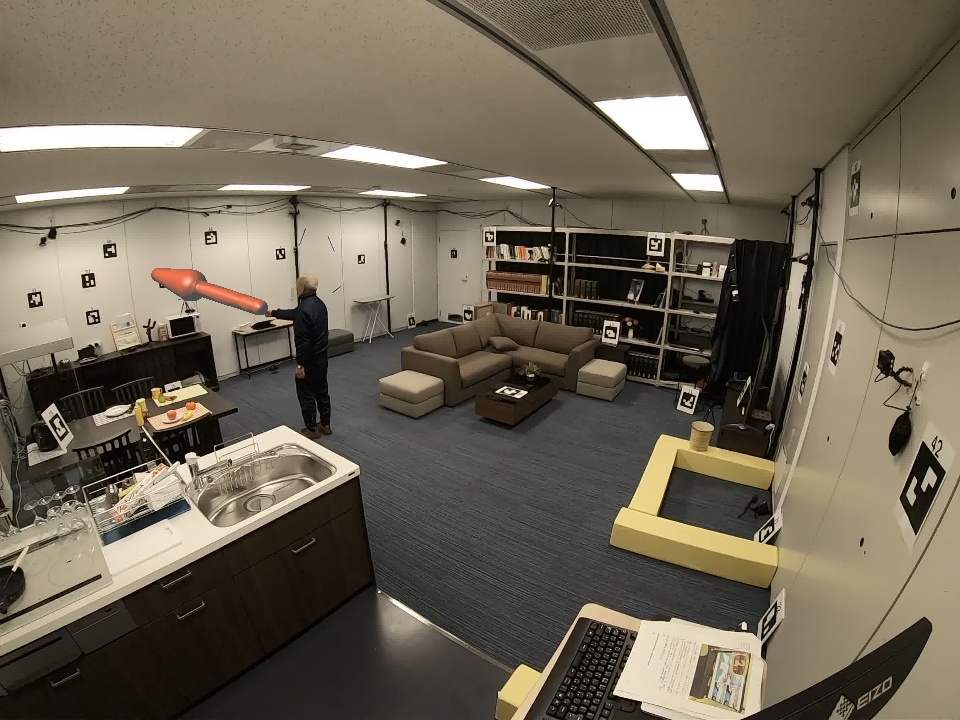}};
    \advance\vY by \vdY
    \node at (\vX,\vY) {\Pb{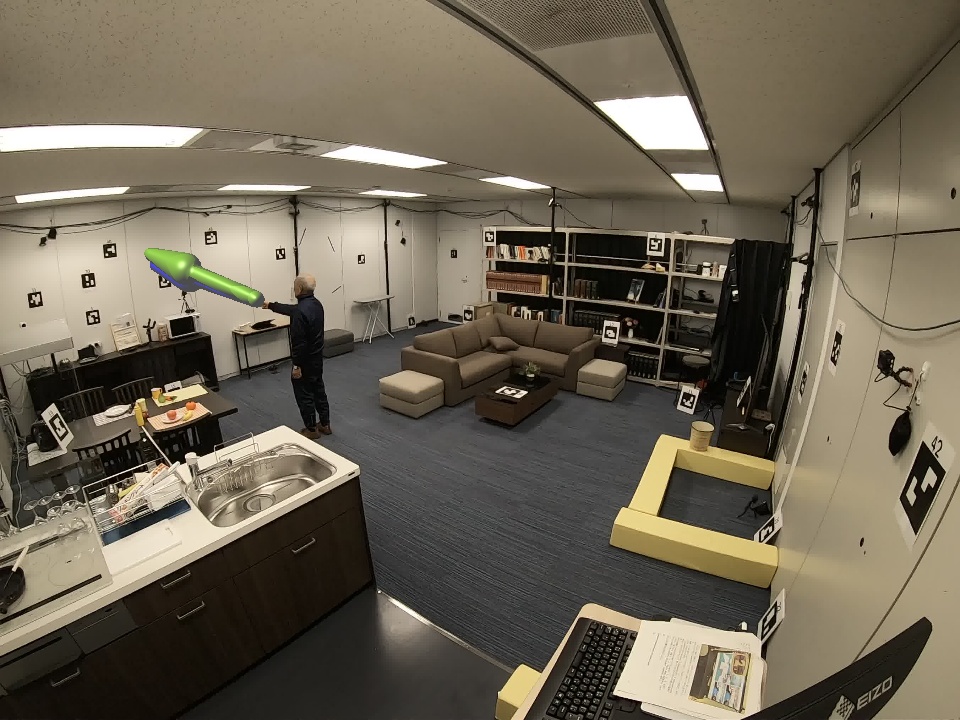}};
    \advance\vY by \vdY
    \node at (\vX,\vY) {\Pb{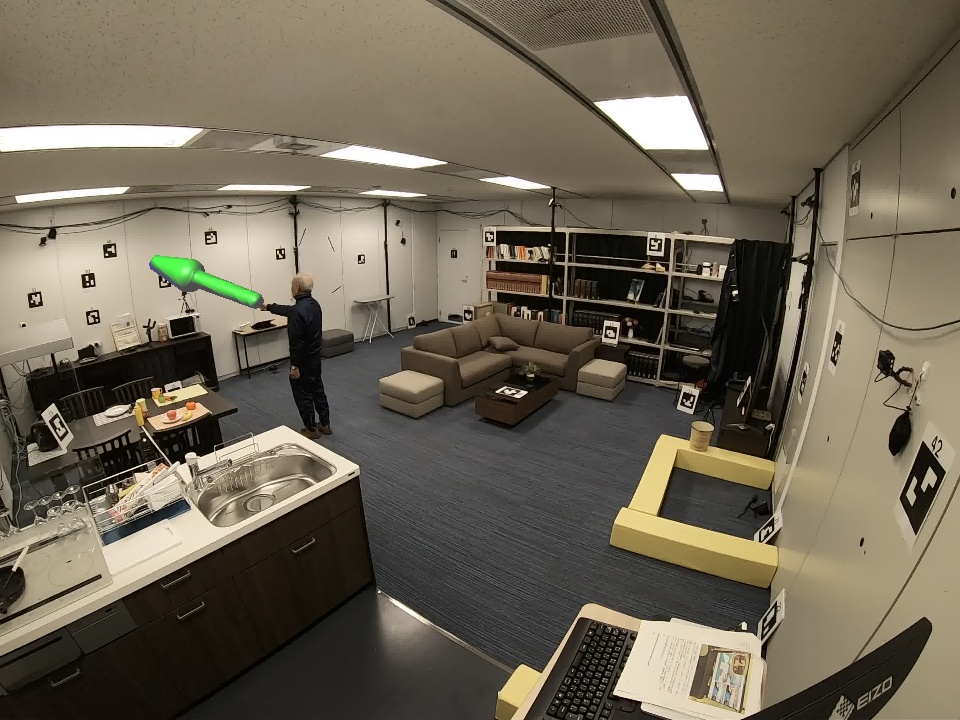}};
    \advance\vY by \vdY
    \node at (\vX,\vY) {\Pb{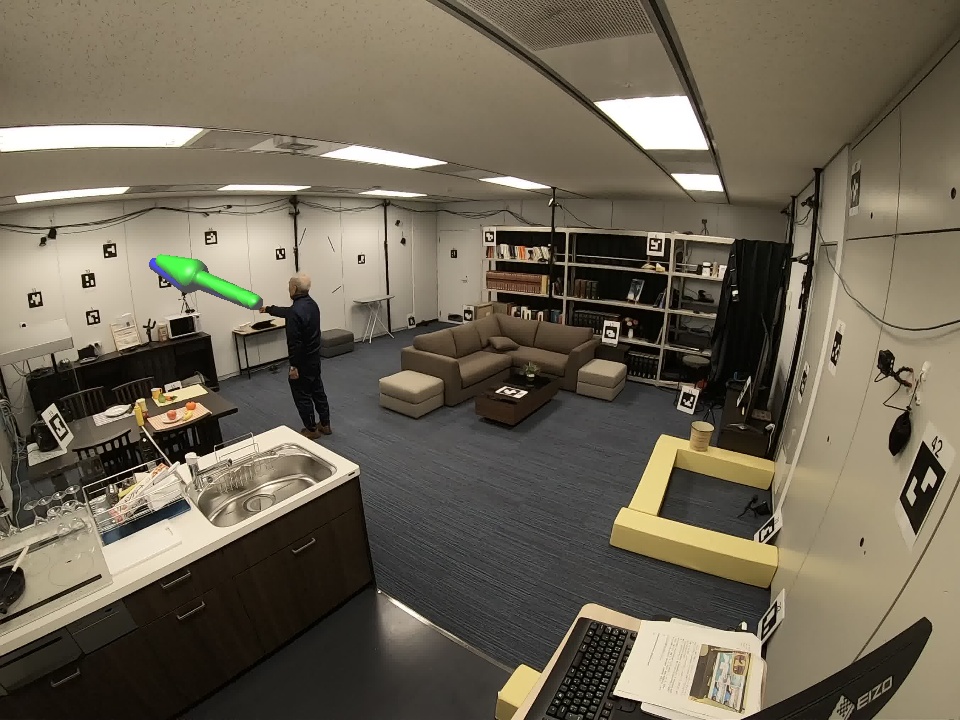}};

    \advance\vX by \vdX
    \vY = 0
    \node at (\vX,\vY) {\Pc{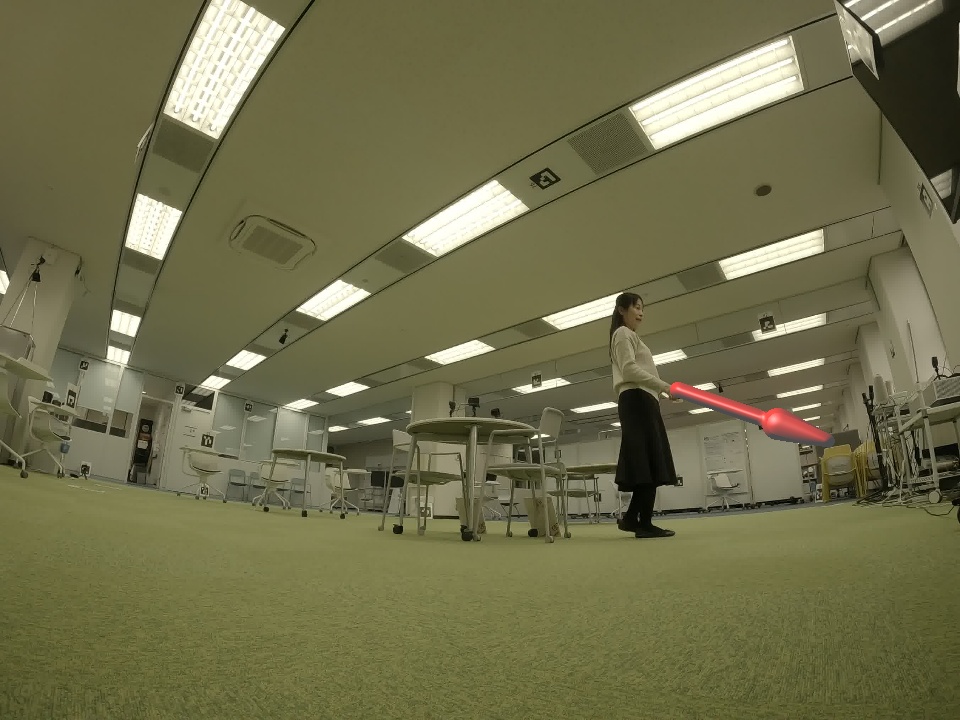}};
    \advance\vY by \vdY
    \node at (\vX,\vY) {\Pc{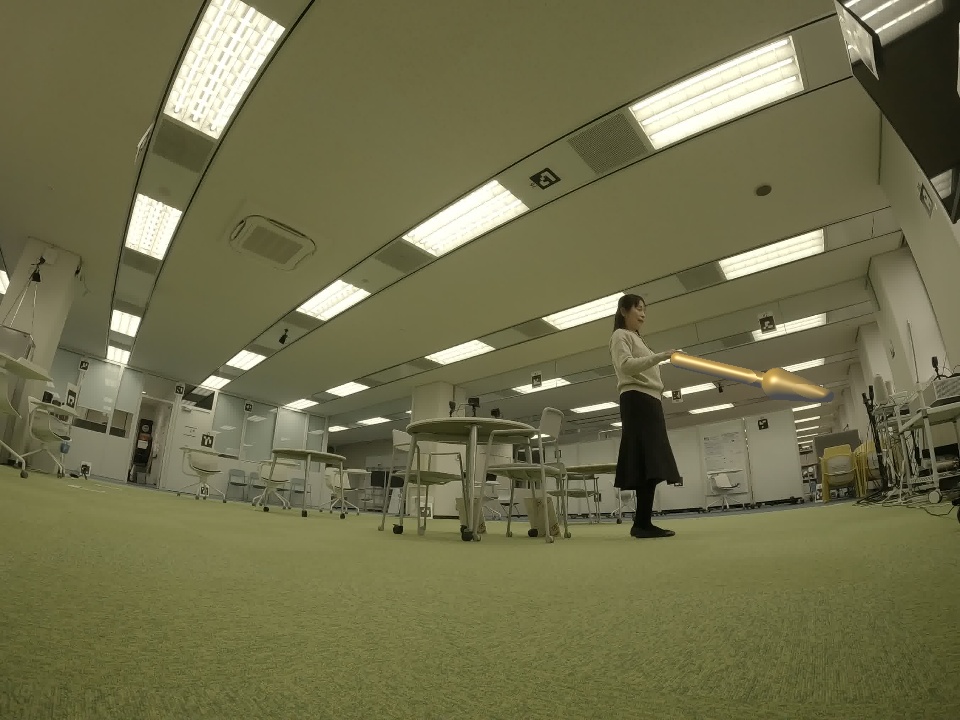}};
    \advance\vY by \vdY
    \node at (\vX,\vY) {\Pc{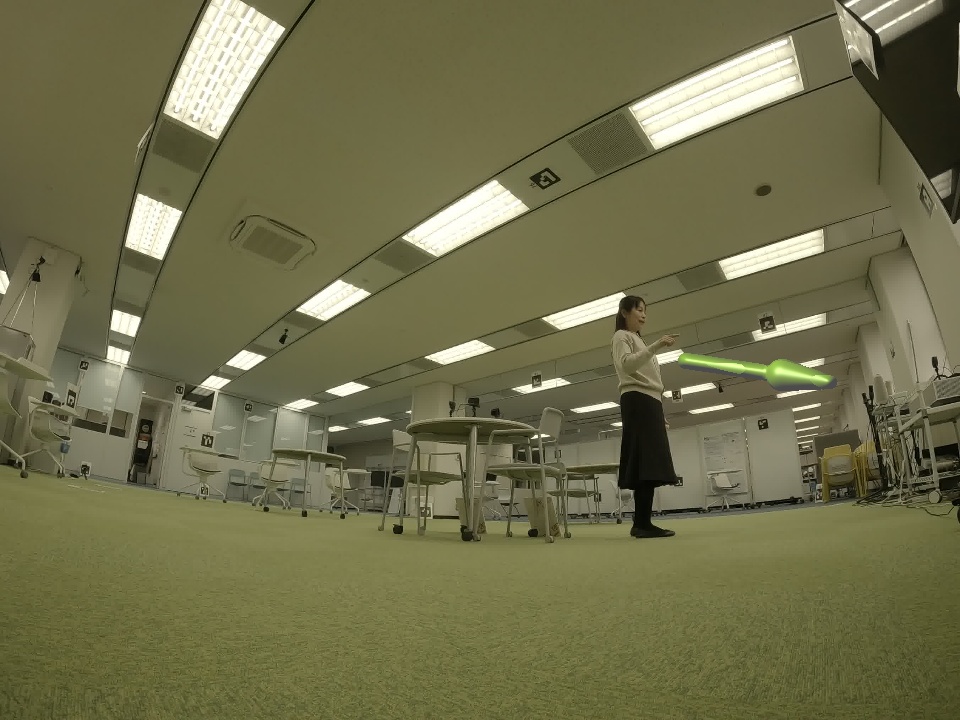}};
    \advance\vY by \vdY
    \node at (\vX,\vY) {\Pc{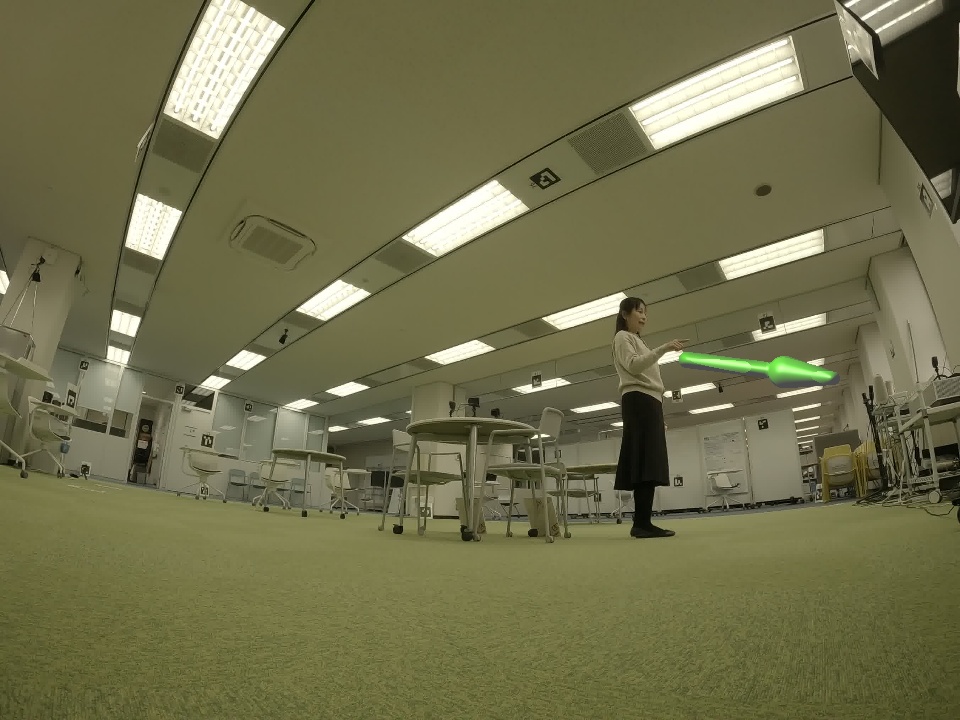}};
    \advance\vY by \vdY
    \node at (\vX,\vY) {\Pc{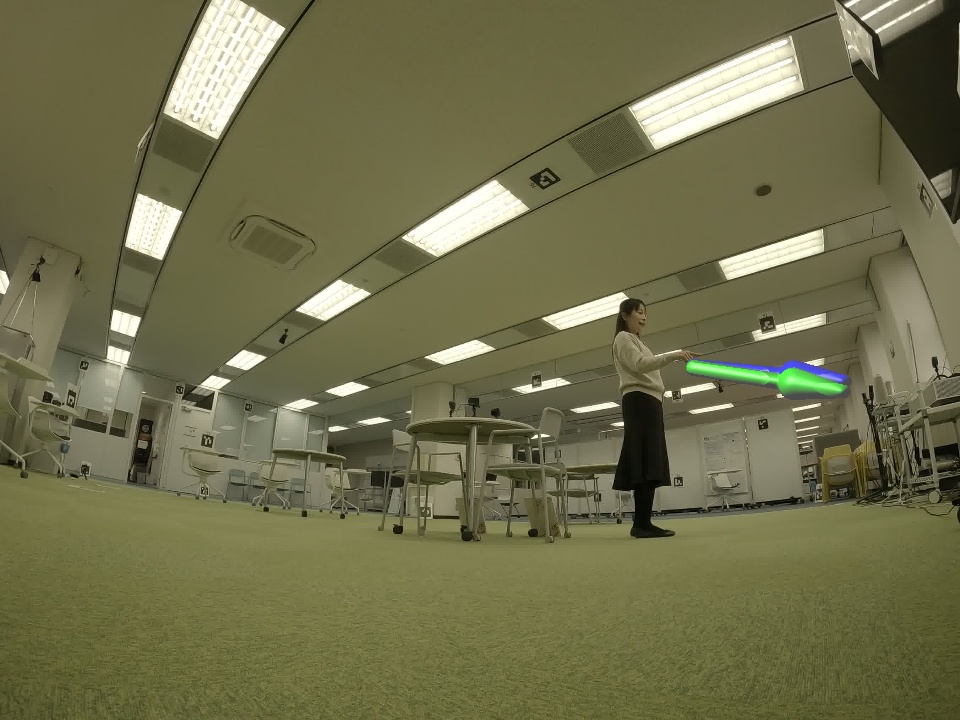}};
    \advance\vY by \vdY
    \node at (\vX,\vY) {\Pc{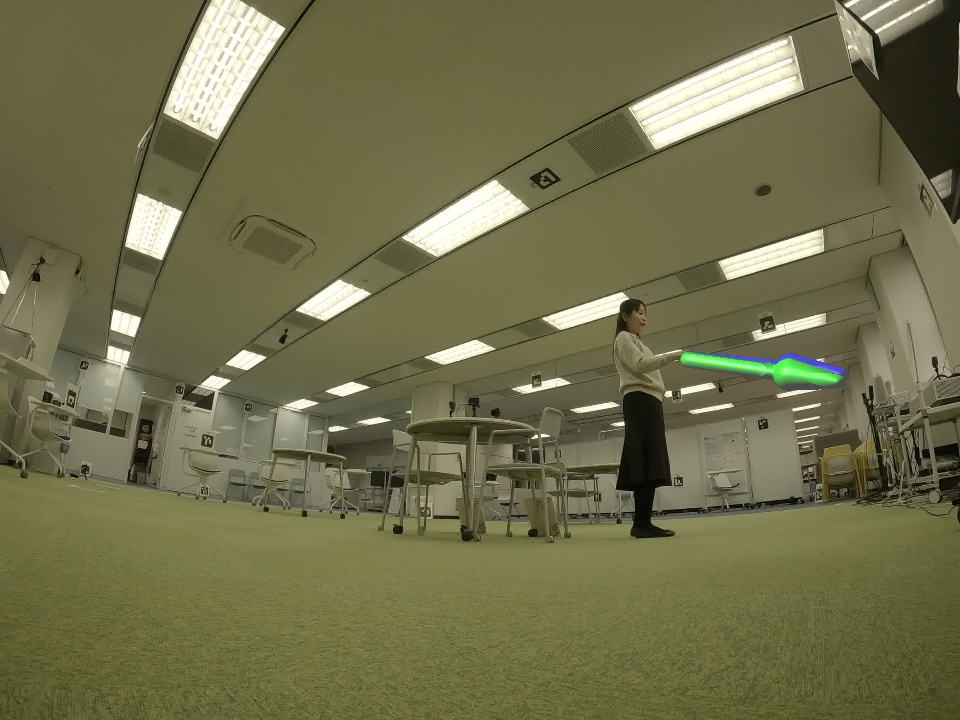}};

    \advance\vX by \vdX
    \vY = 0
    \node at (\vX,\vY) {\Pd{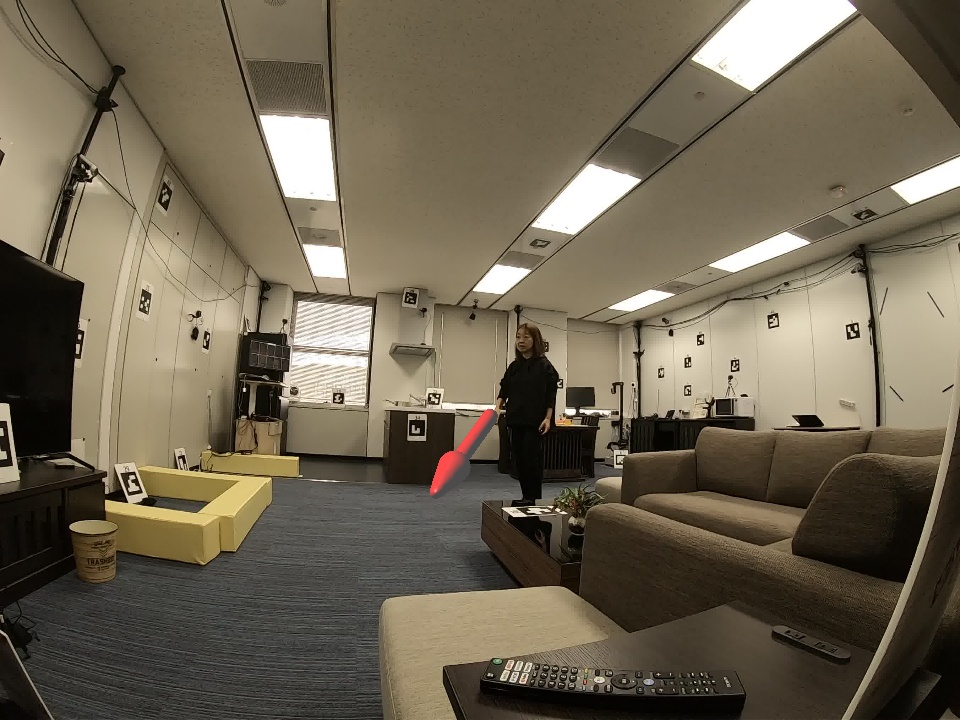}};
    \advance\vY by \vdY
    \node at (\vX,\vY) {\Pd{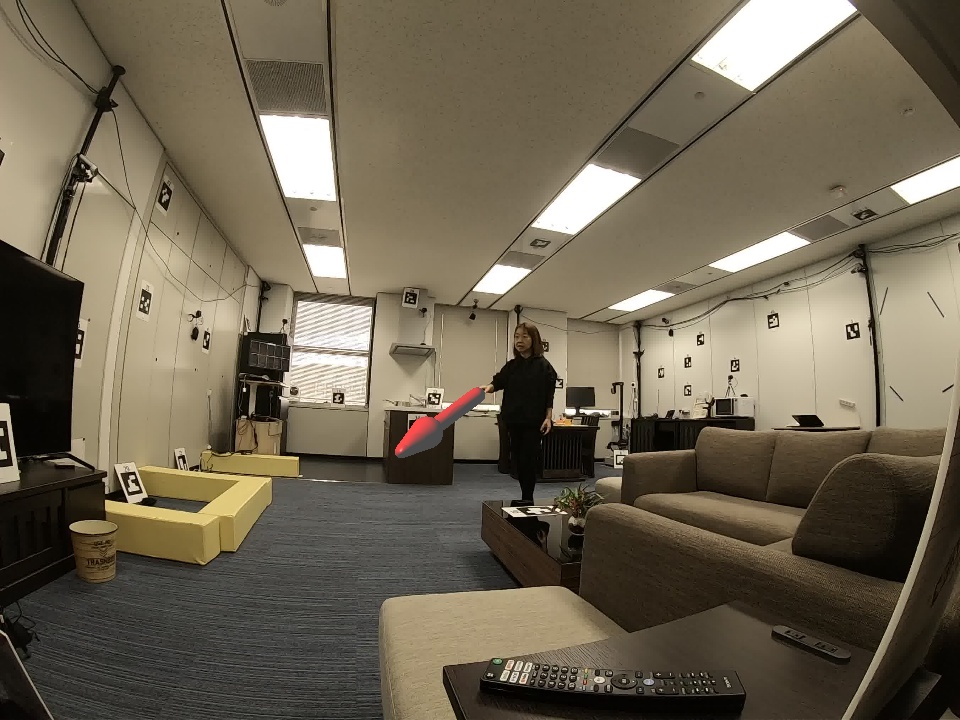}};
    \advance\vY by \vdY
    \node at (\vX,\vY) {\Pd{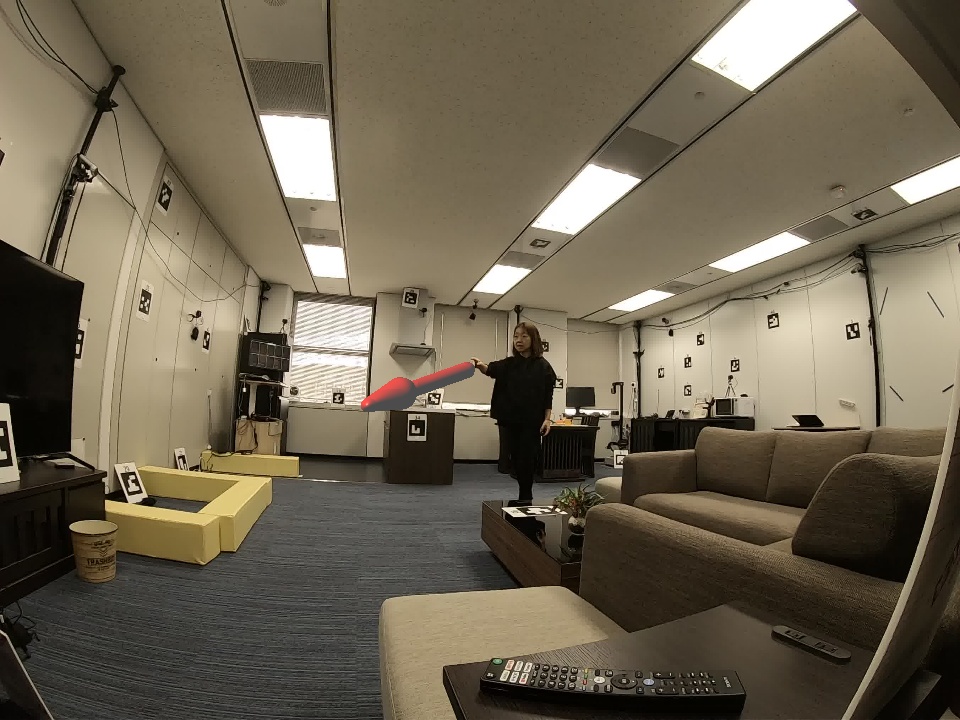}};
    \advance\vY by \vdY
    \node at (\vX,\vY) {\Pd{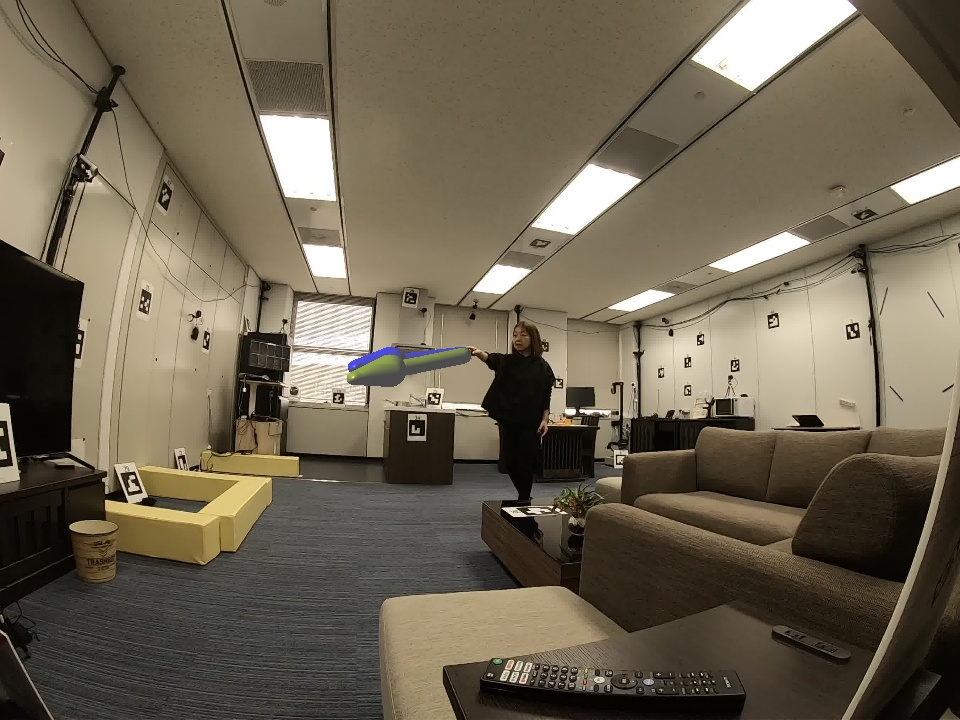}};
    \advance\vY by \vdY
    \node at (\vX,\vY) {\Pd{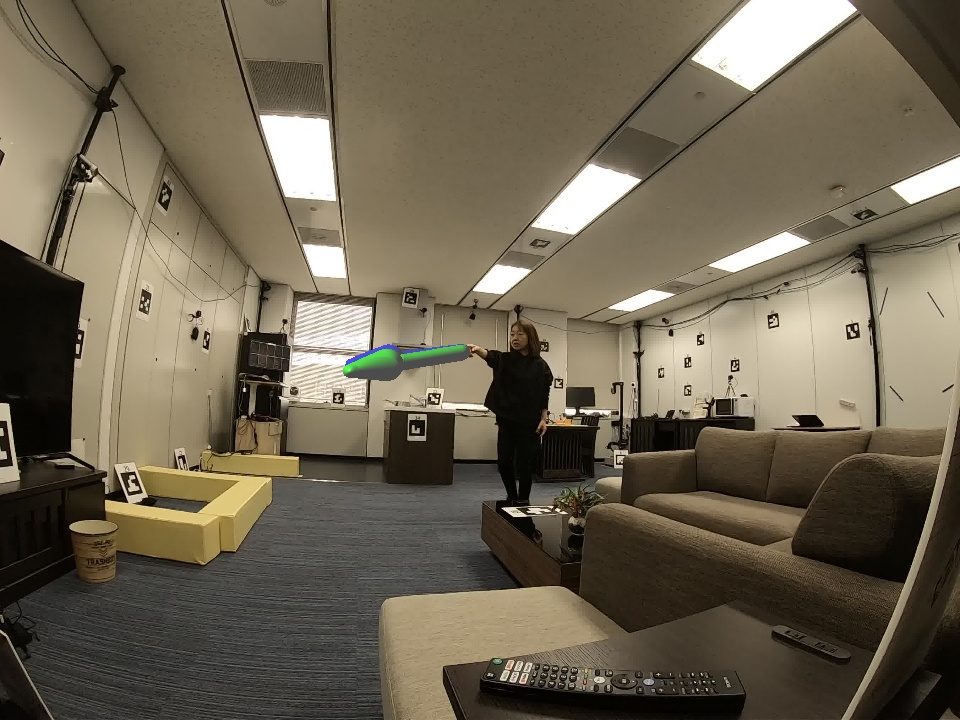}};
    \advance\vY by \vdY
    \node at (\vX,\vY) {\Pd{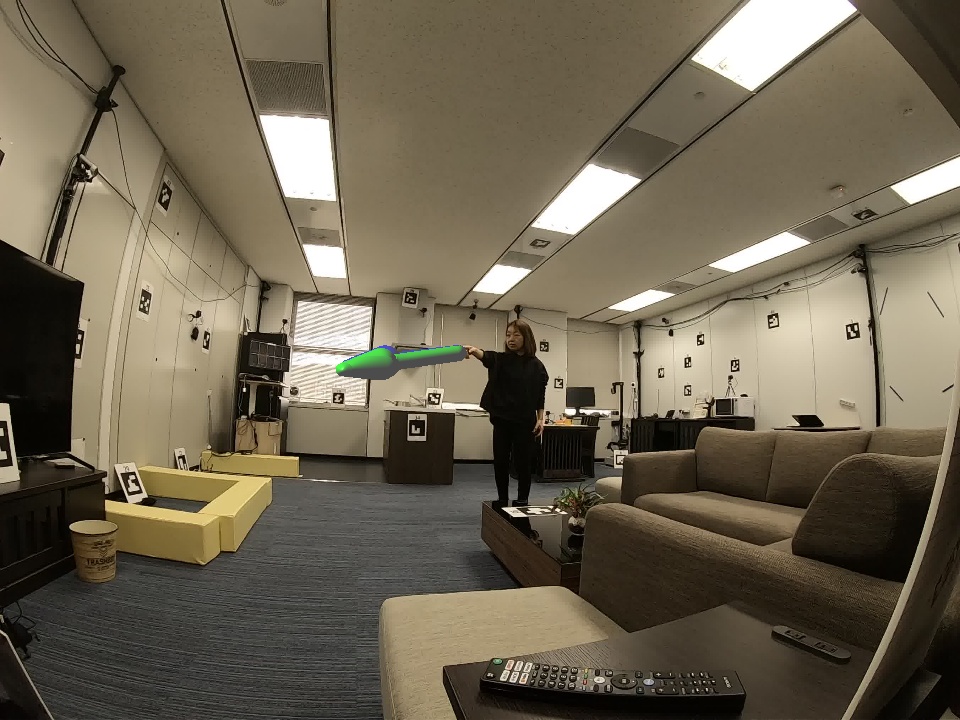}};

    \vX = \vXi
    \vY = \vYi
    \node[inner sep=0pt, font=\scriptsize] (a) at (\vX,\vY) {$t=0$};
    \advance\vY by \vdY
    \node[inner sep=0pt, font=\scriptsize] (a) at (\vX,\vY) {$t=3$};
    \advance\vY by \vdY
    \node[inner sep=0pt, font=\scriptsize] (a) at (\vX,\vY) {$t=6$};
    \advance\vY by \vdY
    \node[inner sep=0pt, font=\scriptsize] (a) at (\vX,\vY) {$t=9$};
    \advance\vY by \vdY
    \node[inner sep=0pt, font=\scriptsize] (a) at (\vX,\vY) {$t=12$};
    \advance\vY by \vdY
    \node[inner sep=0pt, font=\scriptsize] (a) at (\vX,\vY) {$t=15$};
    \advance\vY by \vdY

    \vX = 0
    \vY = \vYi
    \node at (\vX,\vY) {\Pe{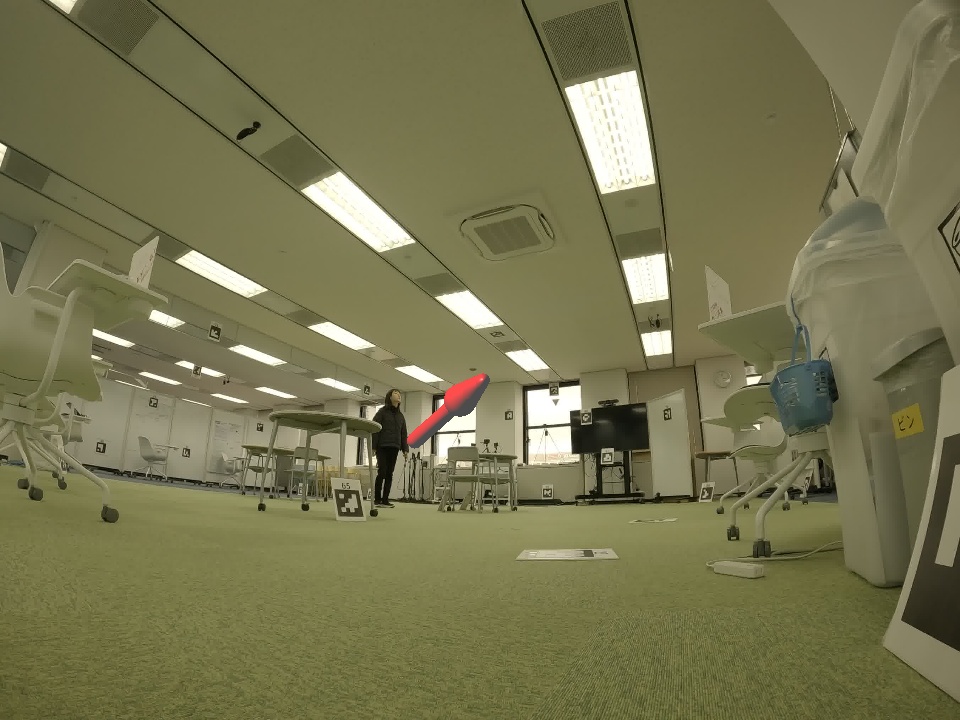}};
    \advance\vY by \vdY
    \node at (\vX,\vY) {\Pe{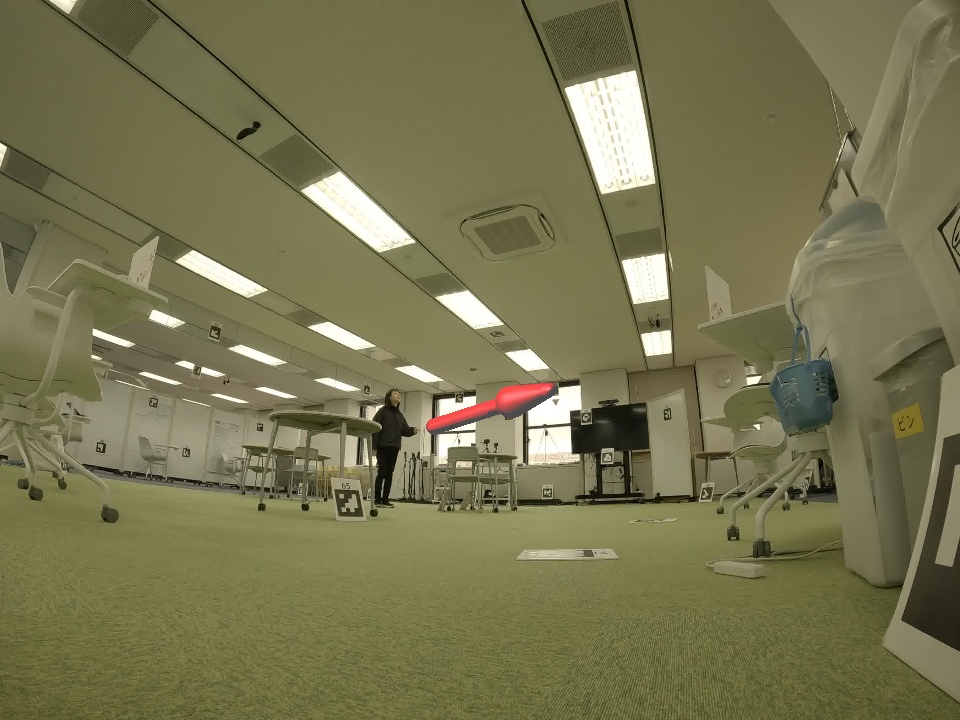}};
    \advance\vY by \vdY
    \node at (\vX,\vY) {\Pe{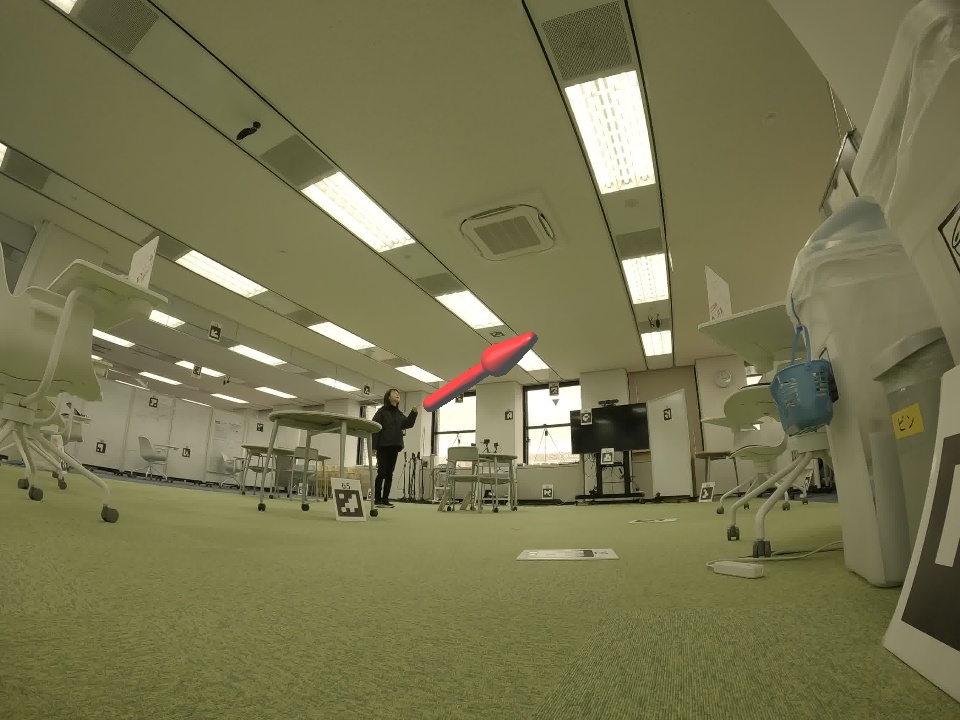}};
    \advance\vY by \vdY
    \node at (\vX,\vY) {\Pe{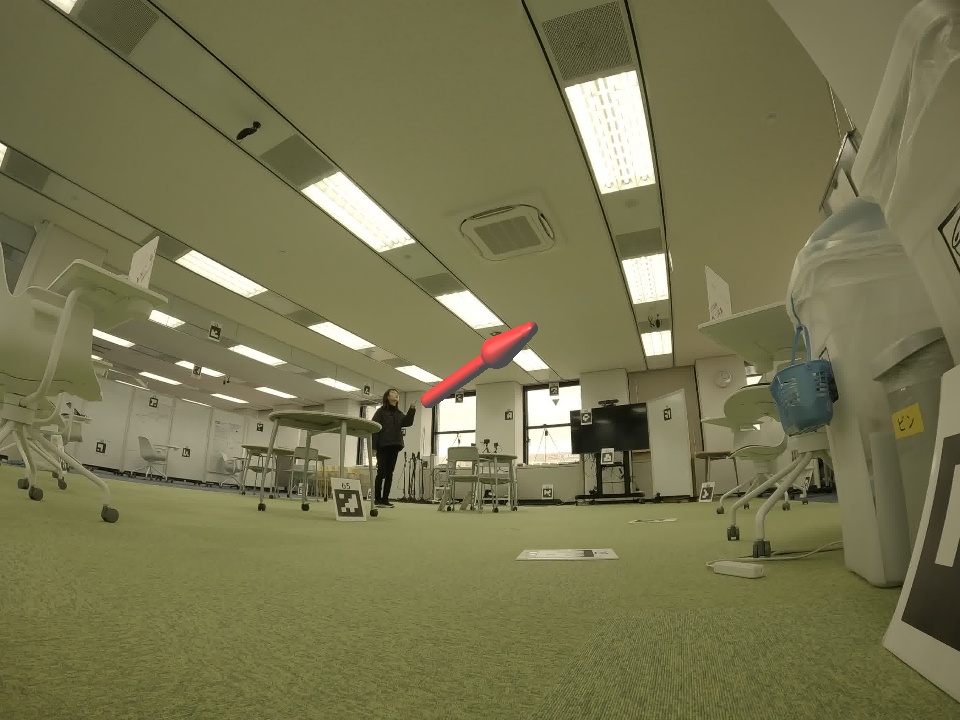}};
    \advance\vY by \vdY
    \node at (\vX,\vY) {\Pe{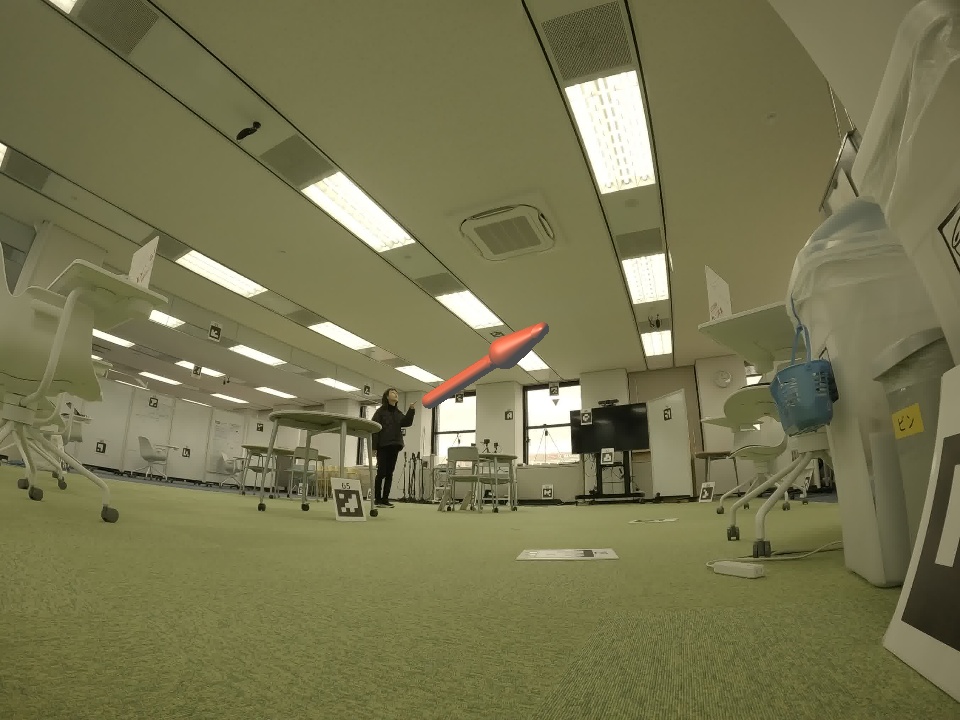}};
    \advance\vY by \vdY
    \node at (\vX,\vY) {\Pe{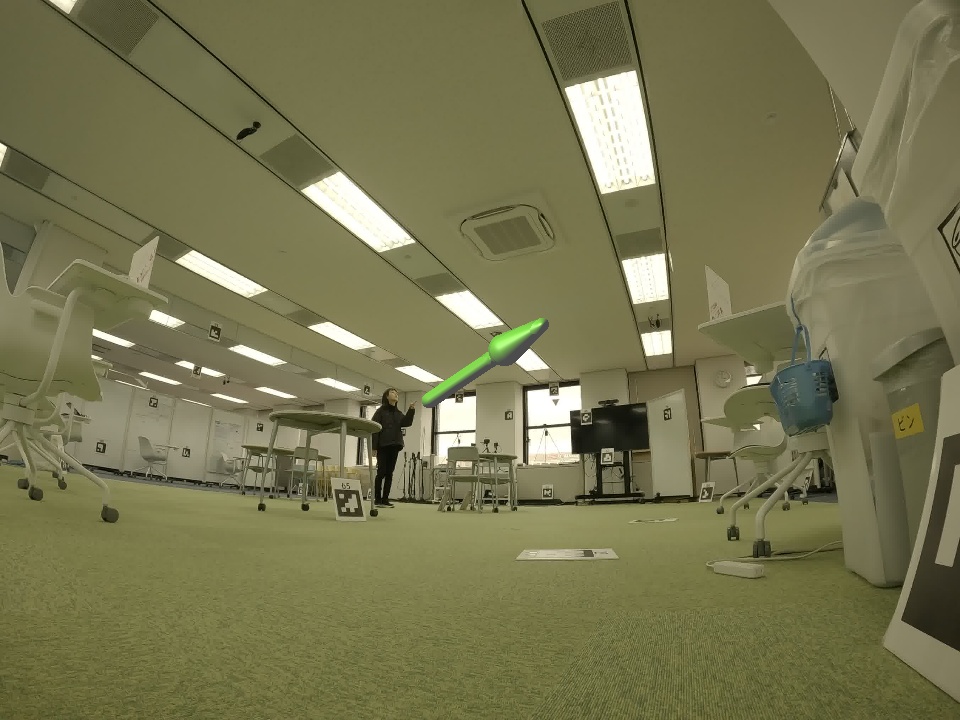}};

    \advance\vX by \vdX
    \vY = \vYi
    \node at (\vX,\vY) {\Pf{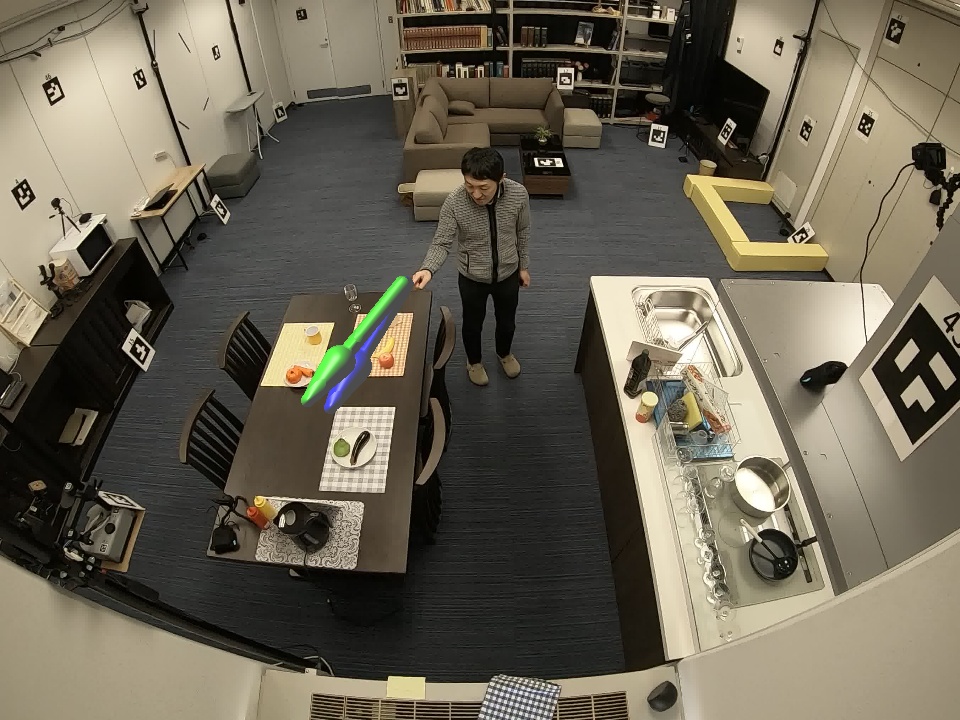}};
    \advance\vY by \vdY
    \node at (\vX,\vY) {\Pf{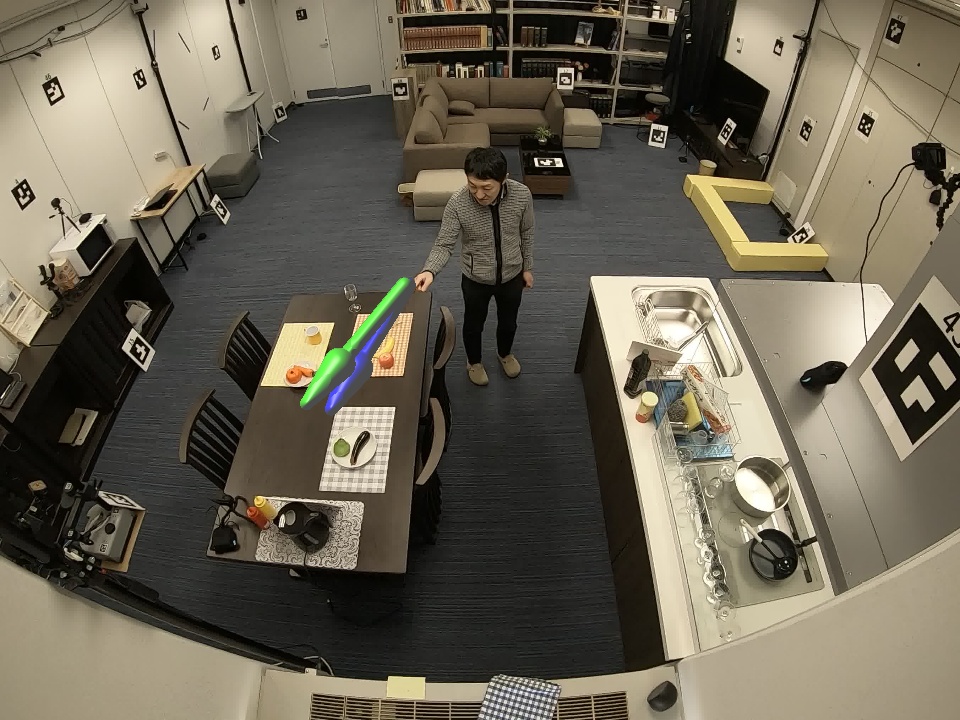}};
    \advance\vY by \vdY
    \node at (\vX,\vY) {\Pf{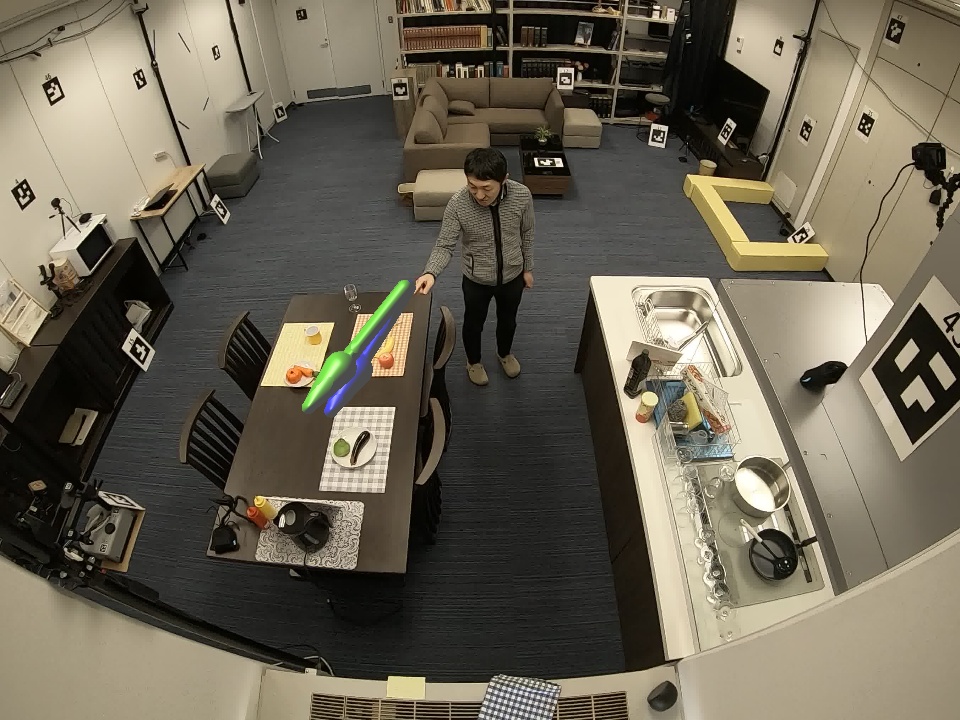}};
    \advance\vY by \vdY
    \node at (\vX,\vY) {\Pf{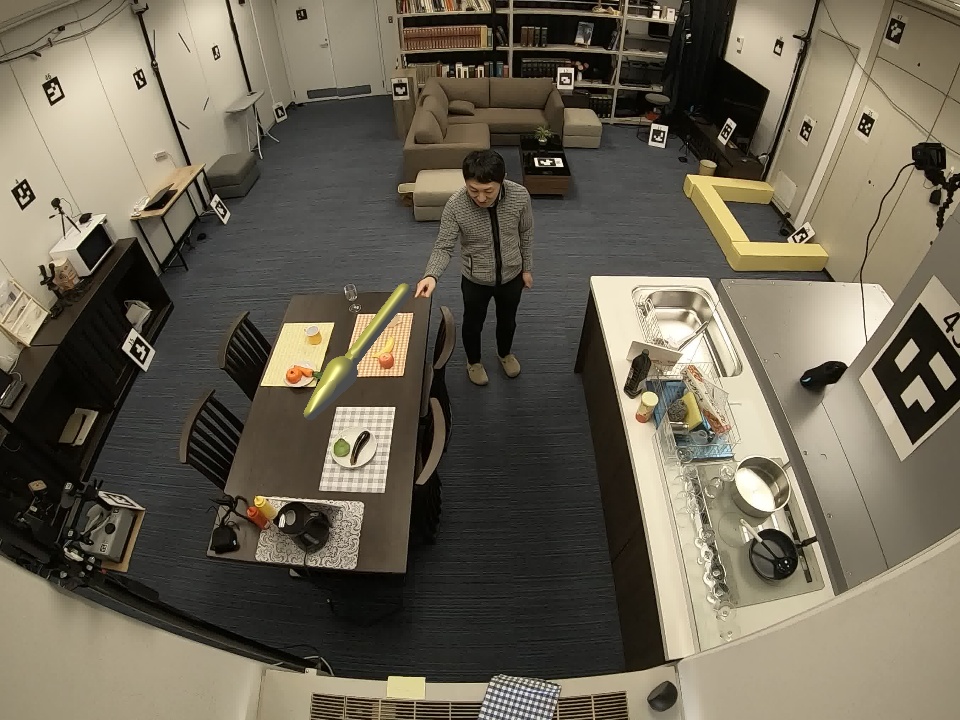}};
    \advance\vY by \vdY
    \node at (\vX,\vY) {\Pf{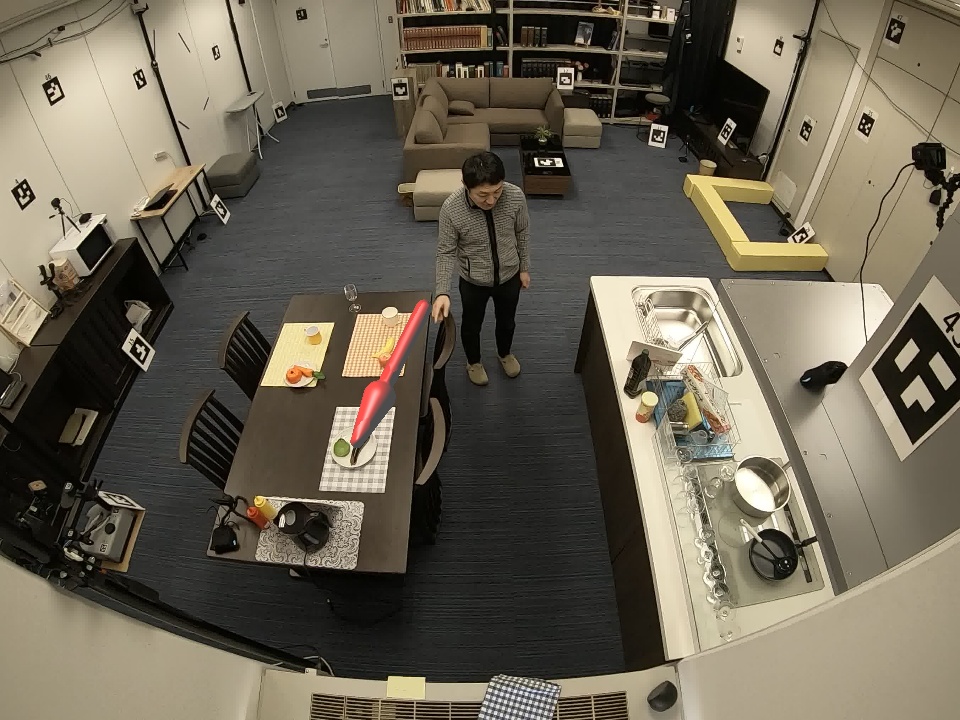}};
    \advance\vY by \vdY
    \node at (\vX,\vY) {\Pf{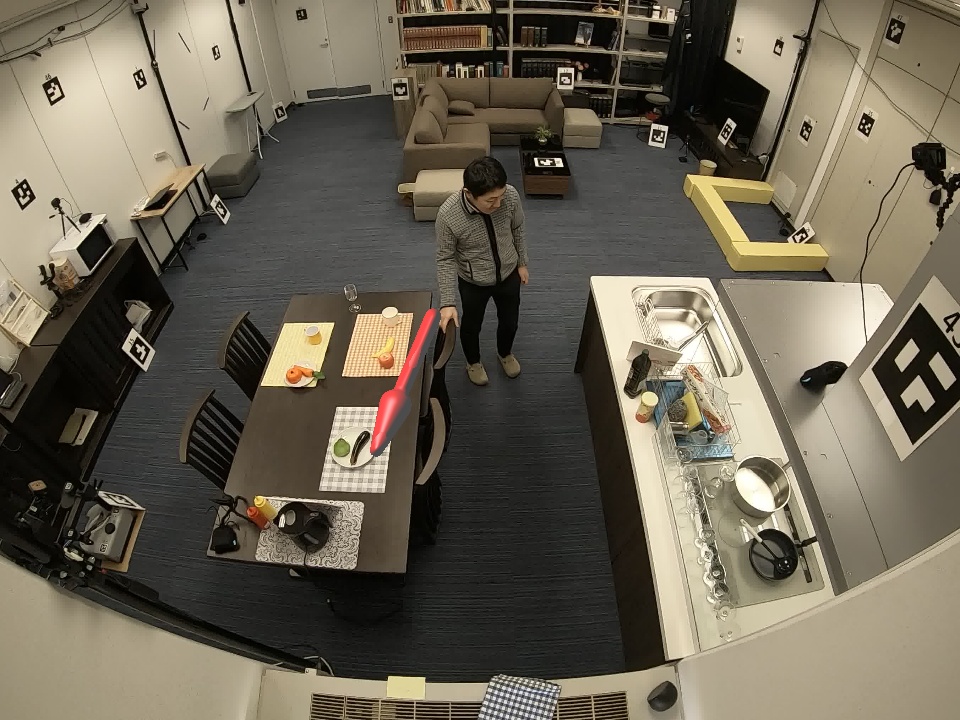}};

    \advance\vX by \vdX
    \vY = \vYi
    \node at (\vX,\vY) {\Pg{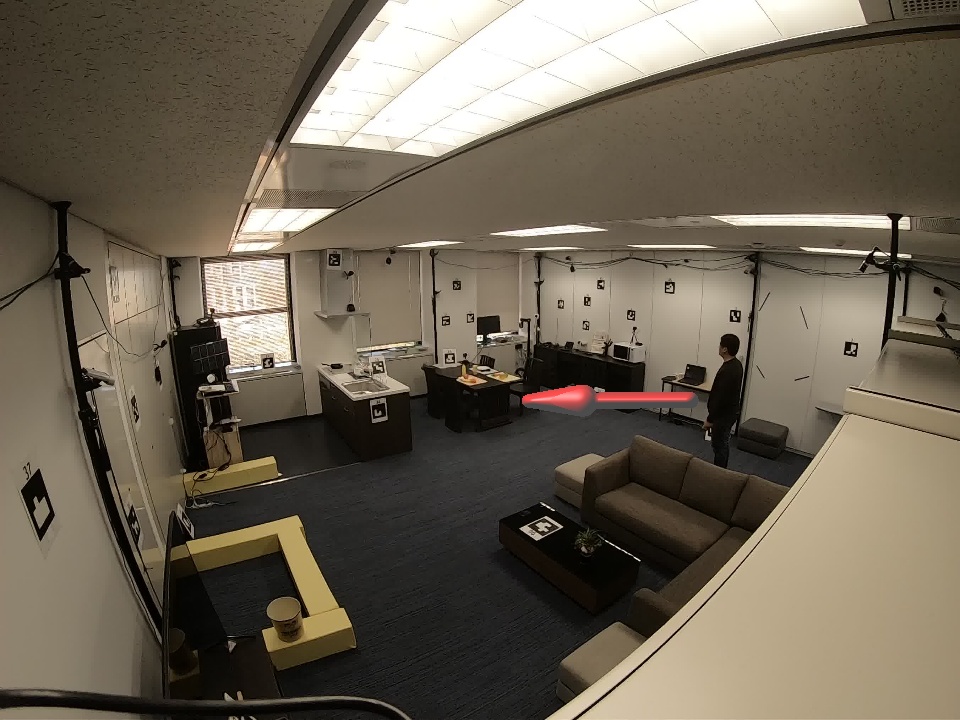}};
    \advance\vY by \vdY
    \node at (\vX,\vY) {\Pg{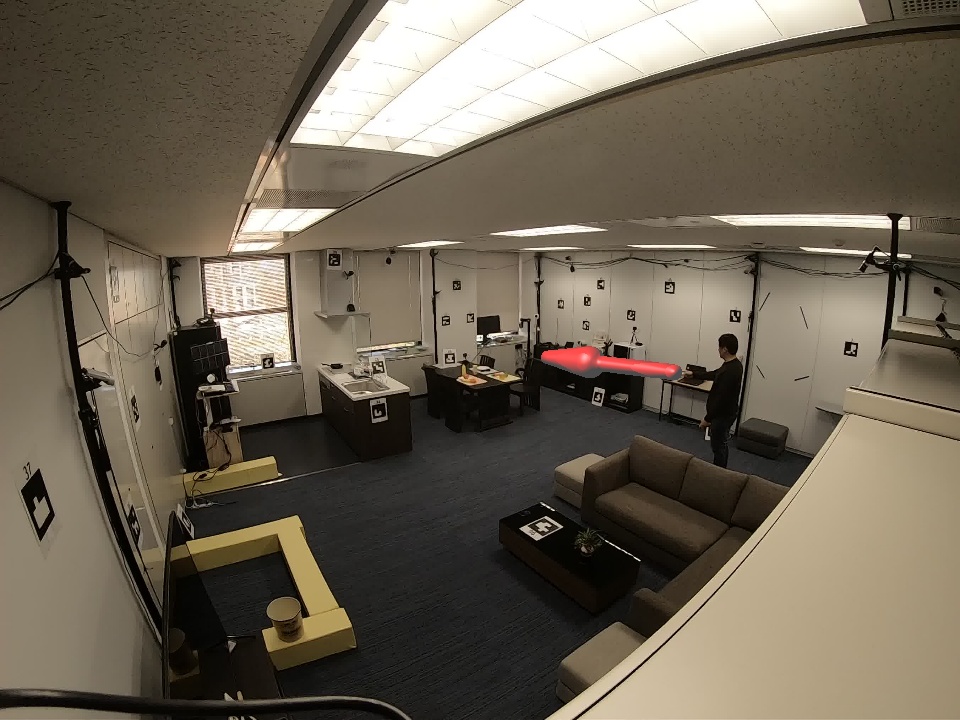}};
    \advance\vY by \vdY
    \node at (\vX,\vY) {\Pg{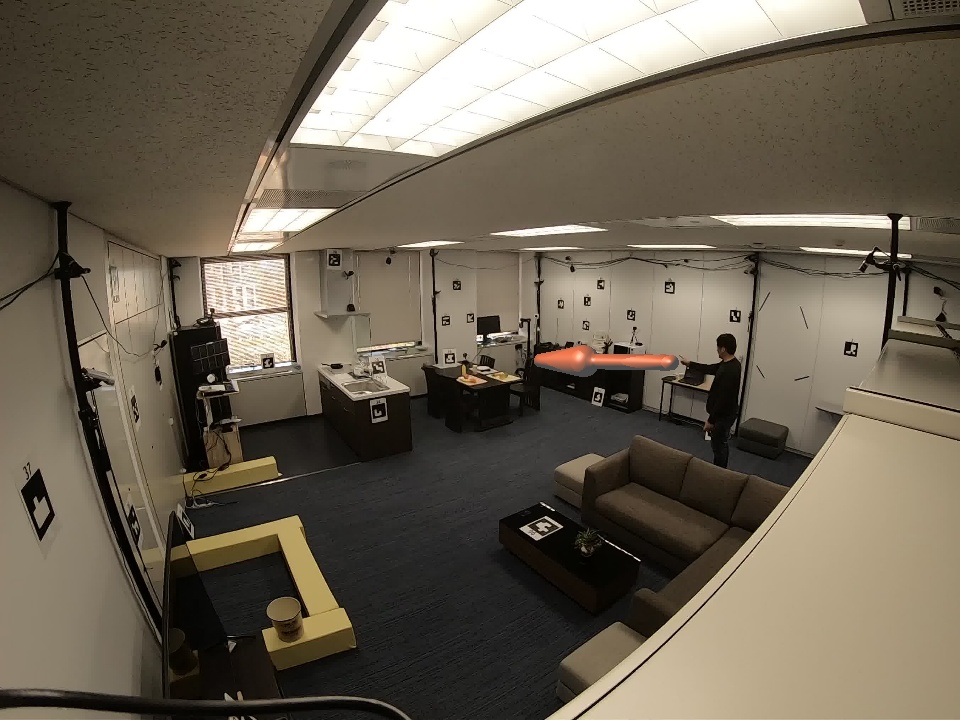}};
    \advance\vY by \vdY
    \node at (\vX,\vY) {\Pg{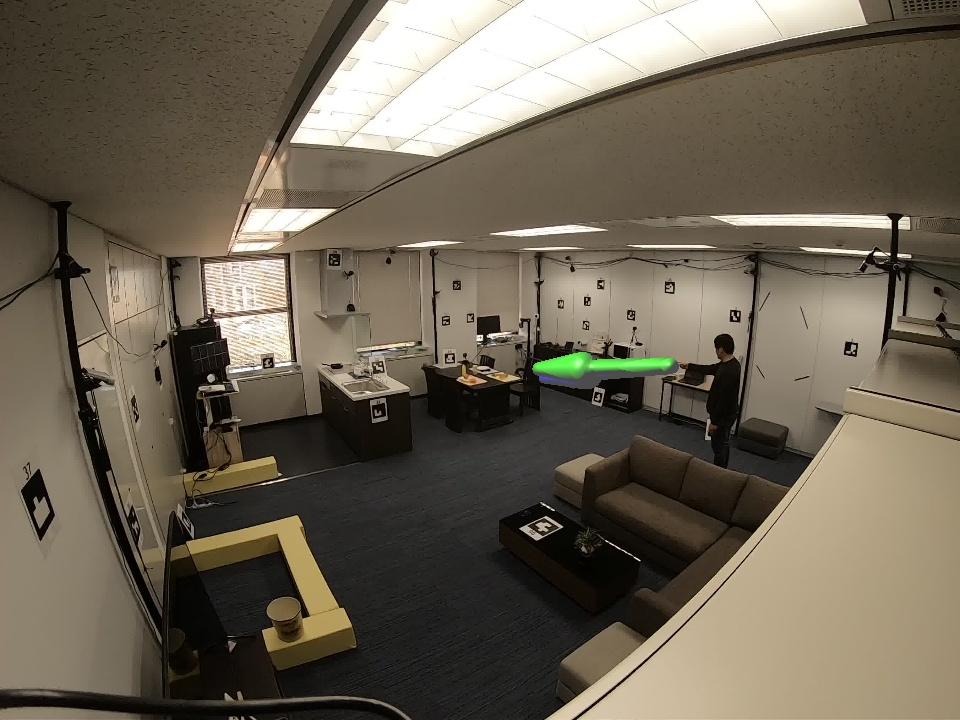}};
    \advance\vY by \vdY
    \node at (\vX,\vY) {\Pg{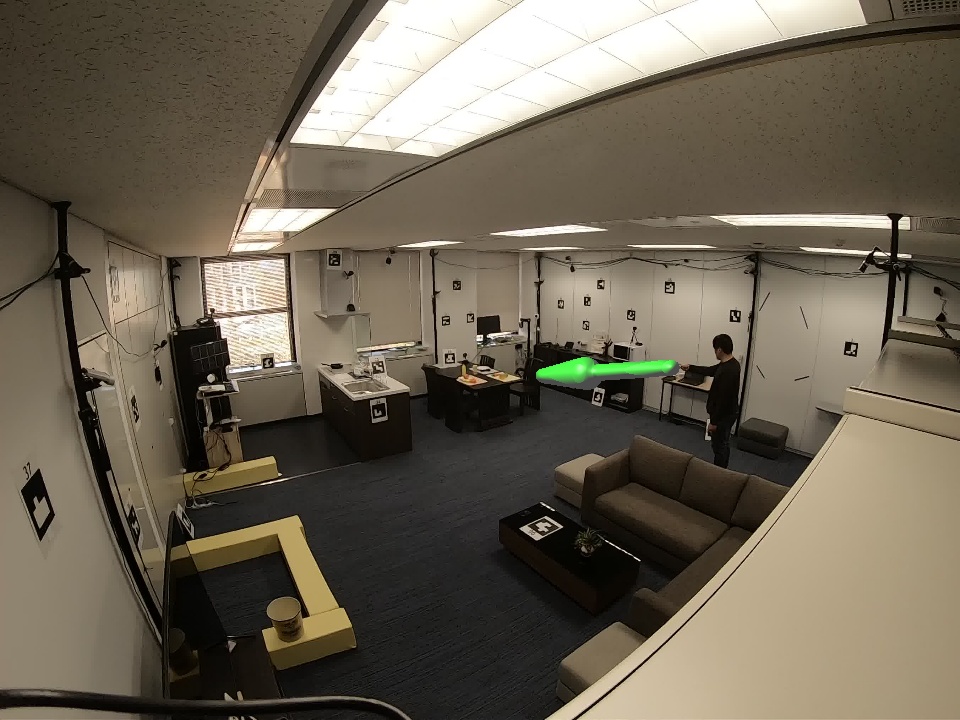}};
    \advance\vY by \vdY
    \node at (\vX,\vY) {\Pg{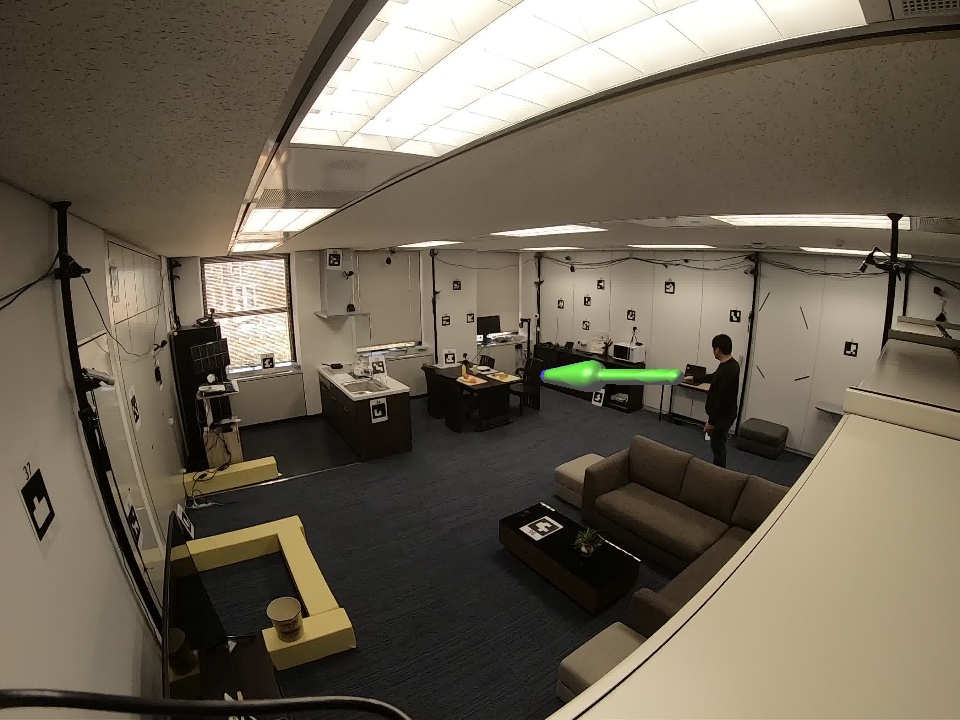}};

    \advance\vX by \vdX
    \vY = \vYi
    \node at (\vX,\vY) {\Ph{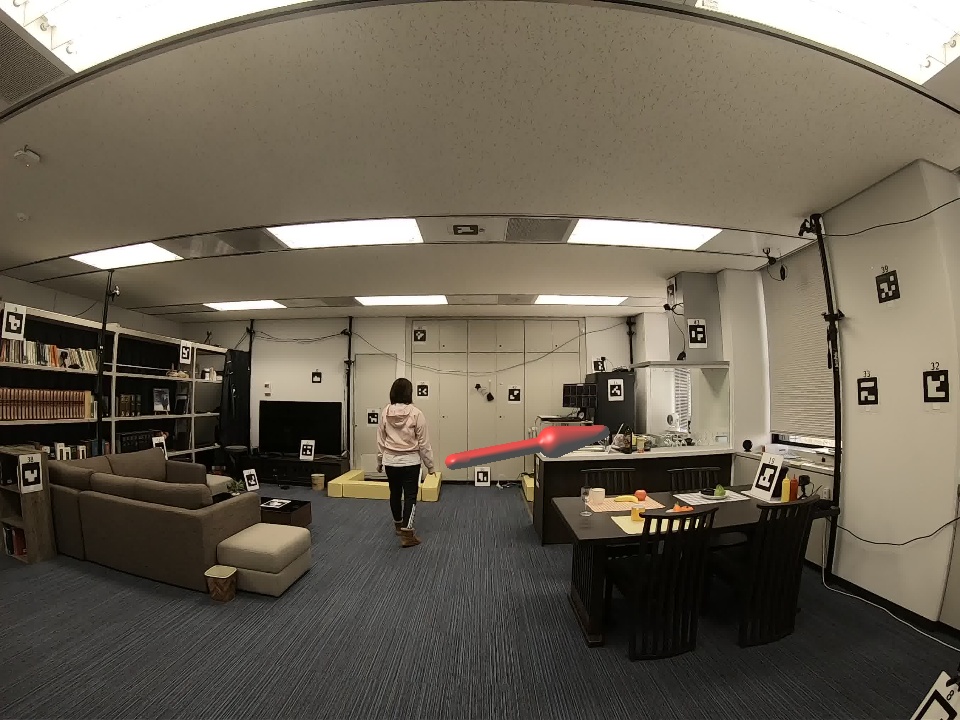}};
    \advance\vY by \vdY
    \node at (\vX,\vY) {\Ph{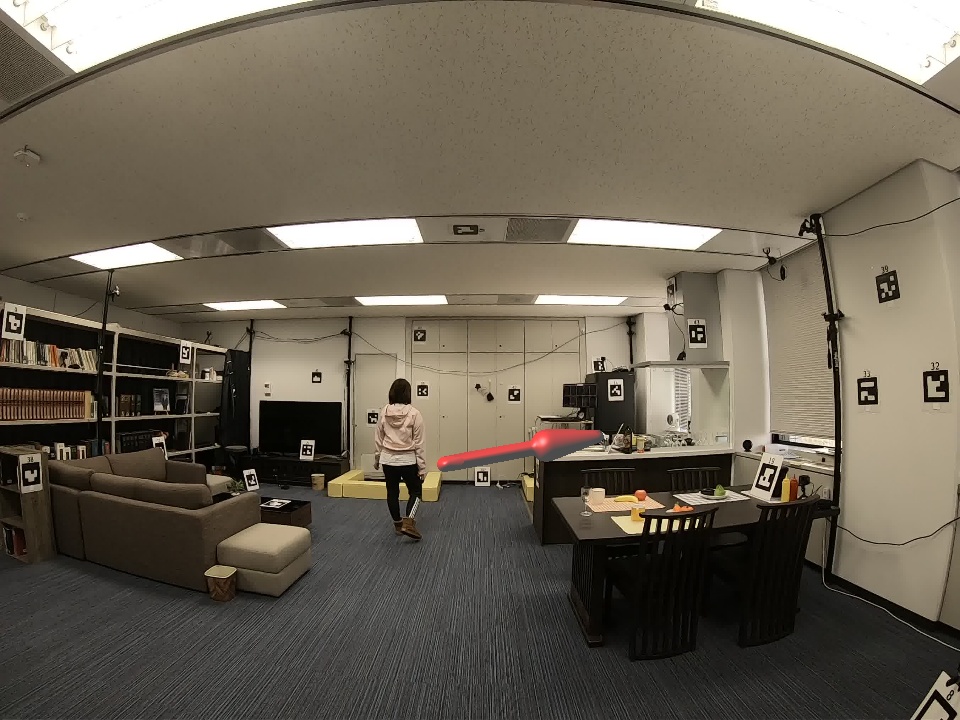}};
    \advance\vY by \vdY
    \node at (\vX,\vY) {\Ph{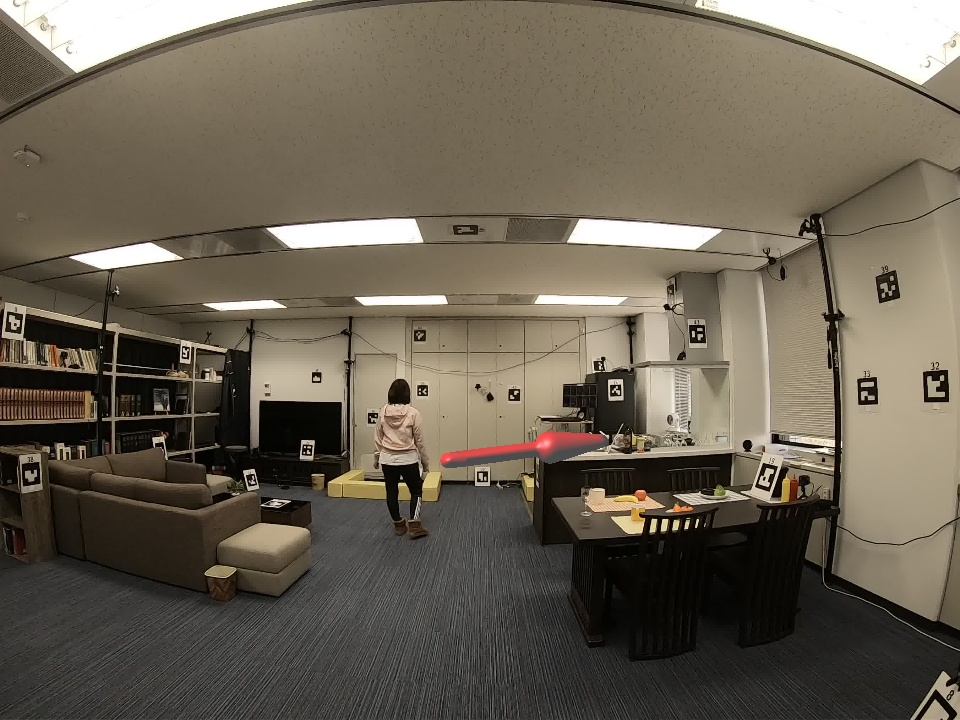}};
    \advance\vY by \vdY
    \node at (\vX,\vY) {\Ph{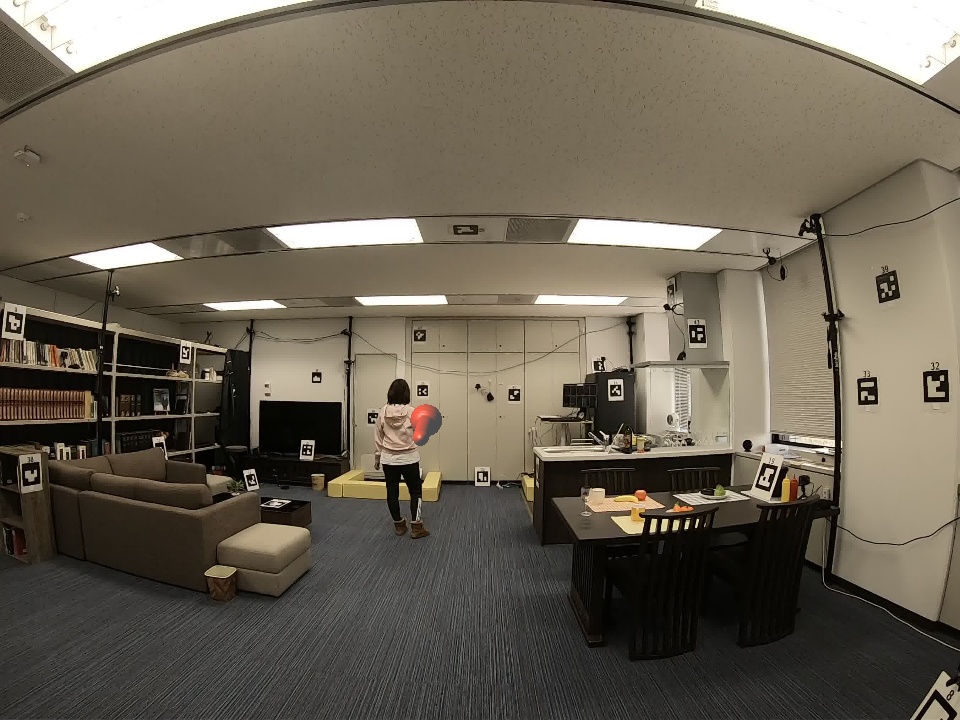}};
    \advance\vY by \vdY
    \node at (\vX,\vY) {\Ph{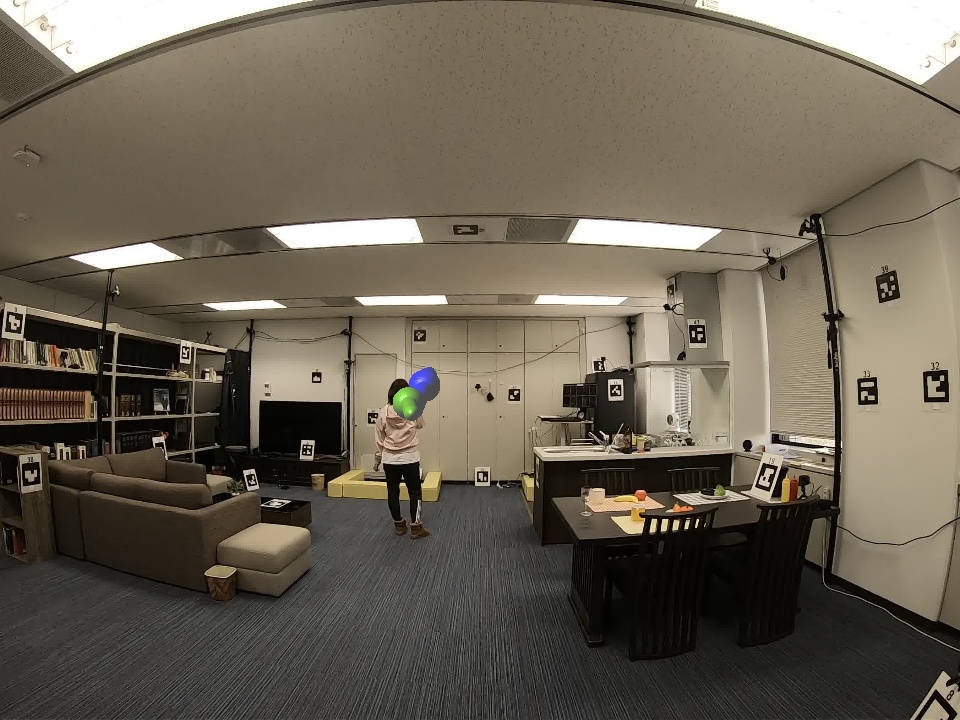}};
    \advance\vY by \vdY
    \node at (\vX,\vY) {\Ph{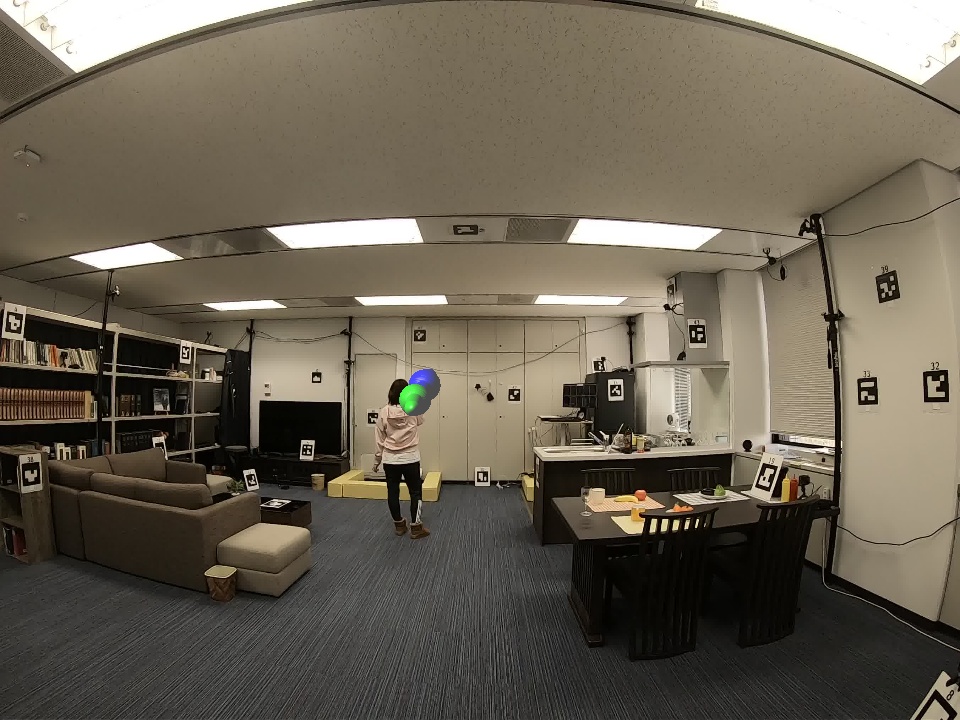}};

    \end{tikzpicture}
    \caption{Pointing direction estimation by DeePoint trained with \SplitT (\ie, \OursA). In each image, the blue arrow denotes the ground truth direction and the other arrow denotes the estimated 3D direction by DeePoint. The color of the prediction arrow represents the result of pointing action recognition. It is green when the person is pointing ($p=1$), red when not ($p=0$), and gradually transitions between the two colors based on estimated probability. Note how DeePoint correctly recognizes the timing of pointing. For instance, it learns to recognize when the person looks away as the finish of pointing and finds the onset of pointing from change in speed of the movements of the body coordination.}
    \label{fig:result}
\end{figure*}

\else

\begin{figure*}[t]
    \centering
    \def\Pa#1{\includegraphics[width=0.22\linewidth,bb={320 166 960 440},clip]{#1}}
    \def\Pb#1{\includegraphics[width=0.22\linewidth,bb={8   267 808 610},clip]{#1}}
    \def\Pc#1{\includegraphics[width=0.22\linewidth,bb={421 213 960 444},clip]{#1}}
    \def\Pd#1{\includegraphics[width=0.22\linewidth,bb={82  171 722 445},clip]{#1}}
    \begin{tikzpicture}[x=1.02mm,y=1mm]
    \newcount\vX
    \newcount\vY
    \newcount\vdX
    \newcount\vdY
    \newcount\vXi
    \newcount\vYi
    \vdX = 41
    \vdY = -16
    \vXi = -24
    \vYi = -100

    \vX = \vXi
    \node[inner sep=0pt, font=\scriptsize] (a) at (\vX,\vY) {$t=0$};
    \advance\vY by \vdY
    \node[inner sep=0pt, font=\scriptsize] (a) at (\vX,\vY) {$t=3$};
    \advance\vY by \vdY
    \node[inner sep=0pt, font=\scriptsize] (a) at (\vX,\vY) {$t=6$};
    \advance\vY by \vdY
    \node[inner sep=0pt, font=\scriptsize] (a) at (\vX,\vY) {$t=9$};
    \advance\vY by \vdY
    \node[inner sep=0pt, font=\scriptsize] (a) at (\vX,\vY) {$t=12$};
    \advance\vY by \vdY
    \node[inner sep=0pt, font=\scriptsize] (a) at (\vX,\vY) {$t=15$};
    \advance\vY by \vdY

    \vX = 0
    \vY = 0
    \node at (\vX,\vY) {\Pa{fig2/s1/out0000000010.jpg}};
    \advance\vY by \vdY
    \node at (\vX,\vY) {\Pa{fig2/s1/out0000000013.jpg}};
    \advance\vY by \vdY
    \node at (\vX,\vY) {\Pa{fig2/s1/out0000000016.jpg}};
    \advance\vY by \vdY
    \node at (\vX,\vY) {\Pa{fig2/s1/out0000000019.jpg}};
    \advance\vY by \vdY
    \node at (\vX,\vY) {\Pa{fig2/s1/out0000000022.jpg}};
    \advance\vY by \vdY
    \node at (\vX,\vY) {\Pa{fig2/s1/out0000000025.jpg}};

    \advance\vX by \vdX
    \vY = 0
    \node at (\vX,\vY) {\Pb{fig2/s2/out0000000065.jpg}};
    \advance\vY by \vdY
    \node at (\vX,\vY) {\Pb{fig2/s2/out0000000068.jpg}};
    \advance\vY by \vdY
    \node at (\vX,\vY) {\Pb{fig2/s2/out0000000071.jpg}};
    \advance\vY by \vdY
    \node at (\vX,\vY) {\Pb{fig2/s2/out0000000074.jpg}};
    \advance\vY by \vdY
    \node at (\vX,\vY) {\Pb{fig2/s2/out0000000077.jpg}};
    \advance\vY by \vdY
    \node at (\vX,\vY) {\Pb{fig2/s2/out0000000080.jpg}};

    \advance\vX by \vdX
    \vY = 0
    \node at (\vX,\vY) {\Pc{fig2/s3/out0000000174.jpg}};
    \advance\vY by \vdY
    \node at (\vX,\vY) {\Pc{fig2/s3/out0000000177.jpg}};
    \advance\vY by \vdY
    \node at (\vX,\vY) {\Pc{fig2/s3/out0000000180.jpg}};
    \advance\vY by \vdY
    \node at (\vX,\vY) {\Pc{fig2/s3/out0000000183.jpg}};
    \advance\vY by \vdY
    \node at (\vX,\vY) {\Pc{fig2/s3/out0000000186.jpg}};
    \advance\vY by \vdY
    \node at (\vX,\vY) {\Pc{fig2/s3/out0000000189.jpg}};

    \advance\vX by \vdX
    \vY = 0
    \node at (\vX,\vY) {\Pd{fig2/s4/out0000000262.jpg}};
    \advance\vY by \vdY
    \node at (\vX,\vY) {\Pd{fig2/s4/out0000000265.jpg}};
    \advance\vY by \vdY
    \node at (\vX,\vY) {\Pd{fig2/s4/out0000000268.jpg}};
    \advance\vY by \vdY
    \node at (\vX,\vY) {\Pd{fig2/s4/out0000000271.jpg}};
    \advance\vY by \vdY
    \node at (\vX,\vY) {\Pd{fig2/s4/out0000000274.jpg}};
    \advance\vY by \vdY
    \node at (\vX,\vY) {\Pd{fig2/s4/out0000000277.jpg}};

    \end{tikzpicture}
    \caption{Pointing direction estimation by DeePoint trained with \SplitT (\ie, \OursA). In each image, the blue arrow denotes the ground truth direction and the other arrow denotes the estimated 3D direction by DeePoint. The color of the prediction arrow represents the result of pointing action recognition. It is green when the person is pointing ($p=1$), red when not ($p=0$), and gradually transitions between the two colors based on estimated probability. Note how DeePoint correctly recognizes the timing of pointing. For instance, it learns to recognize when the person looks away as the finish of pointing and finds the onset of pointing from change in speed of the movements of the body coordination.}
    \label{fig:result}
\end{figure*}
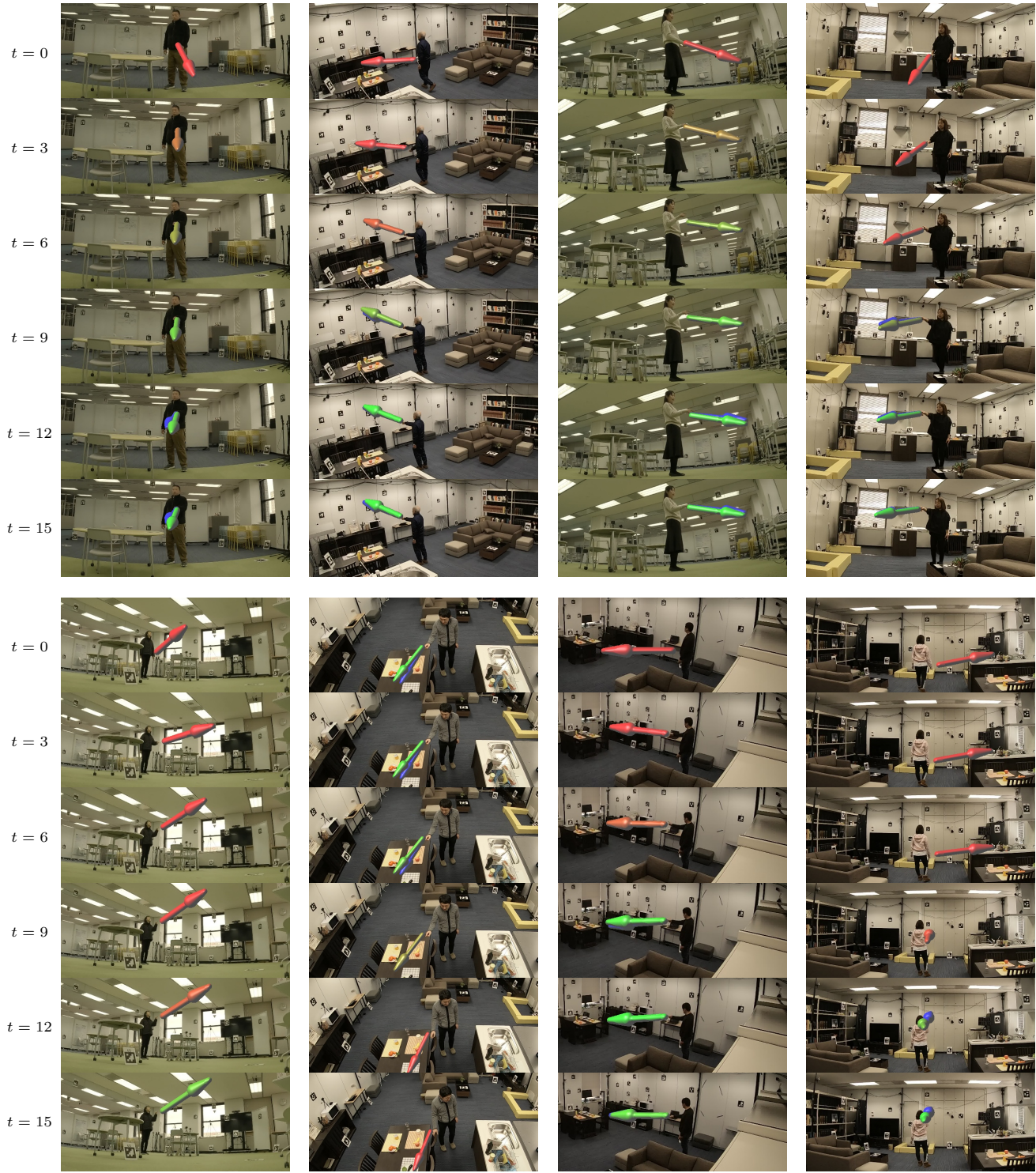
\fi

\paragraph{Network architecture}

DeePoint uses visual features around the detected joints to encode the body posture and its specific instantiation. In addition, we may also leverage visual features of the whole body and even encode the entire captured image. Since each token of \JE is a 192-dimensional vector, we can add these contextual visual features that encode the body and scene appearance into the class token since it is not associated with a specific joint.

As the same for the visual features at each joint, we can apply the same pre-trained ResNet-34 to the image cropped by the bounding box of the detected person and the entire image, apply ROI align to obtain \numproduct{16 x 16 x 256} feature vectors and project them into 192-dimensional vectors with the same linear projection. These vectors are then added to the learnable class token. In what follows, we denote the barebone DeePoint as \OursA, a variant adding the whole-body visual feature to the class token as \OursB, and yet another variant adding both the whole-body and the entire-image visual features to the class token as \OursC.

\paragraph{Data split}
Our DP Dataset consists of roughly \num{2800000} frames of captured sessions of 33 subjects in two different rooms (Living Room and Office). We define the following three splits for evaluation.
\begin{description}[leftmargin=0pt,labelindent=0pt,itemsep=0pt,topsep=0.25\baselineskip,partopsep=0pt]
\item[\it \SplitT (temporal split)] Each session of the subjects is split into 70\%, 15\%, and 15\% from the beginning and used in the training, validation, and test sets, respectively. The three sets share the same subjects and the scenes, but not the pointing instances and their directions. This split lets us evaluate the intra-personal accuracy of DeePoint.
\item[\it \SplitS (scene split)] The training set does not share the same room with the validation and test set. That is, the training set is only taken from Living Room, and the validation and the test sets are from Office. This split lets us study the cross-scene accuracy of DeePoint.
\item[\it \SplitP (person split)] Each of the 33 subjects appears only in one of the training, validation, or test set. We allocated 25, 4, and 4 subjects for the training, validation, and test sets, respectively. This split lets us evaluate the inter-personal accuracy of DeePoint.
\end{description}

\begin{table}[t]
    \centering \begingroup
    \setlength{\tabcolsep}{1.5mm}
    \begin{tabular}{@{}llll@{}}\toprule
        Model                             & \SplitT          & \SplitS          & \SplitP          \\
        \midrule
        \OursA                            & $14.05$          & $\mathbf{17.52}$ & $\mathbf{13.85}$ \\
        \OursB                            & $\mathbf{13.66}$ & $17.63$          & $13.93$          \\
        \OursC                            & $14.12$          & $17.62$          & $14.91$          \\ \bottomrule
    \end{tabular}
    \endgroup
    \vspace{1mm}
    \caption{Pointing direction estimation errors, denoted in degree. We can observe that the proposed model generalizes well in every split.}
    \label{tab:results_split_angle}
\end{table}

\begin{table}[t]
    \centering
    \begingroup
    \setlength{\tabcolsep}{1.5mm}
    \begin{tabular}{@{}llll@{}}\toprule
        Model                              & \SplitT                & \SplitS                         & \SplitP                \\
        \midrule
        \OursA                             & $0.625/0.838$          & $0.629/0.685$                   & $\mathbf{0.476}/0.816$ \\
        \OursB                             & $0.627/\mathbf{0.852}$ & $0.597/0.732$                   & $0.445/0.823$          \\
        \OursC                             & $\mathbf{0.650}/0.837$ & $\mathbf{0.634}/\mathbf{0.740}$ & $0.456/\mathbf{0.855}$ \\ \bottomrule
    \end{tabular}
    \endgroup
    \vspace{1mm}
    \caption{Recall (left) and precision (right) for the pointing action detection. Note that these values are calculated frame by frame and the percentage of pointing actions that are missed completely are much lower.}
    \label{tab:results_split_RP}
\end{table}

\Cref{tab:results_split_angle,tab:results_split_RP} each reports the mean angular errors of pointing direction and recall/precision of pointing detection by frame.
Each model is trained using the training set with learning rate = \num{e-4} with Adam~\cite{Kingma2015Adam} optimizer and batch size = \num{64} until convergence.  We use the best parameter within the epochs in terms of angular error measured with the validation set.

The results provide insights into the role of the body and scene context.
For the intra-personal split (\SplitT), \OursB and \OursC detects pointing better than \OursA as they can leverage the access to scene context.
Performance for direction estimation and precision gets worse for \SplitS, which indicates changes in the way of pointing and background could affect the performance and the importance of training with a dataset that contains multiple venues.
On the other hand, as for \SplitP, recall is relatively low, which indicates the way of pointing differ in people.
While integrating whole-body and scene context contributes to improving the performance for 3D direction estimation, the performance for pointing detection is better without them for \SplitS and \SplitP, likely due to overfitting.
How to encode personal (\ie, body appearance) and scene (\ie, image) context in DeePoint such that we may fully leverage their representational power while avoiding overfitting is a challenge we will explore in future work.

As can be seen in \cref{fig:result}, in general, DeePoint achieves high recall and reasonable angular accuracy, especially for a completely passive method relying only on viewpoint from relatively far distance. Even though roughly \num{18} degrees of error may appear large, given the relative distance to the objects in the scene, in many cases, it is sufficient to identify what is actually pointed at.
Note that the detection recall and precision are calculated by each frame and only a fraction of pointing actions are missed completely. For example, \OursA on \SplitT test set missed only \num{5.8}\% of pointing actions (\num{94.2}\% recall).

\usetikzlibrary{calc}
\begin{figure}
    \centering
    \def\Obj#1{\includegraphics[width=0.21\linewidth,bb={414 30 1364 981},clip]{#1}}
    \def\Pic#1{\includegraphics[width=0.21\linewidth]{#1}}
    \begin{tikzpicture}[x=1pt,y=1pt]
        \newcount\vX
        \newcount\dvX
        \newcount\vY
        \newcount\arrowOffset
        \dvX = 55
        \vX = 0
        \vY = 0
        \arrowOffset = 4
        \node [inner sep=0pt] (pic0) at (\vX,\vY) {\Pic{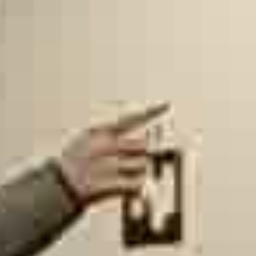}};
        \draw [<->] ($ (pic0.north west) + (0,\arrowOffset) $) -- node [inner xsep=0.5mm, inner ysep=0mm, fill=white, font=\scriptsize] {60px} ($ (pic0.north east) + (0,\arrowOffset) $);
        \advance\vX by \dvX
        \node [inner sep=0pt] at (\vX,\vY) (pic1) {\Pic{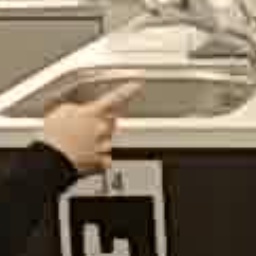}};
        \draw [<->] ($ (pic1.north west) + (0,\arrowOffset) $) -- node [inner xsep=0.5mm, inner ysep=0mm, fill=white, font=\scriptsize] {98px} ($ (pic1.north east) + (0,\arrowOffset) $);
        \advance\vX by \dvX
        \node [inner sep=0pt] at (\vX,\vY) (pic2) {\Pic{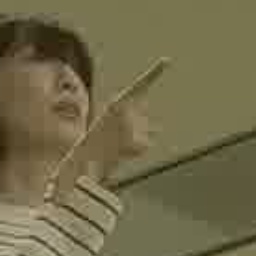}};
        \draw [<->] ($ (pic2.north west) + (0,\arrowOffset) $) -- node [inner xsep=0.5mm, inner ysep=0mm, fill=white, font=\scriptsize] {112px} ($ (pic2.north east) + (0,\arrowOffset) $);
        \advance\vX by \dvX
        \node [inner sep=0pt] at (\vX,\vY) (pic3) {\Pic{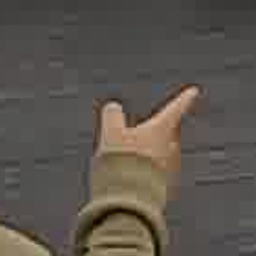}};
        \draw [<->] ($ (pic3.north west) + (0,\arrowOffset) $) -- node [inner xsep=0.5mm, inner ysep=0mm, fill=white, font=\scriptsize] {113px} ($ (pic3.north east) + (0,\arrowOffset) $);
        \vX = 0
        \vY = -50
        \node at (\vX,\vY) {\Obj{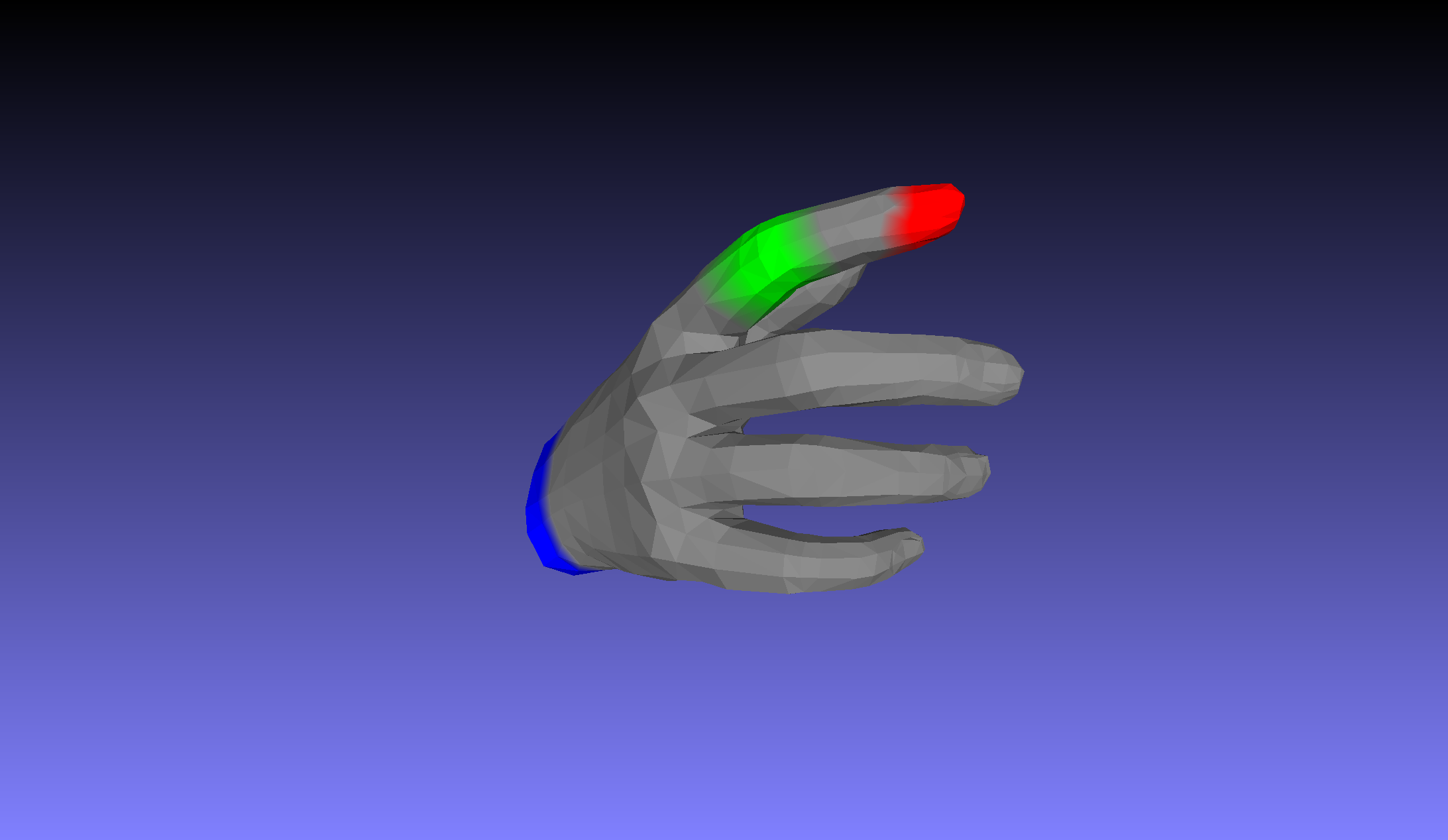}};
        \advance\vX by \dvX
        \node at (\vX,\vY) {\Obj{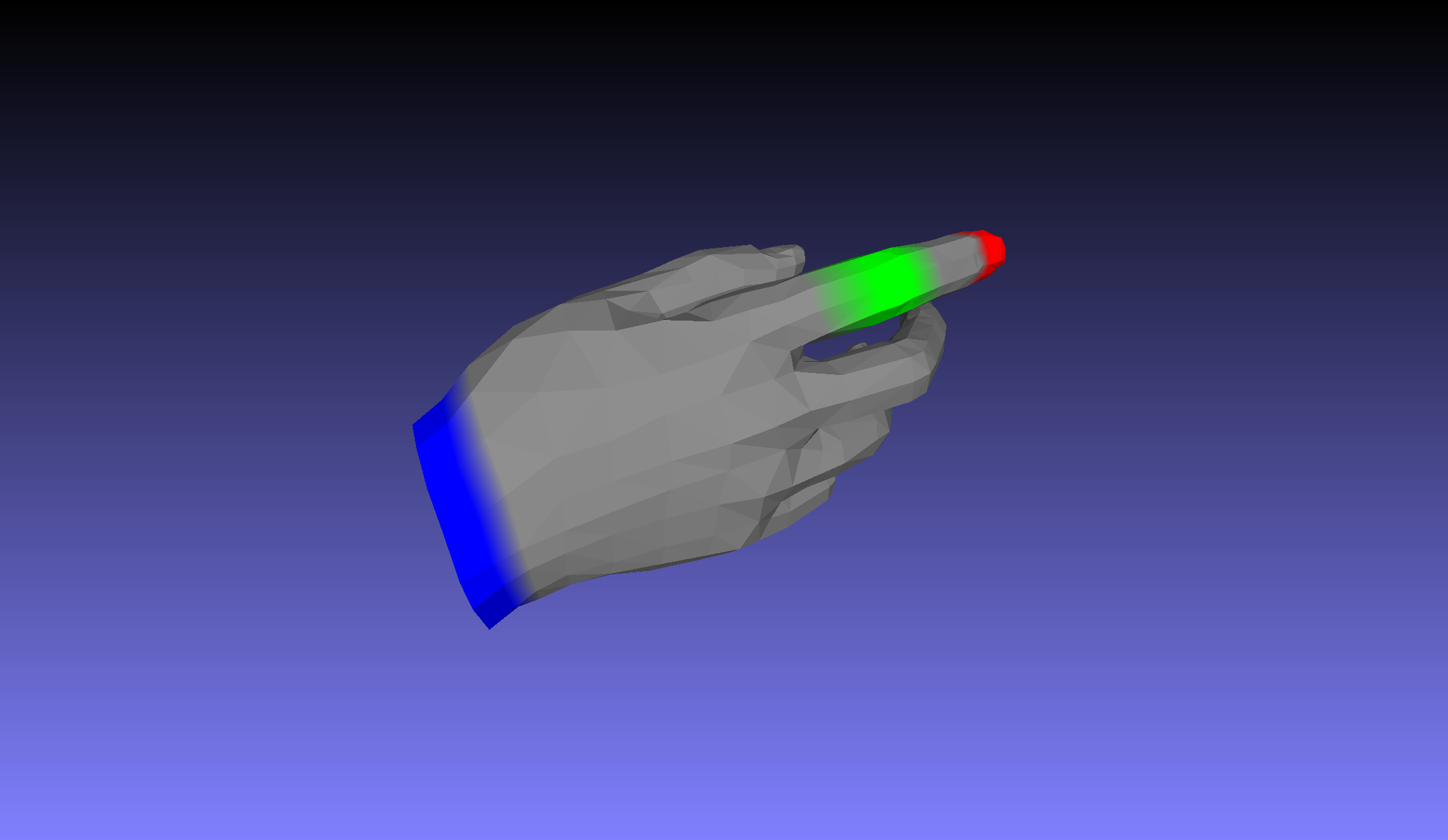}};
        \advance\vX by \dvX
        \node at (\vX,\vY) {\Obj{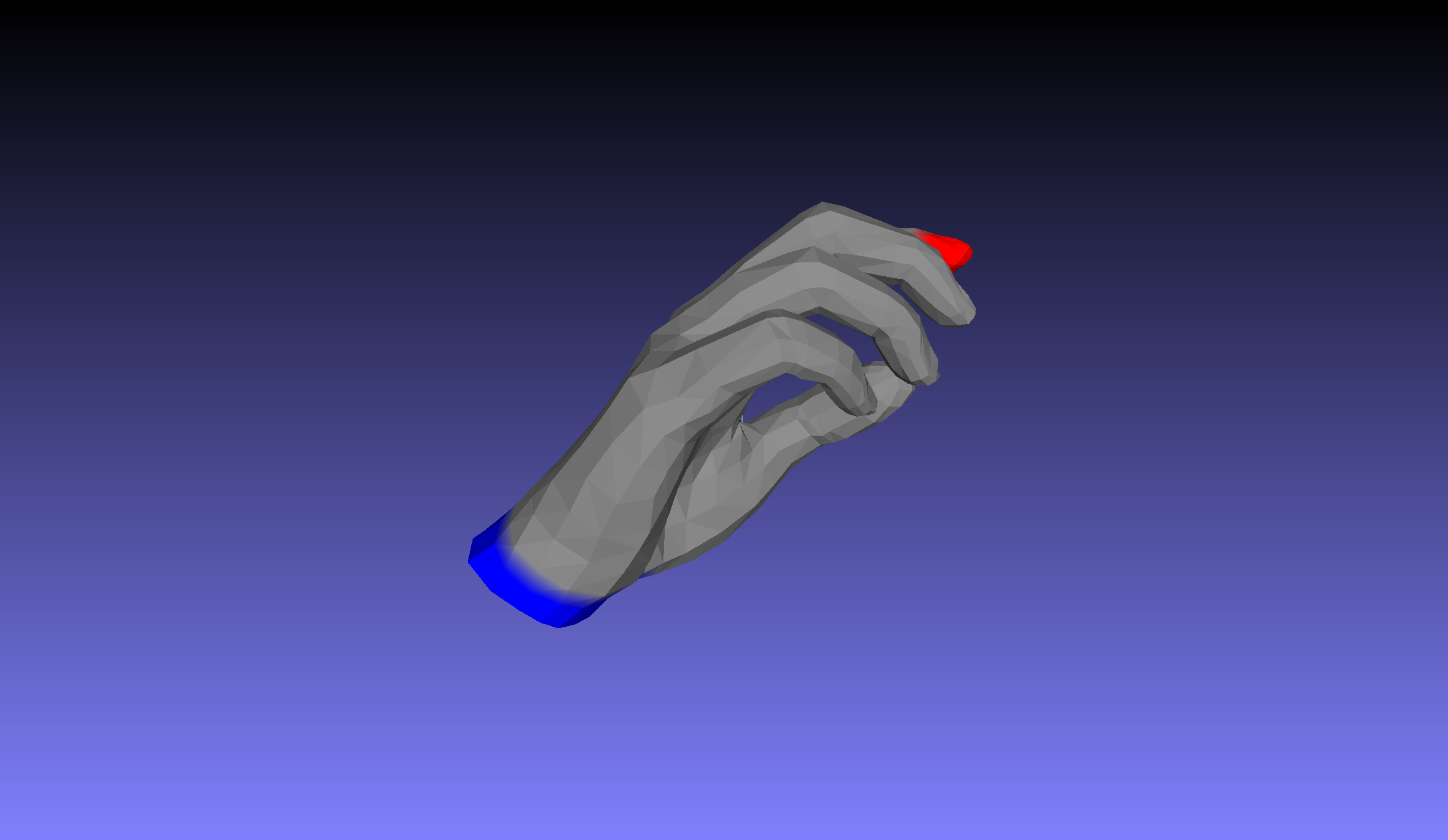}};
        \advance\vX by \dvX
        \node at (\vX,\vY) {\Obj{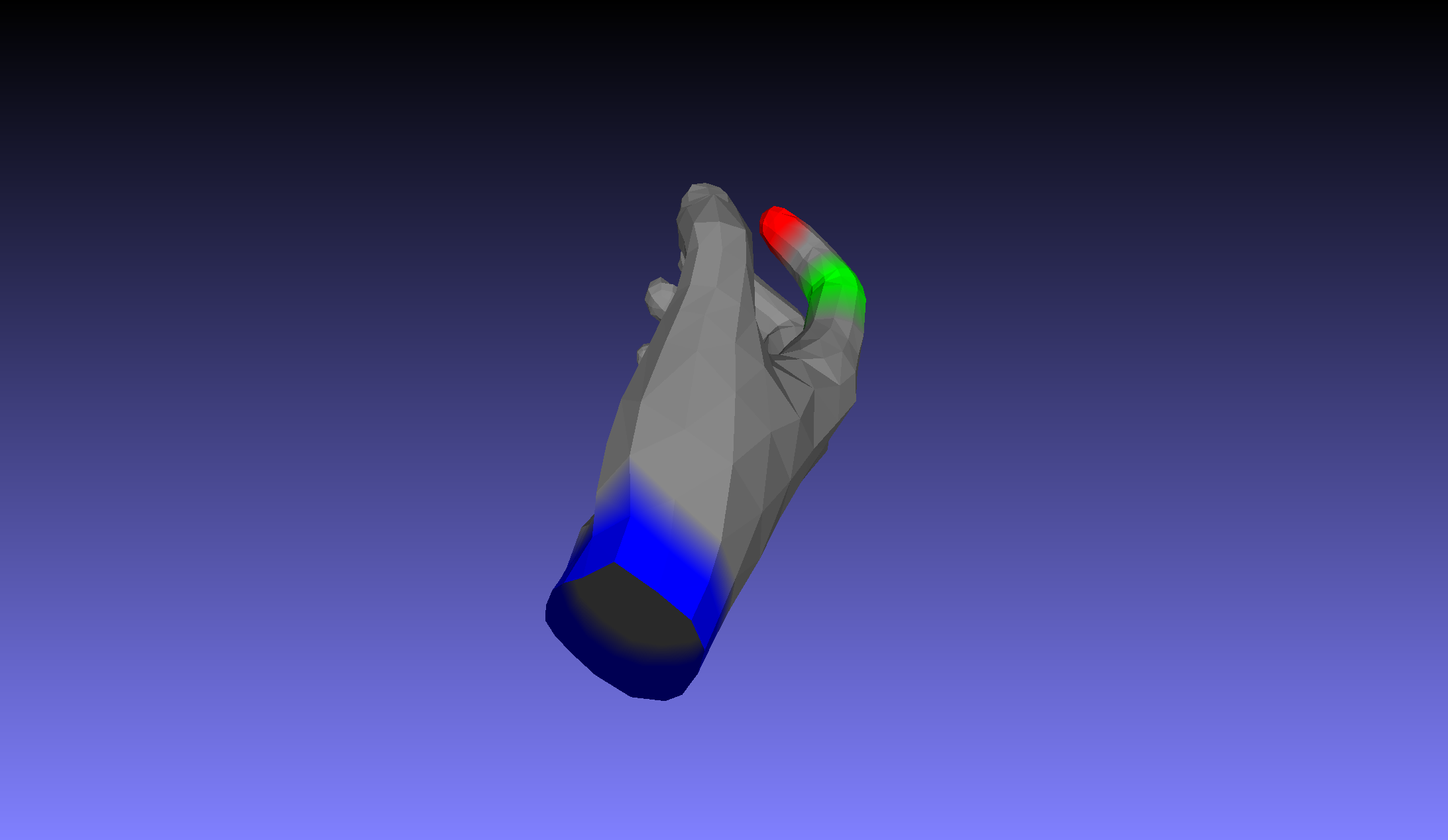}};
        \vX = 0
        \vY = -80
        \node[inner sep=0pt, font=\scriptsize] at (\vX,\vY) {(a) error=$8.38\tcdegree$};
        \advance\vX by \dvX
        \node[inner sep=0pt, font=\scriptsize] at (\vX,\vY) {(b) $22.4\tcdegree$};
        \advance\vX by \dvX
        \node[inner sep=0pt, font=\scriptsize] at (\vX,\vY) {(c) $53.9\tcdegree$};
        \advance\vX by \dvX
        \node[inner sep=0pt, font=\scriptsize] at (\vX,\vY) {(d) $75.4\tcdegree$};
    \end{tikzpicture}
    \caption{Examples of HandOccNet reconstruction on the DP Dataset, sorted by index finger direction error. HandOccNet fails to reconstruct the pointing index finger for most cases, especially as the hands are small in regular videos of people, showing the fragility of 3D hand reconstruction-based pointing understanding.}
    \label{fig:handoccnet_result_examples}
\end{figure}
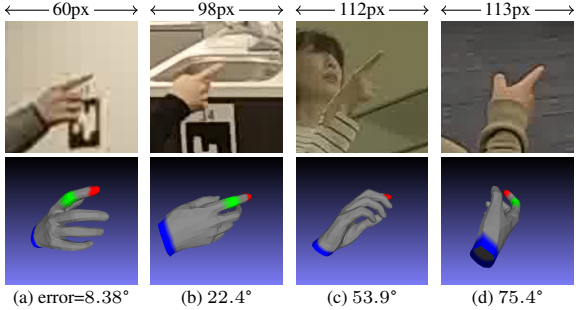

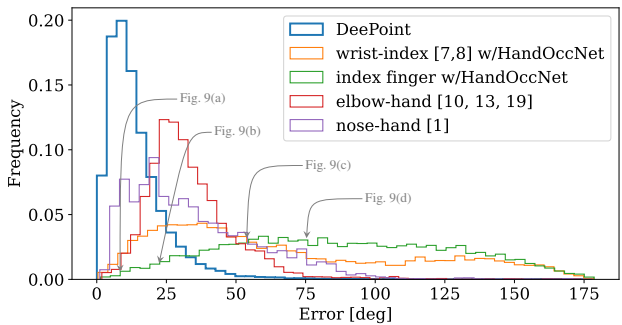
\begin{figure}

    \centering
    \begin{tikzpicture}
        \node [anchor=south west, inner sep=0] (figure) at (0,0) {\includegraphics[width=\linewidth]{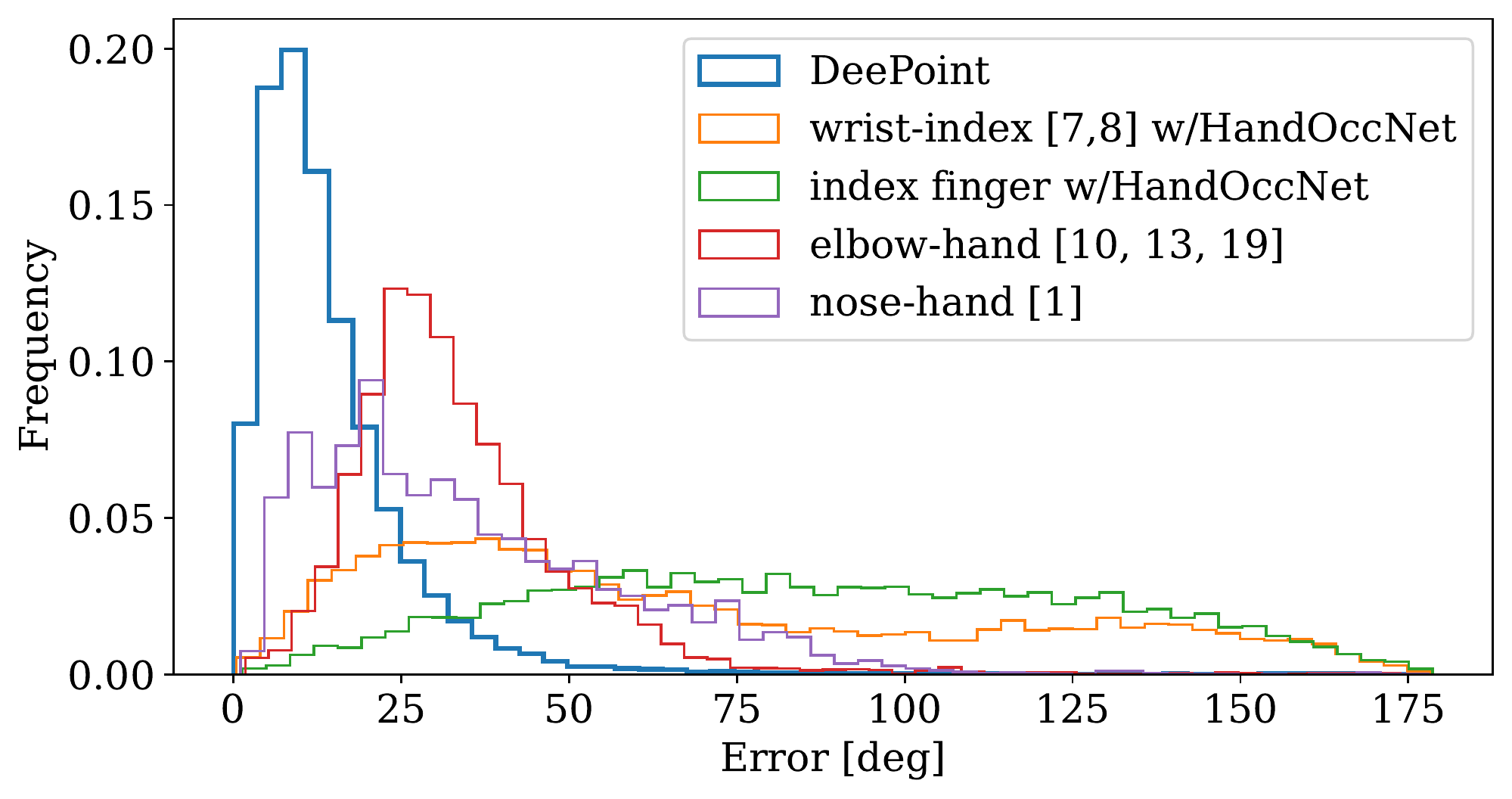}};
            \begin{scope}[x={(figure.south east)},y={(figure.north west)}]
                \draw [very thin, color=gray, <-, >=stealth] (0.192123,0.180) .. controls (0.192123,0.7) .. (0.283,0.7) node [right, inner sep=1pt, font=\tiny] {Fig.\ 9(a)};
                \draw [very thin, color=gray, <-, >=stealth] (0.254232,0.206) .. controls (0.299,0.6)    .. (0.338,0.6) node [right, inner sep=1pt, font=\tiny] {Fig.\ 9(b)};
                \draw [very thin, color=gray, <-, >=stealth] (0.393777,0.281) .. controls (0.398,0.5)    .. (0.483,0.5) node [right, inner sep=1pt, font=\tiny] {Fig.\ 9(c)};
                \draw [very thin, color=gray, <-, >=stealth] (0.489022,0.281) .. controls (0.493,0.4)    .. (0.578,0.4) node [right, inner sep=1pt, font=\tiny] {Fig.\ 9(d)};
            \end{scope}
    \end{tikzpicture}
    \caption{Error histograms of past methods compared with that of DeePoint evaluated on the test data of the DP Dataset. DeePoint clearly outperforms all.}
    \label{fig:comparison}
\end{figure}

\subsection{Ablation Study On Temporal Window}

\begin{table}[t]
    \centering
    \begingroup
    \begin{tabular}{@{}lllll@{}}\toprule
        Temporal window & Angular error $(\downarrow)$ & Prec./Rec. $(\uparrow)$ \\
        \midrule
        $N=1$  & $17.08^\circ$                      & $0.519/0.801$             \\
        $N=5$  & $14.90^\circ$                      & $0.585/0.828$             \\
        $N=15$ & $14.05^\circ$                      & $0.625/\mathbf{0.838}$    \\
        $N=30$ & $\mathbf{13.58^\circ}$             & $\mathbf{0.637}/0.833$ \\
        \bottomrule
    \end{tabular}
    \endgroup
    \vspace{2mm}
    \caption{Contributions of the size of the temporal window $N$.  We can observe that $N=15$ corresponding to 1 second of the observation is a reasonable design choice as the performance gain by $N=30$ is marginal.}
    \label{tab:results_window}
\end{table}

The size of the temporal window, \ie, the number of tokens $N$ given to \TE is set to $N=15$ as the default value.  \Cref{tab:results_window} shows the results with different values of $N$ from $N=1$ to $N=30$. $N=1$ corresponds to single-shot pointing detection and direction estimation, and $N=5$, $15$, and $30$ correspond to 1/3, 1, and 2 seconds of the observed video.

From these results, we can conclude that $N=15$, which is used in DeePoint, is a reasonable design choice as the performance gain by $N=30$ is marginal, while $N=1$ and $N=5$ do not perform well, especially in action detection. This result can be interpreted intuitively that most pointing instances can last up to a second and not shorter than 1/3 seconds, and $N=15$ is a reasonable length to cover such actions.

\begin{figure*}[t]
    \centering
    \includegraphics[width=0.23\linewidth]{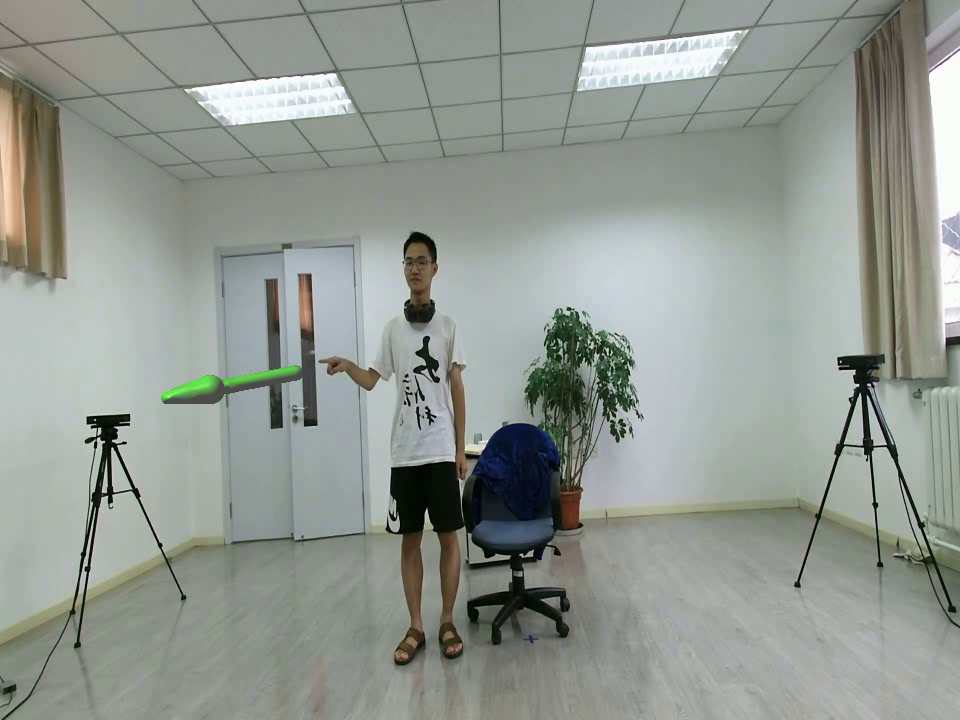}\hfil%
    \includegraphics[width=0.23\linewidth]{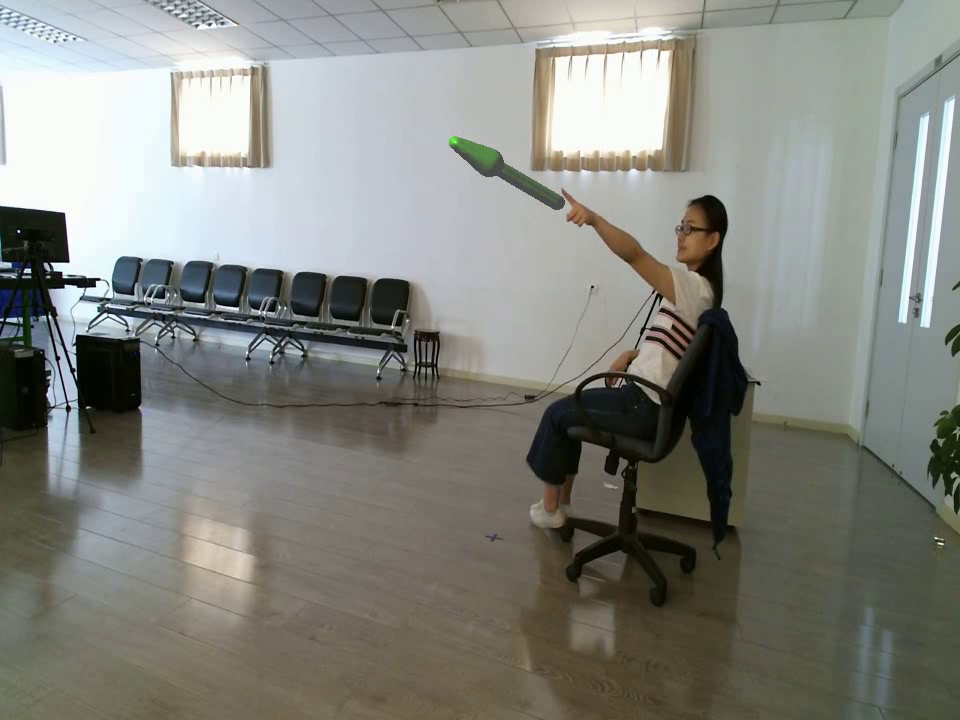}\hfil%
    \includegraphics[width=0.23\linewidth]{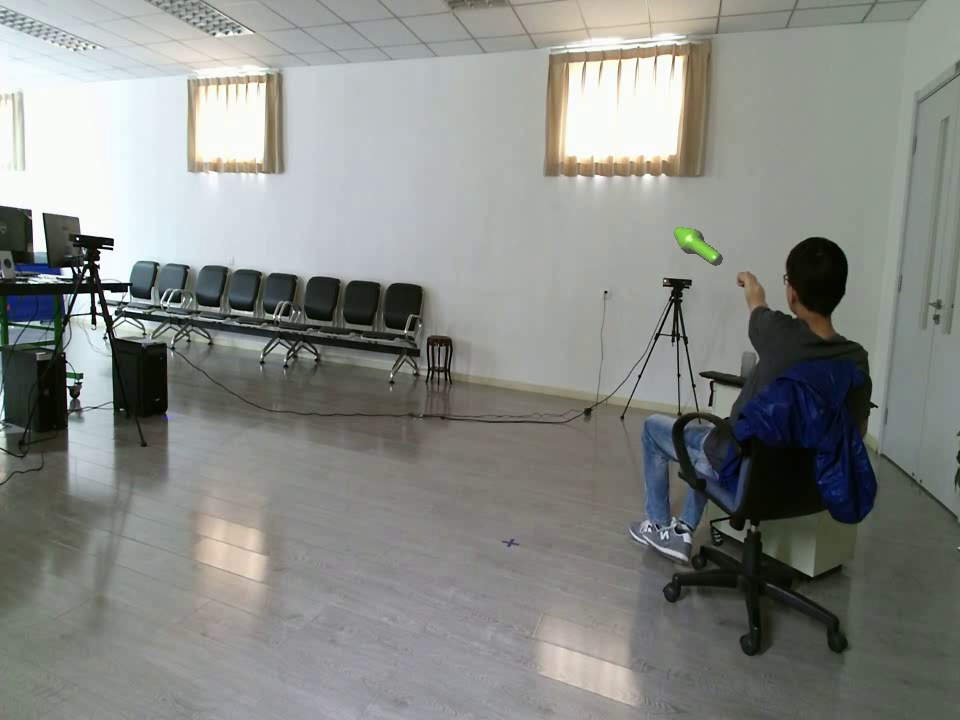}\hfil%
    \includegraphics[width=0.23\linewidth]{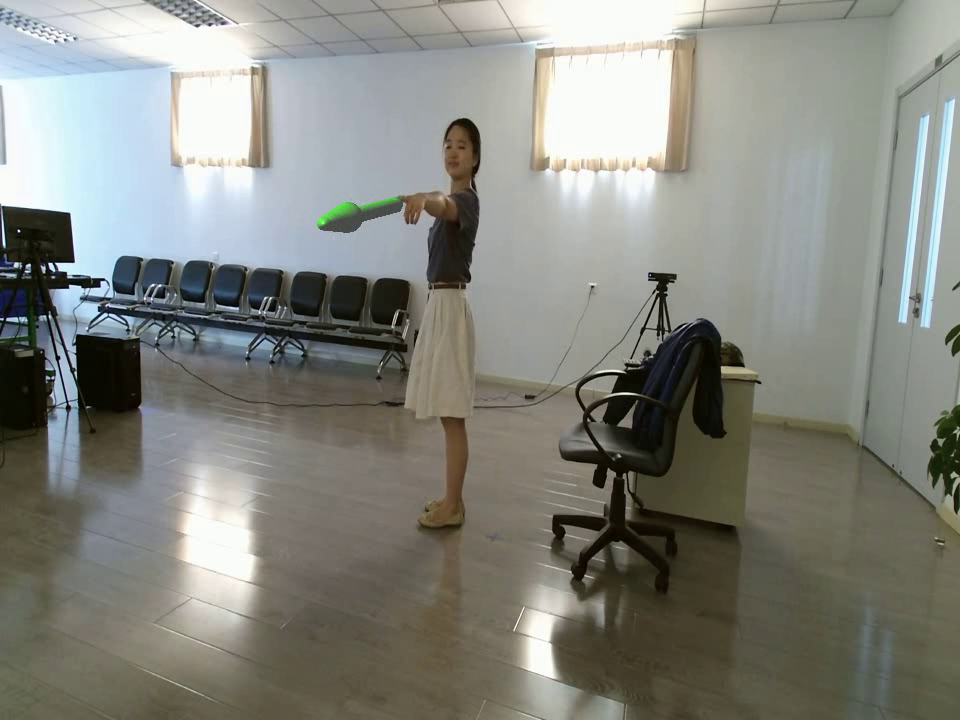}
    \caption{Qualitative evaluation with the PKU-MMD dataset~\cite{Liu17_PKUMMD}. Note that our model is not retrained on the PKU-MMD and applied out-of-the-box. DeePoint generalizes well to a completely different dataset.}
    \label{fig:pkummd}
\end{figure*}

\subsection{Comparison with Baseline Methods}

\newcommand{\blue}{{\color{blue}blue}\xspace}
\newcommand{\green}{{\color{green!70!black}green}\xspace}
\newcommand{\red}{{\color{red}red}\xspace}

We implement baseline methods that represent past methods and evaluate them using the test split of the DP dataset, and compare their results against that of DeePoint.
Directly evaluating DeePoint on the datasets used in the past methods is not possible, as most of them are simply not published~\cite{Azari2019, Zuzana2007, Dhingra2020, Fernandez:2015, Shruti2018, Shiratori2021}. Even when they are, they contain only images (not videos) and capture only hands or arms~\cite{Das2021,Shukla:2015}.
To the best of our knowledge, the only exception is PKU-MMD~\cite{Liu17_PKUMMD}, a video dataset annotated with various action timings, including pointing. We'll discuss it in \cref{subsec:PKUMMD}.

To evaluate the accuracy of a single-view learning-based approach, we use HandOccNet~\cite{Park_2022_CVPR_HandOccNet} to recover a 3D hand mesh from a single image and use the recovered hand to estimate the pointing direction. Since pointing is a manual gesture, it may appear possible to estimate pointing direction by connecting vertices of the mesh.
As shown in \cref{fig:handoccnet_result_examples}, we applied HandOccNet to hand image regions extracted from the DP dataset to test this.
We tried two alternatives for direction estimation: from the wrist to the tip of the index finger (\ie, from the center of the \blue vertices to that of the \red ones in \cref{fig:handoccnet_result_examples})~\cite{Das2018, Das2021} and from the base to the tip of the index finger (from \green to \red).
\Cref{fig:comparison} shows that the results are poor. This is because, as can be seen in \cref{fig:handoccnet_result_examples}, HandOccNet fails to reconstruct the index finger accurately for most cases due to the low resolution of the hand regions.

We also replicated geometry-based approaches using the 3D keypoints of the DP dataset.
Most of these methods calculate the pointing direction by estimating the 3D coordinates of keypoints and connecting them (elbow to wrist~\cite{Dhingra2020,Fernandez:2015,Shruti2018} or face to hand~\cite{Azari2019}). The triangulated keypoints in the DP Dataset can be used to simulate these approaches (elbow to hand or nose to hand).
The results are also depicted in \cref{fig:comparison} and they clearly show that our DeePoint estimates are more accurate by a wide margin. 

These results clearly show that modeling the whole body movements is essential to achieve accurate pointing direction estimation, especially for in-the-wild scenarios in which the person is captured from afar.

\subsection{Cross-Validation with PKU-MMD dataset}
\label{subsec:PKUMMD}
Although there does not exist a large-scale pointing dataset in the community, PKU-MMD~\cite{Liu17_PKUMMD}, a video dataset of various actions, contains a small number of pointing videos in it. We used PKU-MMD Phase \# 2 which originally consists of 7,000 action instances of 41 action classes, performed by 13 subjects. We use one of the 41 action classes named ``pointing to something with finger'' (class 24) for validation. The dataset contains 817,314 non-action frames, 497,296 non-pointing action frames, and 6,708 pointing frames. The pointing directions, however, are not annotated in PKU-MMD and we can only conduct qualitative evaluations.

\Cref{fig:pkummd} shows the pointing directions estimated by DeePoint (\OursA) trained with \SplitT of DP Dataset. Note that the model is not fine-tuned with PKU-MMD (as there are no ground truth directions to fine-tune on). The detection accuracy was 72.2\% for the pointing + non-action frames, and 69.9\% for the entire pointing + non-pointing + non-action frames.  From these results, we can conclude that our DeePoint trained with DP Dataset generalizes reasonably well to new scenes and subjects.

\begin{figure}[t]
    \centering
    \includegraphics[width=\linewidth]{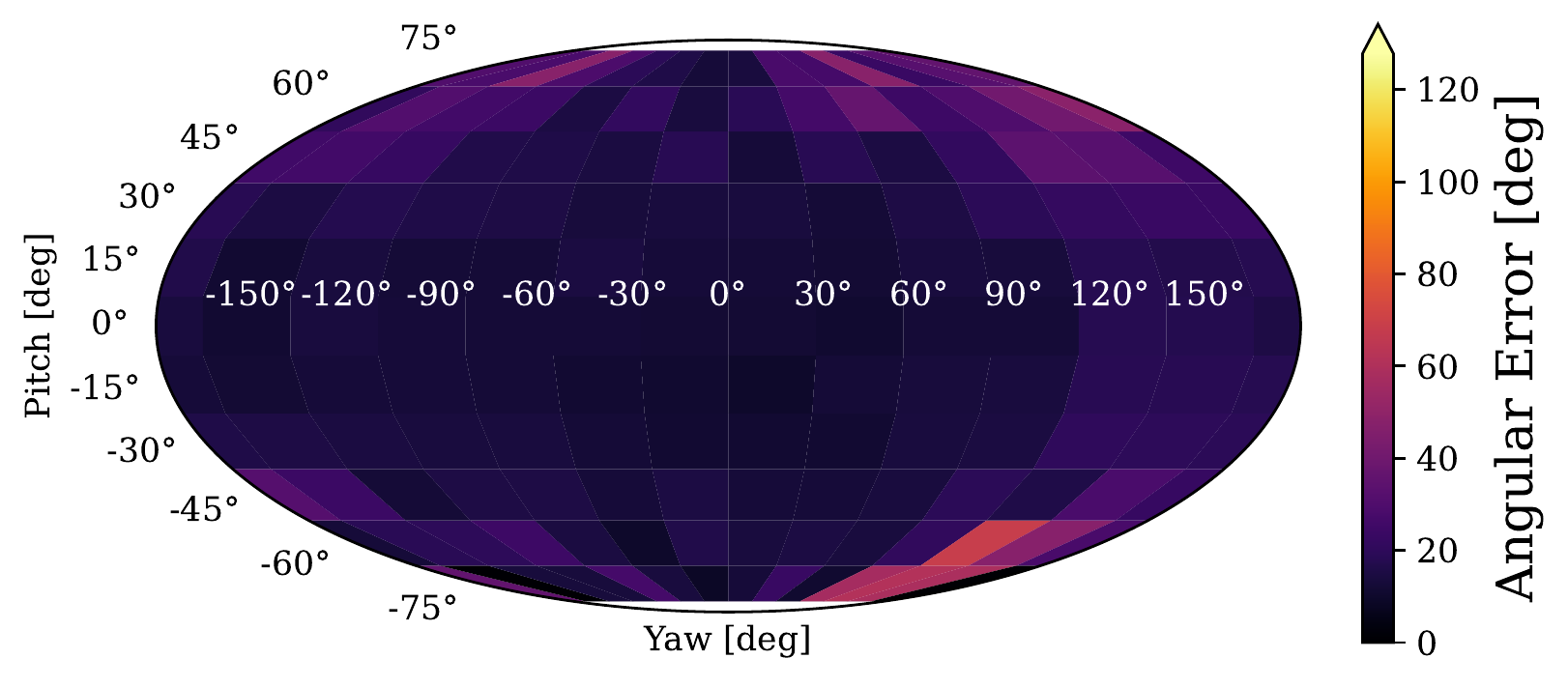}
    \caption{Mean angular error (\ie, 3D direction estimate accuracy) distribution for each ground truth pointing direction. The results show that errors increase for pointing with high/low pitches.}
    \label{fig:projection_angular_error}
\end{figure}

\subsection{3D Direction Accuracy Across Pointing Directions}

To better understand the 3D direction estimation accuracy of DeePoint, we evaluate the relationship between the ground truth pointing directions and the angular error of 3D direction estimates.
\Cref{fig:projection_angular_error} shows the angular error as a distribution over ground-truth pointing directions in Mollweide projection. 
The results show that DeePoint struggles with pointing with high yaw ($>120^\circ$) with high/low pitches ($>60^\circ$ or $<-60^\circ$).
The error is especially high with low pitches, where the person points down while facing away, which means the arms are often occluded by the body.

\section{Conclusion}
In this paper, we introduced a novel method for pointing recognition and 3D direction estimation. DeePoint leverages the spatio-temporal coordination of a person's body to recognize and estimate the timing and direction of pointing from video frames captured from a fixed-view in a relatively far distance. We also introduced the DP Dataset, the first large-scale visual pointing dataset with full annotation of the timings and 3D directions of natural pointing behaviors of a variety of people in different scene contexts. We believe these two fundamental contributions significantly advance visual pointing understanding and serve as a sound foundation for human behavior and intent understanding. We make all the data and code publicly available to catalyze further advances in this field.

\vspace{-12pt}
\paragraph{Limitation}
DeePoint can incorporate scene context but only as 2D images from the fixed viewpoint. 
Our future work includes incorporating such explicit visual cues of the environment, \eg, object detection in the scene to aid in narrowing down the exact object the person is pointing to. We also plan to explore the use of audio, particularly spoken words for this. Incorporating more scene context in these forms has the danger of overfitting to the particular context. We believe DeePoint provides a robust springboard for these further studies. 

\vspace{-12pt}
\paragraph*{Acknowledgement}
This work was in part supported by
JSPS 
20H05951, 
21H04893, 
JST JPMJCR20G7, 
and RIKEN GRP.

\appendix

\def\OursA{\textit{{DP}}\xspace}
\def\OursAwoTE{\textit{{DP w/o TE}}\xspace}
\def\OursAwHH{\textit{{DP-Hand\&Head}}\xspace}
\def\OursAwH{\textit{{DP-Hand}}\xspace}

\ificcvfinal\thispagestyle{empty}\fi



\section{Contribution of \TE}

Our model is composed of two transformer encoders, namely \JE and \TE, cascaded one after another. We evaluate the contribution of \TE by ablating \TE (\OursAwoTE) and comparing it with the full DeePoint (\OursA). 
In \OursAwoTE, the output of \JE is 
concatenated and fed into an MLP instead of being processed by \TE. The hidden layer sizes of the MLP is set to $(2880, 960,960,192)$, to make the number of parameters roughly the same as that of \TE (The number of parameters of the MLP is 3.8 million, while that of \TE is 3.1 million).

\Cref{tab:results_notmp} shows the results of this ablation comparison.
Although \TE requires less parameters than the MLP, they perform better in terms of both precision/recall and angular error, which demonstrates that \TE explicitly and efficiently incorporates temporal coordination.

\begin{table}[t]
    \centering
    \begingroup
    \setlength{\tabcolsep}{1.5mm}
    \footnotesize 
    \begin{tabular}{@{}llll@{}}\toprule
        Model         & Angular error $(\downarrow)$ & Prec./Rec. $(\uparrow)$         \\
        \midrule
        \OursA (Ours) & $\mathbf{14.05}^\circ$       & $\mathbf{0.625}/\mathbf{0.838}$ \\
        \OursAwoTE    & $14.36^\circ$                & $0.610/0.796$                   \\
        \bottomrule
    \end{tabular}
    \endgroup
    \vspace{1mm}
    \caption{Comparison between DeePoint and a variant with \TE 
    replaced with an MLP (\OursAwoTE). 
    \OursA performs better in both angular error and precision/recall of pointing recognition than \OursAwoTE, with a smaller number of parameters.  Modeling temporal movements and their coordination is essential for pointing recognition, which is successfully and efficiently achieved with \TE in DeePoint.}
    \label{tab:results_notmp}
\end{table}

\begin{table}[t]
    \centering
    \begingroup
    \setlength{\tabcolsep}{1.5mm}
    \footnotesize 
    \begin{tabular}{@{}llll@{}}\toprule
        Model         & Angular error $(\downarrow)$ & Prec./Rec. $(\uparrow)$ \\
        \midrule
        \OursA (Ours) & $\mathbf{14.05}^\circ$       & $\mathbf{0.625/0.838}$  \\
        \OursAwHH     & $15.12^\circ$                & $0.613/0.813$           \\
        \OursAwH      & $17.32^\circ$                & $0.601/0.797$           \\
        \bottomrule
    \end{tabular}
    \endgroup
    \vspace{1mm}
    \caption{Ablation of body parts. DeePoint with access to all body parts performs the best in both F-measure and mean angular error. Encoding the appearance, movements, and spatio-temporal coordination of all joints is important for accurate pointing recognition and 3D direction estimation.}
    \label{tab:results_handhead}
\end{table}

\section{Contribution of Body Parts}
In DeePoint, the appearance, movements, and spatial coordination of body parts are encoded by \JE. 
It opportunistically uses as many body parts as detected by pose estimation including the hand, head, elbow, and shoulder joints.
We evaluate the importance of encoding these body parts by ablating them. This is achieved by allowing \JE to have access to only limited detected keypoints by hiding other keypoints with masked attention. In \OursAwHH, the model has access to the keypoints that correspond to the head and the hands, that is, the nose, left/right eyes, left/right ears, and left/right hands.
In \OursAwH, it can only use the keypoints of the left/right hands.

\Cref{tab:results_handhead} shows the results of this ablation study of body parts.
The results show that the encoded tokens from the hand and head are not enough to recognize pointing and estimate its 3D directions. Having access to the tokens of the head or other body parts and joints contributes to improving the accuracy of estimation. These results clearly show that \JE successfully leverages these spatial body configurations encoded in the arrangement of body parts and also eloquently shows that pointing is a full-body gesture.

{\small
\bibliographystyle{ieee_fullname}
\bibliography{egbib}
}

\end{document}